\documentclass[lettersize,journal]{IEEEtran}
\usepackage{amsmath,amsfonts}
\usepackage{algorithmic}
\usepackage{algorithm}
\usepackage{array}
\usepackage[caption=false,font=normalsize,labelfont=sf,textfont=sf]{subfig}
\usepackage{textcomp}
\usepackage{stfloats}
\usepackage{url}
\usepackage{verbatim}
\usepackage{graphicx}
\usepackage{cite}
\usepackage{multirow}
\usepackage{hyperref}
\begin{document}

\title{SE-VGAE: Unsupervised Disentangled Representation Learning for Interpretable Architectural Layout Design Graph Generation}

\author{Jielin Chen, Rudi Stouffs
        % <-this % stops a space
\thanks{The authors are with the Department of Architecture, National University of Singapore, Singapore. E-mail: chen.jielin@u.nus.edu, stouffs@nus.edu.sg}% <-this % stops a space
\thanks{The computational work for this article was performed on resources of the National Supercomputing Centre, Singapore (https://www.nscc.sg). The data sources used in this study are also gratefully acknowledged. This research was supported by the President's Graduate Fellowship of the National University of Singapore and the Singapore Data Science Consortium (SDSC) Dissertation Research Fellowship.

(Corresponding authors: Jielin Chen and Rudi Stouffs)
}% <-this % stops a space
\thanks{Manuscript received June 17, 2024}}%; revised August 16, 2021.

% The paper headers
\markboth{Journal of \LaTeX\ Class Files,~Vol.~x, No.~x, June~2024}%
{Shell \MakeLowercase{\textit{et al.}}: Disentangled Graph Representation Learning for Architectural Layout Design Generation}

% \IEEEpubid{0000--0000/00\$00.00~\copyright~2024 IEEE}
% \IEEEpubidadjcol
% Remember, if you use this you must call \IEEEpubidadjcol in the second
% column for its text to clear the IEEEpubid mark.

\maketitle

\begin{abstract}

Despite the suitability of graphs for capturing the relational structures inherent in architectural layout designs, there is a notable dearth of research on interpreting architectural design space using graph-based representation learning and exploring architectural design graph generation. Concurrently, disentangled representation learning in graph generation faces challenges such as node permutation invariance and representation expressiveness. To address these challenges, we introduce an unsupervised disentangled representation learning framework, Style-based Edge-augmented Variational Graph Auto-Encoder (SE-VGAE), aiming to generate architectural layout in the form of attributed adjacency multi-graphs while prioritizing representation disentanglement. The framework is designed with three alternative pipelines, each integrating a transformer-based edge-augmented encoder, a latent space disentanglement module, and a style-based decoder. These components collectively facilitate the decomposition of latent factors influencing architectural layout graph generation, enhancing generation fidelity and diversity. We also provide insights into optimizing the framework by systematically exploring graph feature augmentation schemes and evaluating their effectiveness for disentangling architectural layout representation through extensive experiments. Additionally, we contribute a new benchmark large-scale architectural layout graph dataset extracted from real-world floor plan images to facilitate the exploration of graph data-based architectural design representation space interpretation. This study pioneered disentangled representation learning for the architectural layout graph generation. The code and dataset of this study will be open-sourced.

\end{abstract}


\section{Introduction}

\IEEEPARstart{A}rchitectural design solutions inherently possess structured information with interdependent scopes, making architectural design data intrinsically relational and suitable for graph-based representations \cite{chen2022AAG-FP}. Graph-structured representations are optimal for accurately depicting complex geometric and semantic information in architectural designs \cite{bronstein2017geometric}, providing an abstract yet robust format for encoding design features and their interrelationships. This suitability extends from macro-level layouts to micro-level construction details, allowing for an interconnected view of design elements and illustrating their cohesive functioning. Architectural design space serves as a fundamental concept in design research \cite{akin2006whittled,woodbury2006whither, gero2013modeling, cross2023design, chen2023atlas}, yet the representation of architectural design space and interpretation of corresponding design representation space using deep learning-based approaches \cite{chen2023deciphering}, especially concerning structured non-Euclidean data like graphs, remains understudied. Most existing studies focus on Euclidean design data formats \cite{chaillou2020archigan, chen2023deciphering, koh2022voxel}, leaving a gap in exploring graph-based representation learning and synthesis in architectural design.

Recent advancements in disentangled representation learning for neural network-based graph generation aim to extract distinct generative factors in observed graph data, crucial for understanding real-world graph distributions. Although studies have shown the potential of disentanglement in deep graph representation learning \cite{stoehr2019disentangling, guo2020interpretable, du2022disentangled}, challenges like overlooking permutation invariance and limited expressiveness remain. Additionally, while significant progress has been made in domains like molecule and protein generation \cite{guo2020interpretable, guo2022systematic, simonovsky2018graphvae, ma2018constrained}, the application of these techniques in architectural design graphs is still largely unexplored.

To bridge these significant research gaps, We propose the \textbf{Style-based Edge-augmented Variational Graph Auto-Encoder (SE-VGAE)}, an unsupervised disentangled representation learning framework for decomposing latent generative factors of architectural layout design graphs represented in the form of attributed adjacency multi-graphs (AAMG). The framework includes three alternative disentanglement pipelines (Fig.~\ref{fig:FP-GNN-frameworks}). All three pipelines are composed of a transformer-based edge-augmented encoder with permutation equivariance property to integrate both node and edge features, a style-based decoder with two sub-decoders (a node-decoder and an edge-decoder) incorporating a layer-wise stochasticity feature decoding strategy, and a latent space disentanglement module. The latter facilitates the decomposition of latent factors influencing architectural layout graph generation. The three alternative disentanglement modules are a vanilla VAE scheme serving as baseline, a Vector Quantisation (VQ) scheme modelling probability density functions through prototype vectors, and a node-edge co-disentanglement scheme to separate features at node, edge, and graph levels using three specialized sub-encoders.

To the best of our knowledge, this study is the first to generate architectural layout design graphs with a focus on representation disentanglement. We investigate various architectural layout graph feature augmentation schemes and their impact on model performance, exploring different framework setups and structural adjustments to understand their influence on interpreting and disentangling complexities in the architectural layout design graph data space. In addition to framework development, we introduce a novel benchmark large-scale architectural layout graph dataset featuring detailed node and edge attributes extracted from real-world floor plan images. Using this dataset, we uncover latent design patterns and relationships through our proposed disentangled graph representation learning schemes. Additionally, we explore interpreting latent architectural design representation space by extracting high-level structural information using graph data. The contribution of this study can be summarized as follows:

\begin{figure}[!t]
  \centering
  \includegraphics[width=1.0\linewidth]{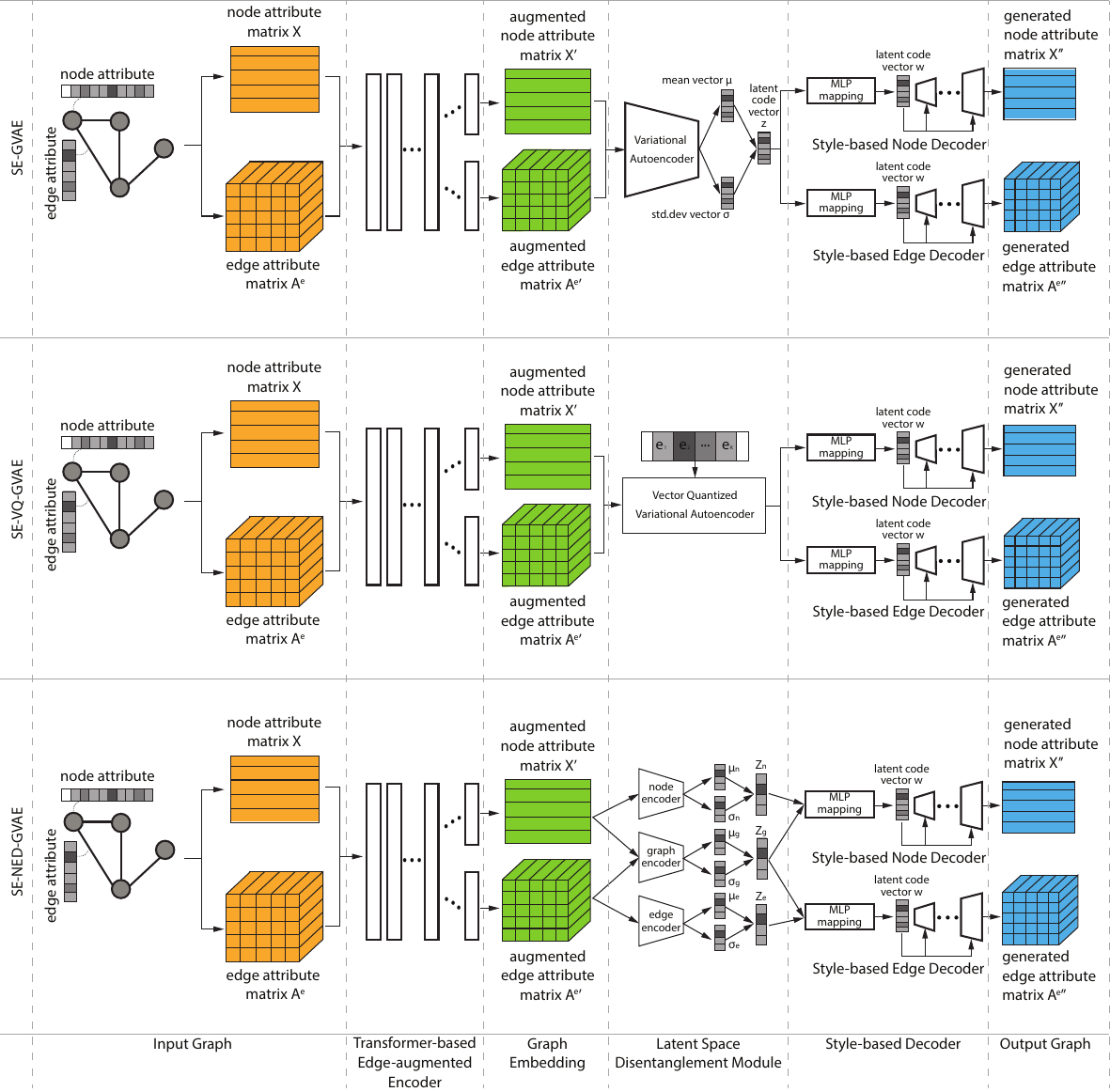}
  % \vspace{\BeforeCaptionVSpace}
  \caption{Overview of the proposed \textbf{Style-based Edge-augmented Variational Graph Auto-Encoder (SE-VGAE)} framework, together with three alternative pipelines, for the latent embedding space disentanglement of architectural layout design representation space}
  \label{fig:FP-GNN-frameworks}
\end{figure}

\begin{itemize}
    \item Introducing the SE-VGAE framework, a pioneering effort to generate architectural layout design graphs with disentanglement, which provides insights into optimizing model structures for interpreting architectural design graph spaces through varied implementations.
    \item Systematically investigating various architectural layout graph feature augmentation schemes and their impact on graph generation and representation disentanglement. Extensive experiments elucidate the efficacy of different augmentation strategies in improving the model's understanding of architectural layout graph complexities.
    \item Offering a new benchmark large-scale architectural layout design graph dataset from real-world floor plan images. This dataset is a valuable resource for training and evaluating disentangled graph representation learning models, enabling researchers to explore latent design patterns and relationships and extract high-level structural information for architectural design data interpretation.
\end{itemize}

\section{Related Work}

This section briefly reviews the research background pertinent to this study, encompassing disentangled graph representation learning, graph generation and evaluation, architectural layout design representation as graphs, and the relevant subject of architectural design representation space interpretation.

\subsection{Disentangled graph representation learning}
\label{dgrl}

Bengio et al. \cite{bengio2013representation} define disentangled representation as the separation of distinct, independent, and informative generative factors in observed data, crucial for understanding real-world data distributions, including complex graph structures. This concept is essential in deep graph representation learning models, where it is beneficial to discern which latent variables influence specific graph generation properties. Studies using disentanglement-oriented neural networks have demonstrated potential in this area. Stoehr et al. \cite{stoehr2019disentangling} used $\beta$-Variational Autoencoders ($\beta$-VAE) \cite{kingma2014auto} to discover generative parameters in graphs but neglected edge features and node order independence, compromising reconstruction fidelity. Guo et al. \cite{guo2020interpretable, du2022disentangled} addressed some of these issues with NED-VAE, a framework with sub-encoders and sub-decoders for disentangling node, edge, and graph-level features, though they overlooked graph node permutation invariance and did not use highly expressive graph aggregation methods. The expressive power of graph representation learning is crucial for identifying and differentiating subtle variations within graph structures. Graph neural networks (GNNs) extend the Weisfeiler-Lehman (WL) isomorphism test by representing graphs as vectors in continuous space, capturing relationships between different topologies \cite{li2022expressive}. However, conventional GNNs are only as powerful as the 1-dimensional WL test \cite{xu2018powerful, morris2019weisfeiler}. Zhang et al. \cite{zhang2023expressive} categorize efforts to enhance GNN expressiveness into three approaches: graph feature enhancement \cite{barcelo2020expressive}, graph topology enhancement \cite{you2019position, murphy2019relational, bouritsas2022improving, morris2019weisfeiler}, and model architecture enhancement \cite{morris2019weisfeiler, bouritsas2022improving}. While model architecture enhancement increases complexity and parameters, feature and topology enhancements are more lightweight and easier to implement. These enhancements include adding local and global topological information to each node and using random node attributes or positional information to improve representation \cite{murphy2019relational, sato2021random, you2019position, srinivasan2019equivalence}.

\subsection{Evaluating interpretable deep graph generation}
\label{gge}

Robust quantitative evaluation is crucial for graph generative modelling, focusing on the difference between learned and reference graph distributions. Evaluating these distributions is challenging due to the unique properties of graph data. Traditional methods calculate the statistical distribution distance between real and generated graphs but often overlook continuous node and edge features \cite{thompson2022evaluation}. Recently, neural network classifier-based metrics have gained popularity for aligning learned and real graph distributions \cite{liu2019auto, thompson2022evaluation}. However, metrics from image generation, which use task-specific neural networks, have limitations in graph generative modelling due to the adaptability issues of pre-trained graph neural networks. To address this, studies have shown that randomly initialized graph neural networks can effectively evaluate graph generative models without the need for further training \cite{morris2019weisfeiler, xu2018powerful, thompson2022evaluation}.

\subsection{Representing architectural layout design as graphs}

Architectural layout design can be naturally represented as graphs. Specifically, a floor plan layout can be converted into a dual graph, emphasizing space adjacency with nodes as spaces and edges as connectivity \cite{park2019dirksen, chen2022AAG-FP}. These adjacency graphs can also embed three-dimensional architectural information \cite{as2018artificial}. Neural network-based graph generation is a burgeoning field explored in various domains like molecules, protein structures, and scene graphs \cite{guo2020interpretable, guo2022systematic, simonovsky2018graphvae, ma2018constrained}. However, in architectural design research, while many studies focus on generating architectural data in Euclidean formats like images and 3D models \cite{chaillou2020archigan, chen2023deciphering, koh2022voxel}, the specific task of generating architectural layout design graphs has been notably unexplored.

\subsection{Architectural design representation space interpretation}

Design space is crucial in architectural design research, with its exploration often used to approximate the design process \cite{akin2006whittled,woodbury2006whither, gero2013modeling, cross2023design, chen2023atlas}. Previous studies primarily focused on superior exploration strategies, leaving the intrinsic structure of design representation space vague \cite{woodbury2006whither}. As design activities are made possible because of designers' mental models of design representation spaces that designers constantly perceive and formulate \cite{woodbury2006whither, cross2023design, chen2023atlas}, Chen \& Stouffs \cite{chen2023atlas, chen2023deciphering} promote two explicit models of design representation spaces: the sparse human-learned model and the compressed machine-learned model, arguing that designers may enhance design performance by interacting with simulated design representation spaces. In this context, converting architectural design data into machine-interpretable formats is necessary, requiring flexible representation learning schemes. Recent data-driven techniques have demonstrated the ability to interpret design concepts by converting data into vectors of neural activities \cite{chen2023deciphering, kim2023text2form}. However, current studies on architectural design representation focus mainly on Euclidean data, with limited exploration of non-Euclidean data like graphs.

\section{Preliminary}

This section introduces the relevant notations used in this study in Section \ref{notations}, the problem formulation of disentangled graph representation learning of architectural layout design graph generation in Section \ref{problem_formulation}, and an overview of our proposed approach in Section \ref{overview}.

\subsection{Notations}
\label{notations}

Formally, a graph is denoted as $G = \left( V, E\right)$ where $V$ is the set of nodes, and $E$ is the set of edges. We denote an edge going from node $v_{i}\in V$ to node $v_{j}\in V$ as $\left( v_{i}, v_{j}\right)\in E$. This study does not consider self-loop edges, namely a node connecting to itself. The number of nodes $n = \left| V\right|$ is called the order of graph $G$, and the number of edges $e=\left| E\right|$ is called the size of graph $G$. In this study, we consider multi-graphs within which there can be more than one edge between a pair of nodes. Also, only undirected graphs are discussed, s.t., $\left(v_{i}, v_{j}\right) \in E\Leftrightarrow \left(v_{j}, v_{i}\right) \in E$.

A convenient way to represent a graph is through an adjacency matrix. An adjacency matrix $A$ for a multi-graph is a symmetric square matrix with $A_{u,v}=a$ if $\left( u, v\right)\in E$ and $a$ is the number of edges connecting nodes $u$ and $v$. To represent a graph $G$ with an adjacency matrix $A$, the node set $V$ of graph $G$ needs to be ordered previously so that every node indexes a particular row and column in the adjacency matrix $A$. There are $n!$ possible node orderings for a graph with order $n$, each corresponding to a unique, arbitrary node ordering $\pi$. Thus, if we choose an ordering $\pi$, the graph can be represented by the corresponding adjacency matrix $A^{\pi }\in \mathbb{R} ^{n\times n}$. Considering the multiplicity of possible representations of a single graph, this necessitates the formulation of training mechanisms of graph representation learning models invariant or equivariant to different node permutations of the same graph. That is to say, any arbitrary node permutations of the same graph should result in identical graph representations, and ideally, deep graph representation learning and generative modelling need to learn permutation-invariant graph distributions.

\subsection{Problem formulation}
\label{problem_formulation}

The learning objective of deep graph representation learning, especially a graph generative model, is to maximize the likelihood of $p\left( G\right) =\sum _{\pi }P\left( G,\pi \right)$. However, while graph encoding can abstract away the ordering of nodes with permutation-invariant node aggregation operations, graph decoding must establish certain node orderings as concrete expressions. Under relatively lenient conditions, graph decoders can attain permutation-equivariance; when presented with a permuted graph, the graph generative model can produce correspondingly permuted graph representations. For sequential generation methods, attaining permutation equivariance is far from trivial and presents a complex challenge, yet achieving this can be straightforward for one-shot generation methods. This is usually done by redefining one or a series of node ordering (e.g., breadth-first search, depth-first search, node degree, or a family of canonical orderings) \cite{liu2019auto, liao2019efficient}. Specifically, canonical ordering refers to systematically arranging the nodes of a graph according to specific rules or algorithms, resulting in a consistent and standardized ordering \cite{liao2019efficient}. A family of canonical orderings can be predefined as $K = \left\{ \pi _{1},\ldots,\pi _{k}\right\}$ and can be used to learn an evidence lower bound (ELBO) of $\sum _{\pi }P\left( G,\pi \right)$, namely $\sum _{\pi \in K}P\left( G,\pi \right)$, as $K$ is a strict subset of the full factorial range of node orderings. It is also a tighter lower bound than any single arbitrary canonical ordering likelihood $P\left( G,\pi \right)$. Additionally, enlarging the size of $K$ can result in a tighter lower bound. Selecting an appropriately sized set $K$ can thus strike an optimal balance between the tightness of the bound – which typically corresponds to improved model quality – and the computational costs involved.

Meanwhile, a graph can have both node and edge attributes; such a graph is referred to as an attributed graph. An attributed graph is defined as $G=\left(V, E, X, A^{e}\right)$. The node feature matrix is denoted as $X\in \mathbb{R} ^{n\times d}$, where we assume that the ordering of the nodes is consistent with the ordering in the corresponding adjacency matrix $A$, with $x_{\nu }\in \mathbb{R}^{d}$ denoting the feature vector of node $v$, while $d$ is the dimension of the node attributes. While the edge feature matrix can be denoted as $A^{e}\in \mathbb{R} ^{n\times n \times c}$, with $x_{u,v}^{e}\in \mathbb{R} ^{c}$ representing the feature vector of edge $\left( u, v\right)\in A$, while $c$ is the dimension of the edge attributes. The edge feature matrix can also be understood as the original adjacency matrix $A$ with the edge feature dimension $c$ added to each $A_{u,v}$.

Given a set of observed graphs, $D_{G}=\left\{ G_{1}, \ldots, G_{s}\right\}$ with underlying data distribution $p\left( G\right)$, where each graph $G_{i}$ may have different order $n_{i}$ and size $e_{i}$, we have $G\sim p\left( G\right)$ for each graph $G$ in the dataset. The goal is to have a deep graph representation learning model that is able to learn a close enough estimation $p_{model}\left( G\right)$ of the real graph distribution $p\left( G\right)$ without being constrained to a predetermined order or size of the graphs. Such a model would be capable of generating novel, previously unseen graphs of various orders and sizes drawn from the learned probabilistic model $p_{model}\left( G\right)$.

The representation mapping encoders and decoders are essential for acquiring such graph representation learning models. Formally, a graph encoder, denoted as $f\left(z|G\right)$, maps a real discrete graph object as a dense, continuous vector $z$ of a low-dimensional stochastic latent space that follows a prior distribution $p\left( z\right)$; the graph encoder $f\left(z|G\right)$ outputs the parameters of the stochastic distribution. While a graph decoder, denoted as $f\left(G|z\right)$, accepts a latent vector $z\sim p\left( z\right)$ sampled from the same stochastic distribution $p\left( z\right)$ and performs the inverse function of the graph encoder. Graphs and corresponding features are transformed into a continuous vector space during encoding. Translating continuous data representations back into discrete graph structures, including nodes and edges, is non-trivial. This reconstruction task can take various forms, ranging from the sequential generation of the nodes and edges of the graphs step by step to the one-shot generation of the adjacency matrices or edge lists. Sequential generation leverages local decision-making efficiently and is flexible when the number of nodes is unknown. However, it struggles with maintaining long-range dependencies, which can result in omitting crucial global graph properties. Conversely, one-shot generation can capture a graph's global properties by simultaneously generating and refining the entire graph structure across multiple iterations \cite{guo2022systematic}. This study adopts the one-shot generation method, as it can learn to map entire graphs into unified latent representations and generate an entire graph directly through a single-step sampling, allowing the extraction of crucial global graph properties without sacrificing computational efficiency.

The ultimate problem this study tries to solve is disentangling local and global graph generative dependencies, which can provide insights into architectural layout design graph topologies. Ideally, upon mastering a latent space that accurately represents the distribution of real graphs, one can sample new latent code $z\sim p\left( z\right)$ from this space to control the characteristics of the generated graphs. Disentangled sampling can be then applied by segmenting the latent vector $z$ into distinct dimensions, with each dimension $z_{n}$ focusing on a unique property. As a result, altering a single latent dimension $z_{n}$ can induce specific property changes in the generated graphs, enabling precise manipulation of graph characteristics.

\subsection{Method overview}
\label{overview}

This study introduces the \textbf{Style-based Edge-augmented Variational Graph Auto-Encoder (SE-VGAE)}, a novel framework designed for unsupervised disentangled representation learning aimed at automatically decomposing latent generative factors within architectural layout design graphs represented in the form of attributed adjacency multi-graphs. The framework comprises three alternative disentanglement pipelines tailored for interpreting the layout design graph data space (refer to Fig.~\ref{fig:FP-GNN-frameworks}). Each pipeline consists of three primary components.

The first component features a transformer-based edge-augmented encoder designed with permutation equivariance to integrate both node and edge features. This encoder takes the node feature matrix and the edge feature matrix of an attributed adjacency multi-graph as input, producing updated node and edge embeddings integrating both local and global nodes and edges' features as output. Addressing the node ordering challenge, we employ a selected family of canonical orderings, enabling the model to consider various orderings with distinct structural biases while circumventing the challenges associated with factorial permutations.

The latent space disentanglement module follows the edge-augmented encoder. It is essential for decomposing latent factors influencing the interpretation and disentanglement of architectural layout design graph representations. This module employs Graph Isomorphism Network (GIN) \cite{xu2018powerful} layers as building blocks, known for their superior expressive power and ability to generalize the Weisfeiler-Lehman (WL) test. This module takes the node and edge embeddings as input and outputs latent code vectors, the specifics of which vary depending on the chosen module scheme. We leverage different disentanglement regularisation methods to guide the representation disentanglement process and promote independence among the learned latent variables. Specifically, we propose three alternative disentanglement module schemes: 1) a vanilla VAE scheme that outputs a single latent code vector embedding the entire input graph, 2) a Vector Quantization (VQ) scheme that models probability density functions through the distribution of prototype vectors, resulting in a quantized latent code vector embedding the entire input graph, and 3) a node-edge co-disentanglement scheme utilizing three specialized sub-encoders to separate features at node, edge, and graph levels, outputting three latent code vectors embedding node, edge, and graph level features of the input graph, respectively.

The final component is a style-based decoder, which incorporates the layer-wise stochasticity feature decoding strategy \cite{karras2019style} by introducing stochastic variations at different layers of the network. The decoder consists of two sub-decoders: a node-decoder and an edge-decoder. The node-decoder reconstructs node features by translating the provided latent representation back into the node-specific attributes of the graph, while the edge-decoder focuses on reconstructing edge features, converting the latent representation into meaningful edge attributes of the graph.

To the best of our knowledge, this study presents a pioneering effort in generating architectural layout design graphs with a primary focus on representation disentanglement. Unlike previous approaches, our work takes into account both node and edge features, as well as the critical issue of graph node permutation invariance. By integrating graph aggregation methods with high expressive power, we aim to acquire high-quality learned graph representations at various levels and, consequently, high effectiveness of feature disentanglement.

\section{Style-based Edge-augmented Variational Graph Auto-Encoder}

In this study, one essential aspect is striking a balance between expressiveness and efficiency when mapping from the spaces of adjacency matrices and node feature matrices to a condensed latent space, as well as the simultaneous generation of graph topology and node/edge attributes. Given that a graph $G$'s topology can be conveniently represented through an edge feature matrix $A^{e}$ (adjacency matrix embedded with edge features) in tandem with a node feature matrix $X$, a prevalent approach is to model the distribution of these matrices in a unified, seamless process \cite{guo2022systematic}. As the one-shot generation method can effectively handle global patterns of graphs, and it is vital to capture global patterns for interpreting architectural layout design graph data, we adopt the adjacency matrix-based one-shot generation approach. We propose a flexible variational autoencoder-based graph representation learning framework designed to learn latent variable distribution at node, edge, and graph levels. The model encodes a comprehensive range of architectural features and relationships inherent in the layout design graph data by simultaneously capturing the nuances at these different levels. Adopting a flexible VAE-based framework paves the way for further systematic implementation and evaluation of a series of structural interventions concerning the model structure. Concretely, we propose \textbf{Style-based Edge-augmented Variational Graph Auto-Encoder (SE-VGAE)}, together with three alternative pipelines, for the latent embedding space disentanglement of architectural layout design graph data, as shown in Fig.~\ref{fig:FP-GNN-frameworks}. 

All three alternative pipelines comprise a transformer-based edge-augmented encoder, a latent space disentanglement module and a style-based decoder. The edge-augmented encoder inputs the node feature matrix $X$ and the edge feature matrix $A^{e}$ of an attributed adjacency multi-graph $G$ of an architectural layout design. The input graph undergoes pre-processing using a predefined family of canonical orderings, allowing the model to account for different orderings with unique structural biases while avoiding the computational challenges of the full space of factorially-many permutations \cite{liao2019efficient}. The encoder outputs the correspondingly updated node feature matrix $X'$ and edge feature matrix $A^{e'}$ with augmented node and edge embeddings that integrate the nodes and edges' intricate relationships and features from local and global levels. Details of the encoder component are further discussed in section \ref{encoder}. The augmented node and edge feature matrices $X'$ and $A^{e'}$ then serve as input to the latent space disentanglement module. We propose three alternative latent space disentanglement modules, offering different disentanglement regularisation methods to guide the representation disentanglement process and promote independence among learned latent variables. We elaborate on the details of the disentanglement modules in section \ref{disentanglement_module}. The disentanglement module further outputs one or three compressed latent code vectors $\textbf{z}$ of the given graph $G$. The specifics of latent code vectors vary depending on the alternative module scheme. The latent code vector(s) $\textbf{z}$ will then be used as the input of the style-based decoder, which generates a node feature matrix $X''$ and edge feature matrix $A^{e''}$ as the final outputs. A more in-depth discussion of the style-based decoder is provided in section \ref{decoder}.

\subsection{Edge-augmented encoder}
\label{encoder}

For the edge-augmented encoder of our proposed model frameworks, we leverage the Edge-augmented Graph Transformer (EGT) \cite{hussain2022global} as the backbone to integrate both node and edge features. The EGT backbone inherits the permutation equivariance characteristic from the original transformer mechanism \cite{vaswani2017attention} and employs global self-attention as its primary aggregation mechanism. This approach markedly differs from the conventional static, localized convolutional node aggregation, enabling the model to facilitate unconstrained long-range dynamic interactions between nodes. A key aspect of the transformer-based edge-augmented encoder is its ability to handle both node and edge features within a unified framework. The residual channels of the original transformer structure are utilized as node channels, while additional edge channels enable the graph's edge information to evolve across different layers of the model, allowing the model to dynamically update and refine the representation of both nodes and edges through successive layers. As a result, the encoder continuously updates both node and edge embeddings at each layer.

The edge-augmented encoder, denoted as $f\left((X', A^{e'})|(X, A^{e})\right)$, intakes the node feature matrix $X\in \mathbb{R} ^{n\times d}$ and the edge feature matrix $A^{e}\in \mathbb{R} ^{n\times n \times c}$ of an attributed adjacency multi-graph $G=\left(V, E, X, A^{e}\right)$ as input, and produces updated node feature matrix $X'\in \mathbb{R} ^{n\times d}$ and edge feature matrix $A^{e'}\in \mathbb{R} ^{n\times n \times c}$ with respectively augmented node and edge embeddings, incorporating intricate relationships and features from both local and global levels of the nodes and edges. Specifically, $x_{v}\in \mathbb{R}^{d}$ denotes the feature vector of node $v$, while $d$ is the dimension of the node features, and $x_{u,v}^{e}\in \mathbb{R} ^{c}$ represents the feature vector of edge $\left( u, v\right)\in E$, while $c$ is the dimension of the edge features.

Concretely, at the $o$-th attention head of the $l$-th layer of the $L$-layer encoder, the attention mechanism is defined as follows: $Attn\left(Q_{n}^{o,l}, K_{n}^{o,l}, V_{n}^{o,l}\right) = softmax\left(clip\left( \dfrac{Q_{n}^{o,l}\cdot\left( K_{n}^{o,l}\right)^{T}}{\sqrt{b_{k}}}\right) + E_{e}^{o,l}\right) \odot \sigma\left( G_{e}^{o,l}\right) \cdot V_{n}^{o,l}$. Here, $Q_{n}^{o,l}, K_{n}^{o,l}, V_{n}^{o,l}\in \mathbb{R} ^{n\times b_{k}}$ represent the queries, keys, and values obtained from linear transformations of node embeddings, with $Q_{n}^{o,l}\cdot\left( k_{n}^{o,l}\right)^{T}\in \mathbb{R} ^{n\times n}$ denoting the dot product of $Q_{n,}^{o,l}$ and $ K_{n}^{o,l}$, $b_{k} = d/O$ is the dimension of the keys for normalizing the dot product and $O$ is the total number of attention heads. The normalized dot product is clipped to a certain range for better numerical stability ($\left[ -5,+5\right]$ is used following \cite{hussain2022global}). $E_{e}^{o,l}, G_{e}^{o,l}\in \mathbb{R} ^{n\times n}$ are the learned linear transformations of the edge embeddings. $E_{e}^{o,l}$ acts as a bias term added to the normalized dot product of the queries and keys of the node embeddings, enabling edge embeddings to influence node embedding attention. While $\sigma \left( G_{e}^{o,l}\right)\in \mathbb{R} ^{n\times n}$ gates the softmax values before aggregation, regulating information flow between nodes, $\odot$ is the element-wise product operation and $\cdot$ is matrix multiplication. With $O$ number of attention heads in total, we have,

\begin{equation}
\label{eq_encoder_edge1}
O_{e}^{l} = ||_{o=1}^{O}\widehat{H}^{o,l}, O_{e}^{l} \in \mathbb{R} ^{n\times n \times O}
\end{equation}
\begin{equation}
\label{eq_encoder_edge2}
where, \widehat{H}^{o,l} = clip\left( \dfrac{Q_{n}^{o,l}\odot\left( K_{n}^{o,l}\right) ^{T}}{\sqrt{b_{k}}}\right) + E_{e}^{o,l}
\end{equation}
\begin{equation}
\label{eq_encoder_edge3}
A^{e,l+1} = LN\left(FFN\left(LN\left(\widehat{A}^{e,1}\right)\right) + \widehat{A}^{e,1}\right)
\end{equation}
\begin{equation}
\label{eq_encoder_edge4}
where, \widehat{A}^{e,1} = \widehat{O_{e}}^{1} + A^{e,1}
\end{equation}
for edge embedding updates at the $l$-th layer, and
\begin{equation}
\label{eq_encoder_node1}
O_{n}^{1} = ||_{o=1}^{O} C^{o,l} \odot \sum ^{n}\widehat{A}^{o,l} \cdot V_{n}^{o,l}, O_{n}^{1} \in \mathbb{R} ^{n\times b_{k}\times O}
\end{equation}
\begin{equation}
\label{eq_encoder_node2}
where, \widehat{A}^{o,l} =  softmax\left(\widehat{H}^{o,l}\right) \odot \sigma\left( G_{e}^{o,l}\right)
\end{equation}
\begin{equation}
\label{eq_encoder_node3}
and, C^{o,l} =  ln\left(1 + \sum^{n}\sigma\left( G_{e}^{o,l}\right)\right)
\end{equation}
\begin{equation}
\label{eq_encoder_node4}
\widehat{O_{n}}^{1} = reshape\left(O_{n}^{1}\right), \widehat{O_{n}}^{1} \in \mathbb{R} ^{n\times d}
\end{equation}
\begin{equation}
\label{eq_encoder_node5}
X^{1+1} = LN\left(FFN\left(LN\left(\widehat{X}^{1}\right)\right) + \widehat{X}^{1}\right)
\end{equation}
\begin{equation}
\label{eq_encoder_node6}
where, \widehat{X}^{1} = \widehat{O_{v}}^{1} + X^{1}
\end{equation}
for node embedding updates at the $l$-th layer, where $LN$ is layer normalization applied right before and after the attention mechanism, $FFN$ is a feed-forward network layer for learnable linear transformation, $||$ refers to the concatenation operation. $C^{o,l}$ represents the logarithm of the sum of Sigmoid-transformed edge embeddings, scaling node centrality to enhance network sensitivity and expressiveness in identifying non-isomorphic (sub-)graphs through adaptive self-attention \cite{hussain2022global}.

For an input attributed adjacency multi-graph $G=\left(V, E, X, A^{e}\right)$, the node feature embedding $\widehat{X}$ and the edge feature embedding $\widehat{A}^{e}$ are obtained through a series of learnable linear transformations using original node feature matrix $X$ and the edge feature matrix $A^{e}$, accommodating both continuous and discrete values. The edge feature embedding $\widehat{A^{e}}$ is further processed by adding the distance matrix $D^{m}$, with $D_{u,v}^{m}\in \{0,1,\ldots,m\}$ being the shortest distances between node $u$ and $v$ while clipped to the $m$-hop distance if exceed. A masking vector is employed in lieu of an edge feature for non-existing edges. The resulting node and edge embeddings are then forwarded to the latent space disentanglement module.

\subsection{Latent space disentanglement modules}
\label{disentanglement_module}

The major difference among the three alternative pipelines is the latent space disentanglement module between the transformer-based edge-augmented encoder and the style-based decoder. Generally, the latent space disentanglement module, denoted as $g\left(z|(X', A^{e'})\right)$, maps the augmented node and edge embeddings $X'$ and $A^{e'}$ to a dense, continuous vector $z$ of a low-dimensional stochastic latent space that follows a prior distribution $p\left( z\right)$; the disentanglement module $g\left(z|(X', A^{e'})\right)$ learns the parameters of the stochastic distribution. We propose three different disentanglement modules to achieve this task, all implemented in an unsupervised manner. 

\subsubsection{SE-VGAE with vanilla VAE module}

The first option of the three proposed pipelines incorporates a traditional VAE scheme (Fig.~\ref{fig:FP-GNN-frameworks}-1), which serves as the baseline of our study. A Graph Isomorphism Network (GIN) \cite{xu2018powerful} is utilized to integrate the augmented node and edge embeddings and produces two subsequent vectors: a mean vector $\mu$ and a standard deviation vector $\nu$. These vectors collectively contribute to the formation of the latent code vector \textit{z}, which encapsulates the essential features and characteristics of both node and edge representations in a distilled form while capturing the underlying patterns and structures of the graph data in a condensed latent space. The obtained latent code vector \textit{z} is then fed into the decoder part, which reconstructs the graph data from the latent representation, allowing the model to effectively learn representations of the original graph data by comparing the original graph and its reconstructed counterpart. By employing this conventional VAE scheme as our baseline, we establish a fundamental framework against which we can compare the effectiveness and efficiency of the other proposed pipelines in disentangling the latent embedding space. This baseline provides a crucial reference point for evaluating proposed alternative pipelines. Please refer to Appendix \ref{method_app} for the pseudo-code of the vanilla VAE module.

\subsubsection{SE-VQ-VGAE with Vector Quantization module}
The second framework option employs a Vector Quantization (VQ) scheme \cite{oord2017neural} (Fig.~\ref{fig:FP-GNN-frameworks}-2), a method particularly adept at modelling probability density functions through the distribution of prototype vectors. It operates by encoding values from a multidimensional vector space into a finite set of discrete values that exist within a subspace of lower dimensions, allowing for a more structured and compact representation of the graph data in the latent space. A critical aspect of the VQ scheme is its output of discrete rather than continuous latent codes produced by the baseline scheme. This shift from continuous to discrete representation can simplify the latent space, potentially making it easier for the model to learn and capture the essential features of the graph data. The major difference between the VQ-based disentanglement module and the vanilla VAE module is the post-process of the latent code $z$, namely projecting $z$ from continuous latent space $Z$ into a discrete latent space $Z_{k}$. Specifically, after obtaining the latent code $z$ using the process provided in Algorithm \ref{alg:alg1}, an intermediate embedding space $K \in \mathbb{R} ^{k\times d}$ with $k$ being the size of the discrete latent space is used to find the nearest neighbour of $z$, namely, its discrete counterpart $z_k$ in the discrete latent space $Z_{k}$. Concretely, we compute the posterior categorical distribution $q(z_{k}|z)$ with $q\left(z_{k}|z\right) =\begin{cases}1, k=argmin\left| \right| z-k_{i}\left| \right| _{2}\\ 0, otherwise\end{cases}$, where $i\in 1,2,\ldots ,k$. The quantised latent code $z_{k}$ is then used as the input of the subsequent decoder.

By introducing the VQ scheme as the second option for latent embedding space disentanglement, we can explore the potential benefits of a discrete latent space in graph representation learning and the generation of architectural layout design. This approach not only provides an alternative perspective to the continuous latent space option but may also enhance our understanding of how different latent space representations can impact the overall performance and effectiveness of the model in capturing and interpreting architectural design graph data.

\subsubsection{SE-NED-VGAE with Node-edge co-disentanglement module}

The third option capitalizes on the node-edge co-disentanglement mechanism (NED) \cite{guo2020interpretable}  (Fig.~\ref{fig:FP-GNN-frameworks}-3) to separate the intricate interplay of features at different levels of the graph using three specialized sub-encoders: a node encoder, an edge encoder, and a node-edge co-encoder. The node encoder focuses on extracting and understanding the features and characteristics unique to individual nodes within the graph, the edge encoder is responsible for capturing and representing the features and properties associated with the edges in the graph, and the node-edge co-encoder works to integrate and comprehend the combined information from both nodes and edges, thereby capturing the overall structural and relational dynamics of the graph. By employing this tripartite encoding strategy, the third option works to model the complex relationships between nodes and edges and disentangle the intertwined node and edge features within the graph. Please refer to Appendix \ref{method_app} for the pseudo-code of the NED-based disentanglement module.

\subsection{Style-based decoder}
\label{decoder}

The style-based decoder of our proposed model frameworks is constructed by incorporating the layer-wise stochasticity mechanism, a feature decoding strategy initially proposed by StyleGAN \cite{karras2019style, karras2020analyzing} for generating image data. This mechanism has proved to be effective in generating high-quality and diverse imagery by introducing stochastic variations at different layers of the network \cite{shen2020interpreting, chen2023deciphering}. By grafting the layer-wise stochasticity feature decoding strategy from StyleGAN into our proposed framework, we aim to enhance the model's capability to generate rich, diverse, and realistic architectural graph representations of both node and edge features, not only leveraging the strengths of StyleGAN's generative capabilities but also tailoring them to the specific needs of graph representation learning, which is crucial for capturing the complexity and diversity inherent in the architectural layout design graph data.

Our proposed style-based decoder is designed to include two sub-decoders: a node-decoder and an edge-decoder. These two sub-decoders are composed of node-transposed (1D) and edge-transposed (2D) convolution layers to decode node and edge representations and generate the features for nodes and edges simultaneously. The node-decoder is responsible for reconstructing the node features, taking the latent representation provided by the encoder and translating it back into the node-specific attributes of the graph. At the same time, the edge-decoder focuses on the reconstruction of edge features, converting the latent representation provided by the encoder into meaningful edge attributes of the graph. Concretely, the node decoder, denoted as $f'_{n}\left((X''|z_{n}\right)$, accepts a latent vector $z_{n}\sim p\left( z\right)$ and produces the reconstructed/generated node feature matrix $X''\in \mathbb{R} ^{n\times d}$, while the edge decoder, denoted as $f'_{e}\left((A^{e''}|z_{e}\right)$, accepts a latent vector $z_{e}\sim p\left( z\right)$ and outputs the reconstructed/generated edge feature matrix $A^{e''}\in \mathbb{R} ^{n\times n \times c}$. The latent code vector $z$ can be either provided by the previous disentanglement module or sampled from the same learned stochastic distribution $p\left( z\right)$. Specifically, a predefined maximum number of nodes needs to be set in advance for the decoder to concurrently output two continuous and dense matrices, $A^{e,\pi''} \in \mathbb{R} ^{n\times n\times c}$ and $\widehat{X}^{\pi''} \in \mathbb{R} ^{n\times d}$ with a particular ordering $\pi$, which define the edge and node attributes of the reconstructed/generated graph $G''$. 

Internally, both the node decoder and the edge decoder are composed of a non-linear 8-layer Multi-layer Perceptron (MLP) mapping network $f'_{1}$ and a synthesis network $f'_{2}$. Each sub-decoder intakes the latent vector $z \in \mathbb{R} ^M$ of input latent space $Z$ with dimension $M$ and processes it through the mapping network $f'_{1}: Z \rightarrow W$ to obtain a transformed latent vector $w \in \mathbb{R} ^M$ of another intermediate latent space $W$ of the same dimensionality. This structural configuration adheres to the established convention outlined in StyleGAN \cite{karras2019style, karras2020analyzing}. The transformed latent vector $w$ is further processed with a learned affine transformation using a fully connected layer \cite{karras2019style, karras2020analyzing}. The processed vector $w$ with varied affinement then serves as synthesis control factors by being fed to different convolution layers. Specifically, at each synthesis layer $l$ of the synthesis network $f'_{2}$, we have:
\begin{equation}
\label{eq_encoder_style1}
\widehat{x}_{l} = x_{l} \odot \widehat w_{l}
\end{equation}
\begin{equation}
\label{eq_encoder_style2}
where, \widehat w_{l} = Affine\left(w\right)
\end{equation}
and for the node decoder, we have:
\begin{equation}
\label{eq_encoder_style3}
X_{l+1} = Conv1D\left(\widehat{X}_{l}\right)
\end{equation}
where $X_{l}, \widehat{X}_{l}, X_{l+1} \in \mathbb{R} ^{c \times n_{l}}$, with $n_{l}$ being the 'resolution' (i.e., number of 'super nodes') of the synthesized graph node feature matrix at layer $l$, and $conv1d$ refers to a 1-dimensional convolution layer. As for the edge decoder, we have:
\begin{equation}
\label{eq_encoder_style4}
A_{l+1} = Conv2D\left(\widehat{A}_{l}\right)
\end{equation}
where $A_{l}, \widehat{A}_{l}, A_{l+1} \in \mathbb{R} ^{d \times n_{l} \times n_{l}}$, where $conv2d$ refers to a 2-dimensional convolution layer.

\section{Model implementation and training}

This section explains the details of our adopted training experiment implementation and training datasets.

\subsection{Training datasets}

\begin{table}
\caption{Two sets of graph datasets with different numbers of architectural element categories}
\centering
% \fontsize{8}{8}\selectfont
\begin{tabular}{p{1cm}|p{1.5cm}|p{4.5cm}}
\hline\hline
&  \multicolumn{2}{c}{\textbf{Number of categories}}\\
\hline
&  \textbf{6}& \textbf{25}\\
\hline
\multirow{2}{4em}{\textbf{Node labels}}&  outdoor, room, stair& outdoor, room, stair, corridor, elevator, escalator, facilities, furniture, greenery, ladder, lavatory, parking, pillar, pool, terrace, skylight, slope, steps, void, \\
&  & \\
\hline
\multirow{2}{4em}{\textbf{Edge labels}}&  wall, door, window& wall, door, window, cased opening, fence, movable partition\\
&  & \\
\hline\hline
\end{tabular}
\label{table:labels_6_25}
\end{table}

\begin{figure}[!t]
  \centering
  \includegraphics[width=1.0\linewidth]{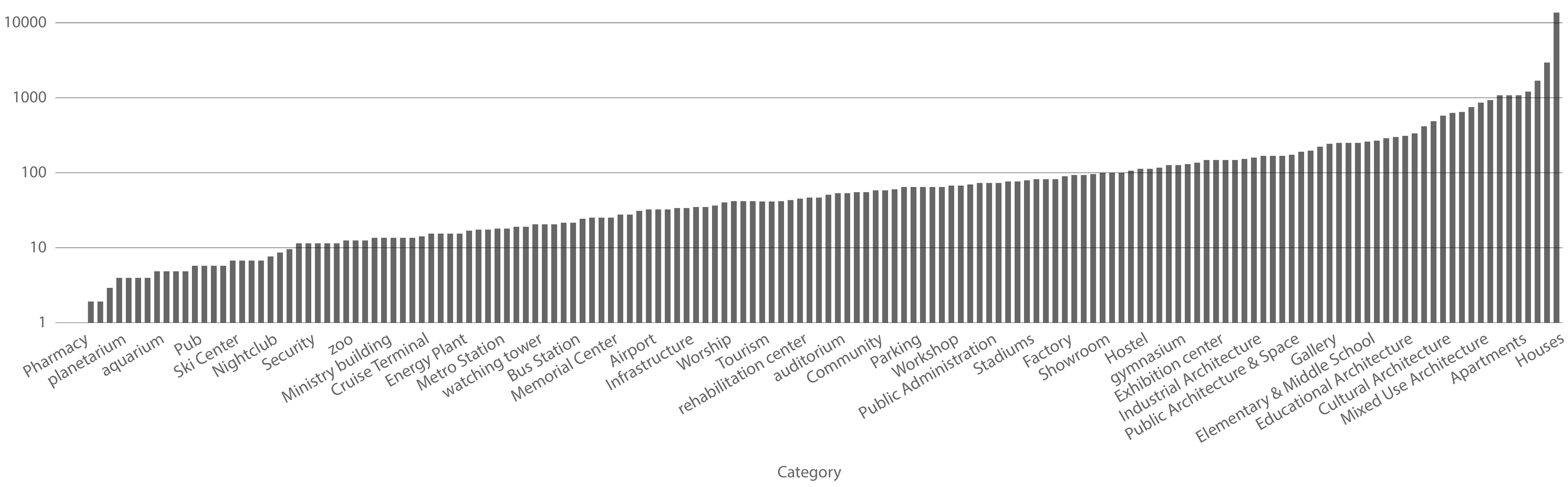}
  % \vspace{\BeforeCaptionVSpace}
  \caption{Distribution of attributed adjacency graphs per architectural category (159 categories in total). The vertical axis is scaled according to the logarithm of image numbers}
  \label{study2:fig:FPimgesDistribution}
\end{figure}

Access to large-scale layout graph databases with quasi-exhaustive coverage and enough instances to cover the diversity of real-world architecture layout design data space is essential for examining the performance of the proposed disentanglement graph representation learning framework. However, large-scale attributed adjacency graph datasets of real-world architectural design are scant in the current literature. Although the graph extraction outputs can be more accurate using 3D building models, the accessibility of real-world 3D architectural design models in bulk is difficult. Given that traditional orthographic drawings of existing architectural design, such as floor plans, are fairly easy to acquire from the internet, it is possible to acquire attributed adjacency graphs from real-world architectural floor plan images and construct large-scale architectural layout design graph datasets. In this study, we harness the floor plan image parsing methods provided in \cite{chen2022AAG-FP} and \cite{chen4727643floorplanseg} to extract attributed adjacency graphs from a customized repository of real-world architectural floor plan images of 159 architectural categories retrieved from \href{https://www.archdaily.com}{ArchDaily®}, a professional architectural design project website. The distribution of extracted layout graphs per architectural category is illustrated in Fig.~\ref{study2:fig:FPimgesDistribution}.  Specifically, we have curated two graph datasets with different numbers of architectural element categories (Table \ref{table:labels_6_25}): the first dataset is constructed using the ensemble-based supervised floor plan parsing schemes provided by \cite{chen2022AAG-FP} with 6 architectural element categories, while the second one is constructed using the semi-weakly-supervised scheme offered in \cite{chen4727643floorplanseg} with 25 architectural element categories. 

We define a concatenated set of baseline node features composed of three major components. 1) Node class represents the categorical classification of each node (as shown in Table \ref{table:labels_6_25}), providing essential information about the type or function of the corresponding element in the architectural layout graph; the class labels are transformed into one-hot encoding. 2) The space area ratio quantifies the proportion of the area occupied by the corresponding original polygon of the node relative to the total layout area. 3) Normalized coordinates of the original space polygon centre; the normalization is to ensure consistency and comparability across different graphs, offering valuable spatial information about the positioning of the spaces within the overall architectural layout. Regarding edge labels, the two graph datasets with different numbers of element categories also provide varied sets of edge labels, as shown in Table \ref{table:labels_6_25}. We transform the edge labels into one-hot encoding for training purposes, denoting the connections between two neighbouring nodes. This feature is instrumental in capturing the architectural layout graph's relationships and interactions among various spaces. These node and edge features jointly offer a rich and detailed representation of architectural layout data, enabling the proposed disentangled graph representation learning framework to learn and interpret architectural layout designs' complex and nuanced characteristics.

With these tailored datasets, we embark on experiments to extract high-level structural features from the attributed adjacency graphs and probe into the potential for interpreting and navigating the latent design representation space that the architectural graphs may reveal using the disentangled graph representation learning frameworks proposed in this study. Please refer to Appendix \ref{AAG_app} for the visualization of the graph extraction process, samples of attributed adjacency graphs and other relevant details.

\subsection{Feature augmentation}

We experiment with a series of attributed adjacency multi-graph feature augmentation schemes and examine their impact on model performance.

\subsubsection{Augmentation with canonical ordering}
\label{position_aug_canonical}

Ideally, we seek to maximize the likelihood of $p\left( G\right) =\sum _{\pi }P\left( G,\pi \right)$. However, this computation is infeasible as the total number of node orderings is factorial in the number of nodes ($n!$). We deal with this issue by applying a selected set of canonical orderings $K = \left\{ \pi _{1},\ldots,\pi _{k}\right\}$ to the input graphs, chosen based on certain criteria: each ordering in the set leads to a unique permutation, ensuring that no two orderings encode the same structural information redundantly, and the selected orderings should collectively capture the essential variations in graph structures. While this approach simplifies the issue, it learns an approximated lower bound (ELBO) of the true likelihood $\sum _{\pi }P\left( G,\pi \right)$, namely $\sum _{\pi \in K}P\left( G,\pi \right)$. The quality of this approximation depends on how well the chosen set of canonical orderings $K$ can represent the space of all orderings. Specifically, we chose the following canonical node orderings anchored in graph properties: 1) arranging nodes in descending order based on node degree, 2) sorting based on average neighbor degree, 3) sequencing by closeness centrality and 4) betweenness centrality. Concretely, we have $p\left( G\right) \geq \sum _{\pi \in K}P\left( G,\pi \right) > \forall P\left( G,\pi \right)$. By maximizing $log \sum _{\pi \in K}P\left( G,\pi \right)$ for a given graph $G$, we implicitly select the optimal combination of node orderings from the set $K$ and maximize observing $G$ under the learned distribution using this optimal ordering \cite{liao2019efficient}.

\subsubsection{Augmentation with positional encoding}
\label{position_aug_positional}

Conventional graph message-passing approaches are usually unaware of the nodes’ different structural roles, as all nodes are treated equally when performing local operations. Despite the initial intuition that neural networks would be able to discover these roles by constructing deeper model structures, it has been shown that vanilla graph neural networks are ill-suited for this purpose and are blind to the existence of structural properties \cite{chen2020substructures}. Thus, positional encoding can play a pivotal role in graph representation learning by embedding global positional information within individual nodes. This feature is essential for distinguishing isomorphic nodes and edges, enhancing the model's capacity to capture and represent complex graph structures and relationships. Hussain et al. \cite{hussain2022global} proposed a unique form of positional encoding that leverages the graph adjacency matrices' pre-calculated Singular Value Decomposition (SVD). It uses the largest \textit{k} singular values and their corresponding left and right singular vectors to construct the positional encoding. This positional encoding approach offers a robust mechanism for encoding positional information across diverse graphs, serving as an absolute global coordinate system. This study explores the SVD-based node positional encoding scheme as an additional feature augmentation. Concretely, given the adjacency matrix $A\in \mathbb{R} ^{n\times n}$ of an input attributed adjacency multi-graph $G$, we have $A \dfrac{SVD}{}U\cdot S\cdot V^{H}$, with $U\in \mathbb{R} ^{n\times n}$ and $V^{H} \in \mathbb{R} ^{n\times n}$ being 2D unitary matrices and $H$ refers to the Hermitian transpose; the rows of $V^{H}$ are the eigenvectors of $A^{H}A$ while the columns of $U$ are the eigenvectors of $AA^{H}$. $S\in \mathbb{R} ^{n\times n}$ is a diagonal matrix with the principal diagonal being $A$’s singular values sorted in descending order; the principal diagonal of $S$ forms the 1D vector $s \in \mathbb{R} ^{n}$, which contains the singular values of $A$. The SVD-based node positional encoding is then calculated as,
\begin{equation}
\label{eq_encoder_svd}
\widehat{\Gamma} = FFN\left(\left(U || {V^{H}}^{T} \right) \odot \sqrt{s}\right), \widehat{\Gamma} \in \mathbb{R} ^{n\times d}
\end{equation}
with $||$ being the concatenation operation along the columns, and the feed-forward network layer being used for learnable projection before integrating the positional encoding into node features. This heuristic approach has been shown to yield improved results \cite{hussain2022global}. Incorporating the SVD positional encoding, our goal is to enrich the node features with nuanced structural information regarding their relationships within the attributed adjacency multi-graphs. This enhancement aims to bolster the framework's expressiveness to capture positional-sensitive layout design patterns effectively, thereby improving its overall performance in disentangling and interpreting the architectural layout design graph data space. 

\subsubsection{Augmentation with extra polygon vertices information}
\label{polygon_aug}

We also try to augment the node features by integrating supplementary information on the coordinates of polygon vertices. This entails incorporating normalized coordinates of polygon boundary vertices as extra node features. The rationale behind this augmentation is to enrich the node representation with finer geometric details, which may potentially improve the model's capacity to comprehend and depict the complex nuances inherent in architectural layout designs. By including this additional information, we further explore the model's capability to capture and articulate the intricate architectural features embedded within the layout design graphs.

\subsection{Training implementation}

We experiment with various training implementation schemes and disentanglement module variations of the proposed framework to explore how varied training schemes and latent space disentanglement modules can influence the model's ability to learn and represent the complexities of architectural layout design graph data and assess their impact on the model's performance regarding representation disentanglement.

\subsubsection{Dimensionality of intermediate latent space}

The latent space in graph representation learning serves as a compressed input data representation, capturing the essential features and patterns in a lower-dimensional form. The choice of dimension for this latent space can be a balance between complexity and expressiveness. A higher-dimensional latent space can potentially capture more detailed and nuanced information about the graph, leading to richer and more accurate representations. However, it can also introduce challenges such as increased computational complexity. Conversely, a lower-dimensional latent space may be computationally more efficient. Still, it might not capture the full complexity of the data, potentially leading to less accurate representations. 

We experiment with the dimensionality of the intermediate latent space $\textbf{Z} \in \mathbb{R} ^M$ and $\textbf{W} \in \mathbb{R} ^M$, and examine whether the latent space dimension $M$ can significantly influence the model performance. Our experimentation will involve varying the dimensions of the latent spaces and observing the resultant effects on the model's performance. Key performance metrics measuring fidelity and diversity will be assessed across different latent space dimensions. This will enable us to determine the optimal size of the latent space that balances expressiveness and computational efficiency while maximizing the performance of the disentangled graph representation learning model.

\subsubsection{Number of architectural element label categories}

we also take into account the number of architectural element label categories utilized during training. This is achieved using the two meticulously curated training graph datasets, each featuring a distinct number of architectural element categories. One dataset encompasses 6 architectural element categories, while the other comprises 25 architectural element categories. Our objective is to investigate whether incorporating an increased number of architectural element label categories provides more intricate and pertinent information for graph representation, potentially influencing the model's performance. Through this evaluation, we aim to discern the impact of varying architectural element label categories on the efficacy of the model and its ability to capture the nuances of architectural layout designs accurately.

\subsubsection{Latent space disentanglement module}

We experiment with the proposed different disentanglement modules, including the vector quantization scheme, the node-edge co-disentanglement scheme, and the layer-wise stochasticity mechanism incorporated in the style-based decoder. Both the vector quantization scheme and the node-edge co-disentanglement scheme are evaluated against the baseline vanilla VAE scheme. To assess the performance of the style-based decoder, we compare it against a vanilla MLP decoder comprising 2 fully connected layers, which also includes two sub-decoders. The MLP node decoder performs the feature transformation as $X'' = l_n\left(pr(\left(l^{nn}_{i\rightarrow n*d}\left(l_n\left(pr(\left(\left(l^{nn}_{M\rightarrow M}\left(z\right)\right)\right)\right)\right)\right)\right)$ and the MLP edge decoder has $A^{e''} = l_n\left(pr(\left(l^{nn}_{i\rightarrow n^{2}*d}\left(l_n\left(pr(\left(\left(l^{nn}_{M\rightarrow M}\left(z\right)\right)\right)\right)\right)\right)\right)$. Each layer in the MLP stack is immediately followed by a PReLU activation layer \cite{he2015prelu} and a layer normalization layer \cite{ba2016layernorm}. Through these experiments, we aim to analyze the effectiveness of different disentanglement modules and decoder architectures in capturing and representing the latent generative factors of architectural layout design graphs.

\subsection{Losses}

The design of appropriate loss functions is essential to ensure that the representation remains disentangled while retaining the information inherent in the data \cite{wang2022disentangled}. 
The components of the loss function $L$ used in this study can be categorised into two parts based on their distinct purposes: reconstruction loss and disentanglement loss. The \textbf{reconstruction loss} is crucial in generation tasks for preserving data integrity by encouraging the accurate reconstruction of the original data, which ensures that learned disentangled representations are semantically meaningful. The \textbf{disentanglement loss} is specifically designed to enforce the separation of the representation, ensuring that each part of the disentangled representation corresponds to unique and independent aspects of the data. These two loss function parts work together to ensure a harmonious balance between maintaining the quality and integrity of the data and effectively achieving disentanglement.

Specifically, we have two respective reconstruction losses for the reconstruction of the node feature matrix and the edge feature matrix. For the node feature matrix reconstruction, we have:
\begin{equation}
\label{loss_r_node}
L_{node} = -\sum_{v=1}^{n} \sum_{i=1}^{d}  [\overline{x_{v}^{i}}\log(x_{v}^{i}) + (1 - \overline{x_{v}^{i}})\log(1 - x_{v}^{i})]
\end{equation}
where $\overline{x_{v}^{i}}$ is the reconstructed feature value at dimension $i$ of node $v$ and $x_{v}^{i}$ is the corresponding ground truth value. Similarly, for the edge feature matrix reconstruction, we have:
\begin{equation}
\label{loss_r_edge}
L_{edge} = -\sum_{u=1}^{n} \sum_{v=1}^{n} \sum_{i=1}^{c}  [\overline{a_{u,v}^{i}}\log(a_{u,v}^{i}) + (1 - \overline{a_{u,v}^{i}})\log(1 - a_{u,v}^{i})]
\end{equation}
where $\overline{a_{u,v}^{i}}$ is the reconstructed feature value at dimension $i$ of edge $(u, v)$ and $a_{u,v}^{i}$ is the corresponding ground truth value. Consequently, we have the total reconstruction loss $L_{rec}$ as:
\begin{equation}
\label{loss_r}
L_{rec} = L_{node} + L_{edge}
\end{equation}

Meanwhile, as the proposed frameworks are VAE-based, we adopt Kullback-Leibler (KL) divergence to optimize the disentanglement latent space by quantifying the distance between the estimated posterior distributions and the isotropic Gaussian prior, and we optimize the KL divergence loss $L_{KL}$ to minimize the estimation losses. Specifically, for the baseline framework with the vanilla VAE module, we have
% https://stats.stackexchange.com/questions/7440/kl-divergence-between-two-univariate-gaussians
\begin{align}
\label{KL_loss}
L_{KL} &= D_{KL}(log q_{\theta }\left( z|(X, A^{e})\right) \left| \right| p\left( z\right)) \\
 &= -\frac{1}{2M} \sum_{m=1}^{M} (1 + z_{\sigma}^{m} - (z_{\mu}^{m})^2 - e^{2z_{\sigma}^{m}})
\end{align}
where $p\left( z\right)$ is the isotropic Gaussian prior $\mathcal{N} \left( 0,\textbf{I}\right)$ and $q_{\theta }\left( z|(X, A^{e})\right)$ is the estimated posterior distribution, $z_{\mu}^{m}$ and $z_{\sigma}^{m}$ are respectively the estimated mean value and log-variance at latent dimension $m$. While for the framework with the NED-based disentanglement module, we have
% https://stats.stackexchange.com/questions/7440/kl-divergence-between-two-univariate-gaussians
\begin{align}
\label{KL_NED_loss}
L_{KL}^{NED} &= D_{KL}(log q_{\theta }\left( z_{graph}|(X, A^{e})\right) \left| \right| p\left( z\right)) \\
&+ D_{KL}(log q_{\theta }\left( z_{node}|(X, A^{e})\right) \left| \right| p\left( z\right)) \\
&+ D_{KL}(log q_{\theta }\left( z_{edge}|(X, A^{e})\right) \left| \right| p\left( z\right)) \\
&= -\frac{1}{2M} \cdot \\
&(\sum_{m=1}^{M} (1 + z_{\sigma}^{graph, m} - (z_{\mu}^{graph, m})^2 - e^{2z_{\sigma}^{graph, m}}) \\
&+ \sum_{m=1}^{M} (1 + z_{\sigma}^{node, m} - (z_{\mu}^{node, m})^2 - e^{2z_{\sigma}^{node, m}}) \\
&+ \sum_{m=1}^{M} (1 + z_{\sigma}^{edge, m} - (z_{\mu}^{edge, m})^2 - e^{2z_{\sigma}^{edge, m}}))
\end{align}

For the Vector Quantization module, the distance between the embedding vectors $k$ and the latent codes $z$ is optimized using Mean Squared Error (MSE). Specifically, the VQ loss $L_{VQ}$ is composed of two parts: the dictionary learning component and the commitment loss component. The former aligns the embedding vectors $k$ towards the latent codes $z$, thereby refining the quantization dictionary. The latter guarantees that the latent codes $z$ reliably correspond to specific embeddings $k$ within the quantization dictionary, preventing unchecked dimensionality expansion of the intermediate embedding space $K$. Concretely, we have:
\begin{equation}
\label{VQ_loss}
L_{VQ} = \sum_{i=1}^{K}(const(z_{k}^{i})-k^{i})^2 + \sum_{i=1}^{K}(z_{k}^{i}-const(k^{i}))^2
\end{equation}
where $const$ denotes the operation of detaching from the computational graph and renders the variable as a constant during optimization. The first term of the RHS of equation \ref{VQ_loss} refers to the dictionary learning loss, and the second term is the commitment loss. When both the Vector Quantization module and the Node-edge co-disentanglement module are implemented, we have:
\begin{align}
\label{VQ_NED_loss}
L_{VQ}^{NED} &= \sum_{i=1}^{K}(const(z_{k}^{node+graph, i})-k^{node+graph, i})^2\\
&+ \sum_{i=1}^{K}(z_{k}^{node+graph, i}-const(k^{node+graph, i}))^2\\
&+ \sum_{i=1}^{K}(const(z_{k}^{edge+graph, i})-k^{edge+graph, i})^2\\
&+ \sum_{i=1}^{K}(z_{k}^{edge+graph, i}-const(k^{edge+graph, i}))^2
\end{align}
which incorporates the vector quantization of $z^{node+graph}$ and $z^{edge+graph}$, namely the node and edge-level latent codes fused with the graph-level latent code.

For different framework implementation schemes, we apply different combinations of the loss function components; details are demonstrated in Table \ref{table:losses}.

\begin{table}
\caption{Different combinations of loss function components for various framework implementation schemes. "EA-encoder" stands for Edge-augmented encoder, "VAE", "VQ", and "NED" refer to the different disentanglement modules introduced in this study, "Style-decoder" indicates the style-based decoder, and "MLP-decoder" refers to the vanilla MLP decoder for comparison}
\centering
% \fontsize{8}{8}\selectfont
\begin{tabular}{p{5cm}|p{3cm}}
\hline\hline
\textbf{Framework scheme}& \textbf{Objective function}\\
\hline
EA-encoder + VAE + Style-decoder& $L_{rec} + L_{KL}$\\
\hline
EA-encoder + VQ + Style-decoder& $L_{rec} + L_{KL} + L_{VQ}$\\
\hline
EA-encoder + NED + Style-decoder& $L_{rec} + L_{KL}^{NED}$\\
\hline
EA-encoder + VQ + NED + Style-decoder& $L_{rec} + L_{KL}^{NED} + L_{VQ}$\\
\hline
EA-encoder + VAE + MLP-decoder& $L_{rec} + L_{KL}$\\
\hline
EA-encoder + VQ + MLP-decoder& $L_{rec} + L_{KL} + L_{VQ}$\\
\hline
EA-encoder + NED + MLP-decoder& $L_{rec} + L_{KL}^{NED}$\\
\hline
EA-encoder + VQ + NED + MLP-decoder& $L_{rec} + L_{KL}^{NED} + L_{VQ}$\\
\hline\hline
\end{tabular}
\label{table:losses}
\end{table}

\section{Empirical Experiments}

This section introduces our experimental setup and discusses our experiments' quantitative and qualitative evaluation results. 

\subsection{Experimental setup}

% https://help.nscc.sg/softwarehardware-information/
All experiments of this study are performed with PyTorch \cite{paszke2019pytorch} using the following system setup:
\begin{itemize}
    \item Operating system: Red Hat Enterprise Linux release 8.4 (Ootpa).
    \item CPU: AMD EPYC 7713P 64-Core Processor.
    \item GPU: 1x NVIDIA A100-SXM4-40GB.
    \item RAM: 500GB DDR4 ECC RAM
\end{itemize}

We initialize the model weights using the normal initialization method. The edge-augmented encoder comprises a total of 8.48 million trainable parameters. The vanilla VAE disentanglement module consists of 1.32 million trainable parameters, while the VQ-based disentanglement module and the NED-based disentanglement module have 1.58 million and 12.34 million trainable parameters, respectively. The style-based node decoder contains 14.04 million trainable parameters, whereas the style-based edge decoder has 29.34 million trainable parameters. The vanilla MLP node decoder comprises 1.71 million trainable parameters, while the MLP edge decoder comprises 50.89 million. We set the maximum number of nodes ($n$) to 128 in our experiments, a value calculated based on the characteristics of the training datasets.

\subsection{Model Performance}

This section demonstrates the results of our experimental endeavours, employing both quantitative and qualitative methods to offer an exhaustive assessment of the efficacy of our proposed frameworks.

\subsubsection{Quantitative evaluation}
\label{results:gnn:quant}

For quantitative evaluation of the performance of the proposed frameworks, we adopt a series of domain-agnostic, scalable and expressive evaluation metrics recommended by Thompson et al. \cite{thompson2022evaluation}, tailored for easy and accurate evaluating and ranking of graph generative models. Specifically, \textbf{Frechet Distance} (FD, or FID)\cite{heusel2017gans} approximates the graph embeddings as continuous multivariate Gaussians with sample mean and covariance and the distance between distributions is computed as an approximate measure of the sample qualities. \textbf{Precision \& Recall} (P\&R) \cite{kynkaanniemi2019improved} decouples a generator's quality into two distinct values to detect mode collapse and mode dropping, constructing manifolds by extending a radius from each sample in a set to its $k^th$ nearest neighbour to form hyperspheres. The union of these hyperspheres represents a manifold: precision measures the percentage of generated samples within the real samples' manifold, while recall measures the percentage of real samples within the generated samples' manifold. The harmonic mean (``F1 PR'') of P\&R, a scalar metric, can further provide meaningful decomposable values in experiments \cite{lucic2018gans}. \textbf{Density \& Coverage} (D\&C) \cite{naeem2020reliable}, developed as robust alternatives to P\&R, differ by creating a single manifold from the union of all hyperspheres for each set and treating each sample's hypersphere independently. Density is calculated based on the number of real hyperspheres that a generated sample falls within on average, while coverage, on the other hand, is the percentage of real hyperspheres containing at least one generated sample. These hyperspheres are determined using the $k^th$ nearest neighbour method, similar to P\&R, and just like with P\&R, a scalar metric can be formed using the harmonic mean (``F1 DC'') of D\&C for a comprehensive evaluation \cite{lucic2018gans}. \textbf{Maximum Mean Discrepancy} (MMD) \cite{gretton2006kernel} is a versatile measure used to quantify the dissimilarity between two sets of graphs regardless of the domain, utilizing different kernel functions. The original Kernel Inception Distance (KID) applied a polynomial kernel in conjunction with MMD, while MMD Linear employs a parameter-free linear kernel, offering a simpler approach, and the RBF kernel (MMD RBF) is also widely utilized for its effectiveness in capturing dissimilarities \cite{xu2018empirical}. According to Thompson et al.\cite{thompson2022evaluation}, \textit{recall}, \textit{coverage} and \textit{F1 PR} exhibit strong positive correlations with the diversity level, while \textit{precision} and \textit{density} are negatively correlated with diversity. Meanwhile, \textit{MMD RBF} has slightly stronger correlations with both fidelity and diversity of generated samples compared to other metrics. In addition, MMD RBF and F1 PR are both capable of detecting changes in node and edge feature distributions.

We present the t-test statistic alongside the corresponding p-value to demonstrate the significance of the differences between the means of each of the two comparison groups (Table \ref{table:FDMMD} and Table \ref{table:PRDC}); larger magnitudes (absolute values) of the t-statistic suggest a higher likelihood that the observed difference between the group means is not a product of random chance. This statistical method provides a rigorous approach to assessing the significance of the differences observed in our data, ensuring our conclusions are robust and reliable.

\begin{table*}[h!]
\caption{Report of t-test statistic of each evaluation metric, calculated based on all training epochs (Style: apply style-based decoder or the MLP-based counterpart; SVD: involve singular value decomposition-based positional encoding or not; NED: involve node-edge co-disentanglement or not; VQ: involve vector quantization or not; poly: involve polygon vertices information or not; label: number of architectural element label categories involved; z-dim: dimensions of latent codes; FD: lower is better; F1 PR: higher is better; F1 DC: higher is better; MMD Linear: lower is better; MMD RBF: lower is better.)}
\centering
\tiny
\fontsize{8}{8}\selectfont
\begin{tabular}{p{0.6cm}|p{0.6cm}|p{0.8cm}|p{1.5cm}|p{0.6cm}|p{1.5cm}|p{0.6cm}|p{1.5cm}|p{0.6cm}|p{1.5cm}|p{0.6cm}|p{1.5cm}} 
\hline\hline
\multicolumn{2}{c|}{ } & \multicolumn{2}{c|}{FD}&  \multicolumn{2}{c|}{F1 PR}&  \multicolumn{2}{c|}{F1 DC}& \multicolumn{2}{c|}{MMD Linear}&\multicolumn{2}{c}{MMD RBF}\\
\hline
\multicolumn{2}{c|}{\textbf{Module}} & mean& t-statistic& mean& t-statistic&  mean&t-statistic&  mean&t-statistic& mean&t-statistic\\
\hline
\multirow{2}{4em}{\textbf{Style}}&True&86.97&\multirow{2}{4em}{-16;p=.00}&.004&\multirow{2}{4em}{-13;p=.00}&.04&\multirow{2}{4em}{-43;p=.00}&18.58&\multirow{2}{4em}{-10;p=.00}&.88&\multirow{2}{4em}{58.3;p=.00}\\
&False&138.51&&.02&&.14&&30.93&&.53&\\ 
\hline
\multirow{2}{4em}{\textbf{SVD}}&True&114.24&\multirow{2}{4em}{2.96;p=.00}&.01&\multirow{2}{4em}{1.62;p=.11}&.09&\multirow{2}{4em}{5.81;p=.00}&23.67&\multirow{2}{4em}{1.09;p=.28}&.69&\multirow{2}{4em}{-3;p=.00}\\
&False&106.42&&.01&&.08&&22.61&&.72&\\ 
\hline 
\multirow{2}{4em}{\textbf{NED}}&True&112.87&\multirow{2}{4em}{1.93;p=.05}&.01&\multirow{2}{4em}{-1;p=.28}&.08&\multirow{2}{4em}{-6;p=.00}&23.13&\multirow{2}{4em}{.03;p=.98}&.73&\multirow{2}{4em}{7.72;p=.00}\\
&False&107.76&&.01&&.09&&23.10&&.68&\\ 
\hline 
\multirow{2}{4em}{\textbf{VQ}}&True&96.80&\multirow{2}{4em}{-10;p=.00}&.004&\multirow{2}{4em}{-14;p=.00}&.05&\multirow{2}{4em}{-37;p=.00}&22.57&\multirow{2}{4em}{-1;p=.3}&.81&\multirow{2}{4em}{30.18;p=.00}\\
&False&122.89&&.022&&.13&&23.64&&.61&\\ 
\hline 
\multirow{2}{4em}{\textbf{poly}}&True&105.12&\multirow{2}{4em}{-5;p=.00}&.01&\multirow{2}{4em}{-10;p=.00}&.10&\multirow{2}{4em}{17.45;p=.00}&22.87&\multirow{2}{4em}{-7;p=.51}&.74&\multirow{2}{4em}{14.89;p=.00}\\
&False&118.76&&.02&&.06&&23.53&&.64&\\ 
\hline 
\multirow{2}{4em}{\textbf{label}}&6&103.07&\multirow{2}{4em}{-7;p=.00}&.01&\multirow{2}{4em}{-10;p=.00}&.09&\multirow{2}{4em}{7.38;p=.00}&20.09&\multirow{2}{4em}{-8;p=.00}&.68&\multirow{2}{4em}{-9;p=.00}\\
& 25&  122.47&&  .02&&  .07&&  28.37&& .75&\\ 
\hline 
\multirow{2}{4em}{\textbf{z-dim}}&512&110.87&\multirow{2}{4em}{.80;p=.42}&.015&\multirow{2}{4em}{2.45;p=.01}&.08&\multirow{2}{4em}{-11;p=.00}&22.51&\multirow{2}{4em}{-2;p=.06}&.69&\multirow{2}{4em}{-6;p=.00}\\
&1024&108.57&&.011&&.11&&24.49&&.74&\\ 
\hline\hline
\end{tabular}
\label{table:FDMMD}
\end{table*}

\begin{table*}[h!]
\caption{Report of t-test statistic of each evaluation metric, calculated based on all training epochs (Style: apply style-based decoder or the MLP-based counterpart; SVD: involve singular value decomposition-based positional encoding or not; NED: involve node-edge co-disentanglement or not; VQ: involve vector quantization or not; poly: involve polygon vertices information or not; label: number of architectural element label categories involved; z-dim: dimensions of latent codes; Precision: higher is better; Recall: higher is better; Density: higher is better; Coverage: higher is better.)}
\centering
\tiny
\fontsize{8}{8}\selectfont
\begin{tabular}{p{0.6cm}|p{0.6cm}|p{0.6cm}|p{1.7cm}|p{0.6cm}|p{1.7cm}|p{0.6cm}|p{1.7cm}|p{0.6cm}|p{1.7cm}} 
\hline\hline
\multicolumn{2}{c|}{ } & \multicolumn{2}{c|}{Precision}&  \multicolumn{2}{c|}{Recall}&  \multicolumn{2}{c|}{Density}& \multicolumn{2}{c}{Coverage}\\
\hline
\multicolumn{2}{c|}{\textbf{Module}} & mean& t-statistic& mean& t-statistic&  mean&t-statistic&  mean&t-statistic\\
\hline
\multirow{2}{4em}{\textbf{Style}}&True&.98&\multirow{2}{4em}{9.78;p=.00}&.003&\multirow{2}{4em}{-9.6;p=.00}&1.87&\multirow{2}{4em}{9.04;p=.00}&.02&\multirow{2}{4em}{-41;p=.00}\\
&False&.96&&.01&&1.60&&.08&\\ 
\hline 
\multirow{2}{4em}{\textbf{SVD}}&True&.97&\multirow{2}{4em}{-.1;p=.90}&.01&\multirow{2}{4em}{.31;p=.75}&1.76&\multirow{2}{4em}{1.04;p=.30}&.05&\multirow{2}{4em}{5.97;p=.00}\\
&False&.97&&.01&&1.73&&.04&\\ 
\hline 
\multirow{2}{4em}{\textbf{NED}}&True&.97&\multirow{2}{4em}{.75;p=.45}&.01&\multirow{2}{4em}{.03;p=.98}&1.73&\multirow{2}{4em}{-1;p=.33}&.04&\multirow{2}{4em}{-6;p=.00}\\
&False&.97&&.01&&1.76&&.05&\\ 
\hline 
\multirow{2}{4em}{\textbf{VQ}}&True&.973&\multirow{2}{4em}{2.96;p=.00}&.003&\multirow{2}{4em}{-11;p=.00}&1.70&\multirow{2}{4em}{-4;p=.00}&.02&\multirow{2}{4em}{-36;p=.00}\\
&False&.966&&.013&&1.79&&.07&\\ 
\hline 
\multirow{2}{4em}{\textbf{poly}}&True&.99&\multirow{2}{4em}{25.80;p=.00}&.005&\multirow{2}{4em}{-8;p=.00}&2.30&\multirow{2}{4em}{80.67;p=.00}&.06&\multirow{2}{4em}{13.86;p=.00}\\
&False&.93&&.013&&.81&&.04&\\ 
\hline 
\multirow{2}{4em}{\textbf{label}}&6&.968&\multirow{2}{4em}{-2;p=.04}&.005&\multirow{2}{4em}{-9;p=.00}&1.87&\multirow{2}{4em}{13.66;p=.00}&.05&\multirow{2}{4em}{4.20;p=.00}\\
&25&.973&&.014&&1.53&&.04&\\ 
\hline 
\multirow{2}{4em}{\textbf{z-dim}}&512&.96&\multirow{2}{4em}{-12;p=.00}&.009&\multirow{2}{4em}{2.16;p=.03}&1.53&\multirow{2}{4em}{-28;p=.00}&.04&\multirow{2}{4em}{-10;p=.00}\\
&1024&.99&&.007&&2.24&&.06&\\ 
\hline\hline
\end{tabular}
\label{table:PRDC}
\end{table*}

The results highlight several key findings. Incorporating a style-based decoder significantly enhances fidelity, as evidenced by lower ``FD'' and ``MMD Linear'' values and higher ``Precision'' and ``Density'' values, without affecting diversity. Adding SVD positional encoding improves both fidelity and diversity, indicated by lower ``MMD RBF'' and higher ``Coverage'' values. Including polygon vertices coordinates enriches node features, boosting fidelity (lower ``FD'', higher ``F1 DC'', ``Precision'', ``Density'', ``Coverage'') but reducing diversity (higher ``MMD RBF'', lower ``F1 PR'', ``Recall''). Increasing architectural element label categories also enhances fidelity (higher ``F1 PR'', ``Precision'', ``Recall'') at the expense of diversity (higher ``MMD Linear'', ``MMD RBF'', lower ``F1 DC'', ``Density'', ``Coverage''). Implementing vector quantization improves fidelity (lower ``FD'', higher ``Precision'') but reduces diversity (higher ``MMD RBF'', lower ``Recall'', ``Density'', ``Coverage''). Increasing latent code dimensions enhances diversity (higher ``F1 DC'', ``Precision'', ``Density'', ``Coverage'') but slightly reduces fidelity (higher ``MMD RBF''). Incorporating node-edge co-disentanglement significantly increases diversity (higher ``F1 DC'', ``Coverage'') but slightly impacts fidelity (higher ``MMD RBF''). These findings underscore the delicate balance between fidelity and diversity in graph representation learning tasks for architectural layout graphs. More detailed documentation of the quantitative evaluation results can be found in Appendix \ref{quan_app}.

\begin{table}
\caption{Summary of impacts towards the graph representation learning performance with the intervention of different model structural or feature modules}
\centering
\begin{tabular}{ccc}
\hline\hline
Modules &  Fidelity& Diversity\\
\hline\hline
+Style&  +& -\\
\hline
+SVD&  +& +\\
\hline
+NED&  -& +\\
\hline
+VQ&  +& -\\
\hline
+poly&  +& -\\
\hline
+label&  +& -\\
\hline
+zdim&  -& +\\
\hline\hline
\end{tabular}
\label{table:summary_FD}
\end{table}

We summarize the impacts of various structural and feature interventions on the performance of the graph representation learning model in Table \ref{table:summary_FD}, which provides a detailed overview of how different modifications to the model's structure or its feature modules affect the fidelity and diversity levels of the learned graph representation. By organizing this information, we can more easily understand the complex interplay among various model implementation schemes concerning graph representation learning performance, offering a valuable reference point for elucidating the trade-offs and synergies inherent in different model design choices.

\begin{figure*}[!t]
  \centering
  \includegraphics[width=1.0\linewidth]{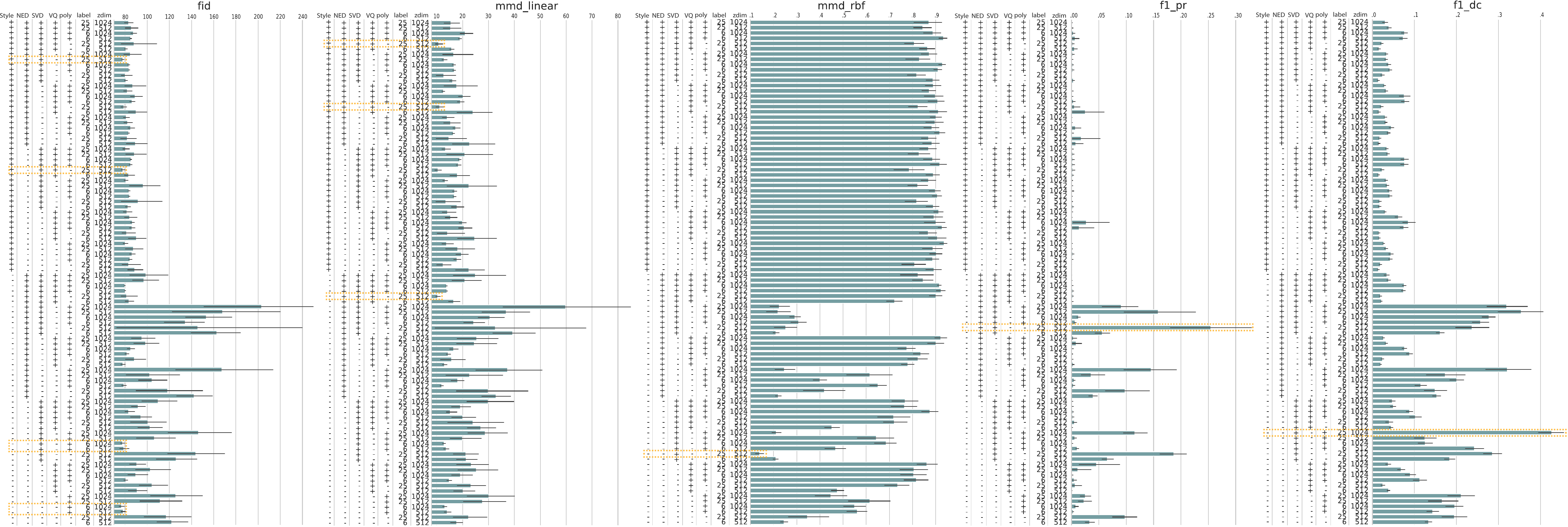}
  % \vspace{\BeforeCaptionVSpace}
  \caption{Comparison of different graph representation model design choices and their corresponding metric values, including FID(FD), MMD Linear, MMD RBF, F1 PR, and F1 DC; the sweet spots for each metric measure are highlighted in dashed box}
  \label{fig:anovas-mmdf1}
\end{figure*}

\begin{figure*}[!t]
  \centering
  \includegraphics[width=1.0\linewidth]{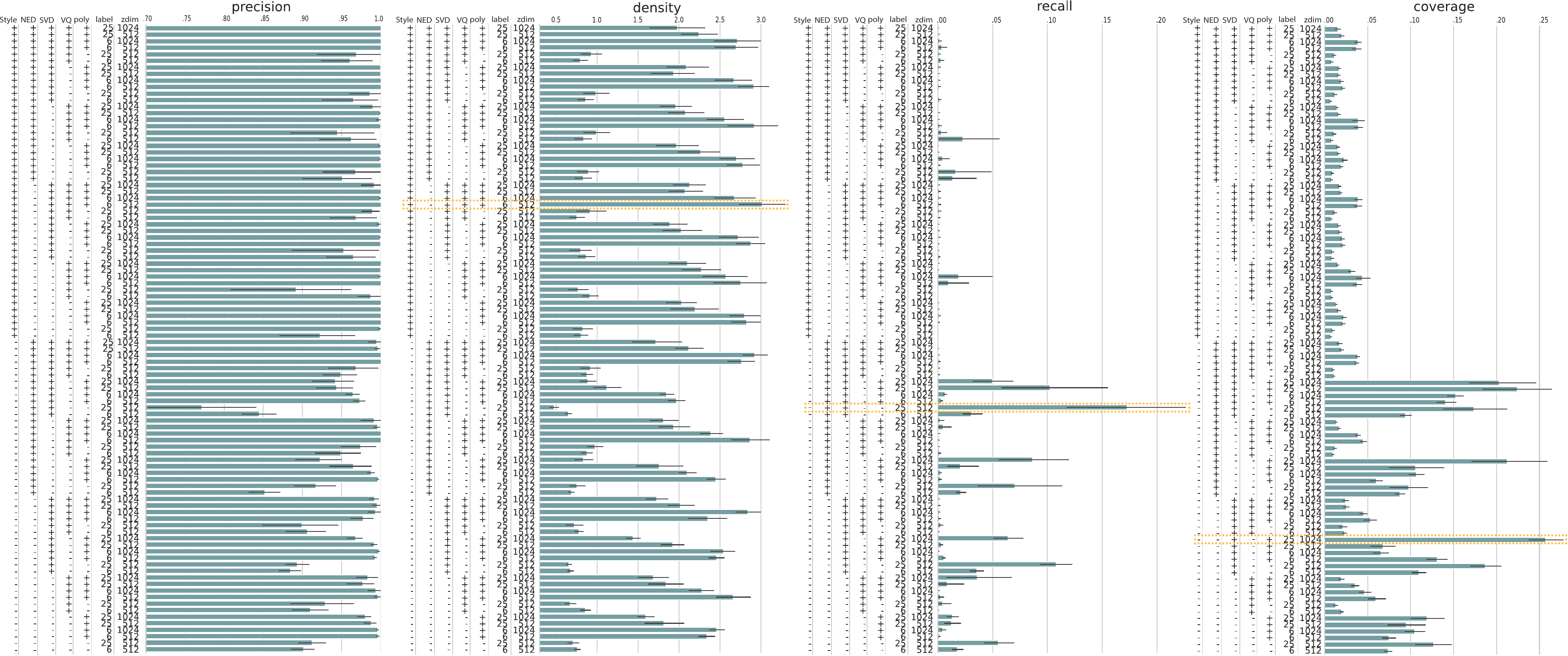}
  % \vspace{\BeforeCaptionVSpace}
  \caption{Comparison of different graph representation model design choices and their corresponding metric values, including precision, density, recall, and coverage; the sweet spots for each metric measure are highlighted in dashed box}
  \label{fig:anovas-PRDC}
\end{figure*}

To further deepen our understanding of the impacts of various implementation choices on the model performance, we systematically compare the effects of different combinations of model design choices on the performance metrics of the graph representation learning models. By applying this method to the diverse range of model design choices and their respective performance metrics, we aimed to identify the sweet spot of graph representation model structure and feature interventions. We highlight the sweet spot combinations of graph representation model structural intervention and feature augmentation choices in Fig.~\ref{fig:anovas-mmdf1} and Fig.~\ref{fig:anovas-PRDC}. Additionally, we conduct One-way ANOVA analysis across all possible combination groups to identify the statistical significance of differences among the various model design choice combinations; the One-way ANOVA results of all group comparisons yield significant F statistics (please refer to Appendix \ref{quan_app} for more details), indicating that the differences in performance metrics across various model design choice combinations––whether they involve structural modifications or feature enhancements––are statistically significant and not due to random variations.

Fig.~\ref{fig:anovas-mmdf1} and Fig.~\ref{fig:anovas-PRDC} jointly reveal some interesting insights about the impact of different model design choices on the performance of graph representation learning concerning both fidelity and diversity. It can be observed that the involvement of certain model design choices can have dominant impacts on certain graph representation learning performance metrics. Notably, the inclusion of a style-based decoder results in mutually significantly improved metrics for fidelity, including ``FD'', ``MMD Linear'', and ``Precision''. These findings underscore the effectiveness of the layer-wise stochasticity mechanism in enhancing the fidelity of generated graphs. Contrary to the results observed with the style-based decoder, the involvement of a vector quantisation module only positively impacts ``FD'' and ``MMD Linear'' metrics without significant improvement in ``Precision''. This suggests that while vector quantization mechanisms contribute to fidelity, their effectiveness is less pronounced than the layer-wise stochasticity mechanism. Interestingly, the implementation of layer-wise stochasticity and vector quantization mechanisms negatively affects diversity-relevant metrics, including ``MMD RBF'', ``F1 PR'', ``Recall'', and ``Coverage''. This indicates that these mechanisms do not inherently contribute to the diversity level of generated graphs.

The comparison analysis further highlights a consistent trend across various diversity-relevant metrics, including ``MMD RBF'', ``F1 PR'', ``Recall'', and ``Coverage'', wherein models incorporating the SVD encoding scheme consistently outperform others. This observation underscores the critical role of positional encoding mechanisms, such as SVD, in facilitating diverse graph generation. This leads to the conclusion that by incorporating positional information into the learning process, graph representation learning models can effectively capture spatial relationships and structural nuances within the graph data, thus enhancing the generated graphs' diversity. Moreover, the significance of positional encoding extends beyond individual metrics, emphasising the importance of considering and integrating positional encoding techniques in developing graph generation models.

\subsubsection{Qualitative evaluation}
\label{results:gnn:qual}

We convert the generated attributed adjacency multigraph into graphical floor plans to clearly delineate the complex information within the edge and node feature matrices, facilitating a direct comparison of model-generated layouts. Such comparisons are crucial for assessing model fidelity, identifying strengths, and pinpointing areas for refinement. Detailed conversion steps are provided in Appendix \ref{study2:results:gnn:visualization}.

Based on the outcomes of our quantitative evaluation in Section \ref{results:gnn:quant}, we carefully select a series of models with representative framework setups. These models are chosen due to their distinct structural configurations and varied approaches to processing the graph data, offering a diverse perspective on model performance. To further analyse and visualise these models' performance, we randomly sample 1000 latent codes $\textbf{z}$ to ensure pseudo-exhaustive coverage of the distribution of the learned latent space for each selected model. These latent codes are high-dimensional vectors that represent the compressed, encoded information derived from the training graph data. To visualize these high-dimensional latent codes in a more interpretable manner, we use Uniform Manifold Approximation and Projection (UMAP) \cite{mcinnes2018umap} to map the latent codes into a 2-dimensional space while preserving the essential structures and relationships, allowing us to observe patterns, clusters, and variations among the encoded representations of different models. 

\begin{figure}[!t]
  \centering
  \includegraphics[width=0.8\linewidth]{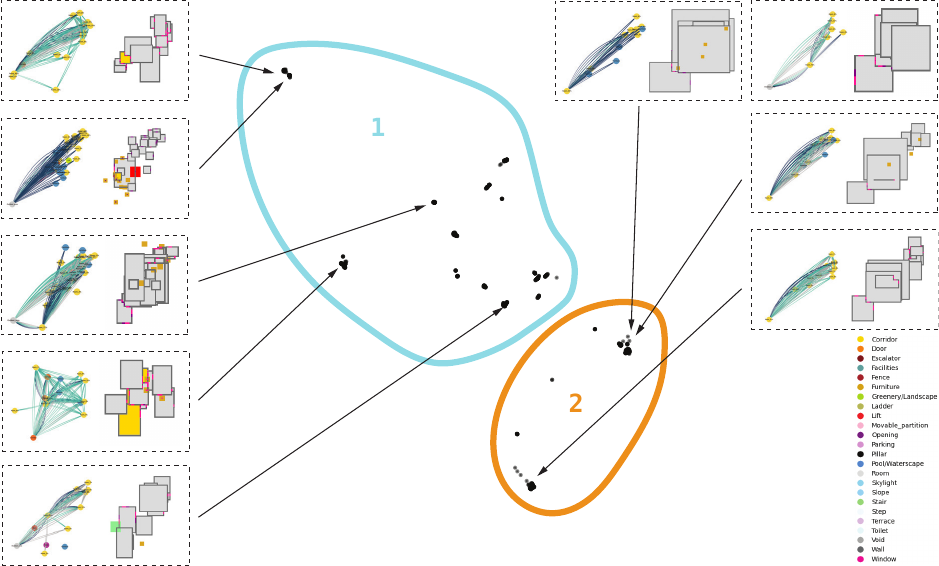}
  % \vspace{\BeforeCaptionVSpace}
  \caption{Generated graph samples and their corresponding locations in the learned latent space using a trained framework with edge-augmented encoder, vanilla VAE disentanglement module, MLP-based decoder, SVD embeddings and 25 categories of architectural elements}
  \label{fig:sum_svd8_betavae_FP25_undirected_basicNode_multiEdge_512_21}
\end{figure}

\begin{figure}[!t]
  \centering
  \includegraphics[width=0.5\linewidth]{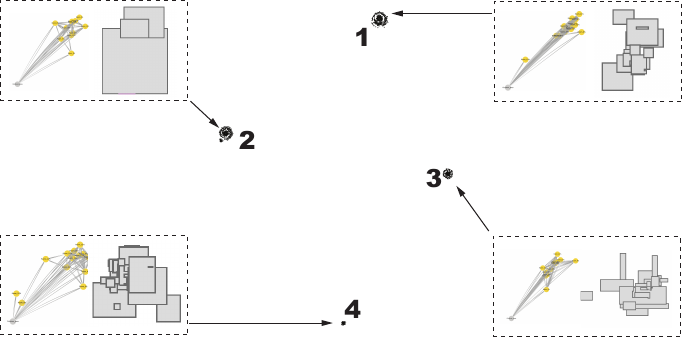}
  % \vspace{\BeforeCaptionVSpace}
  \caption{Generated graph samples and their corresponding locations in the learned latent space using a trained framework with edge-augmented encoder, vector quantisation disentanglement module, MLP-based decoder, SVD embeddings and 6 categories of architectural elements}
  \label{fig:sum_svd8_vqvae_FP6_undirected_basicNode_multiEdge_512_41}
\end{figure}

\begin{figure}[!t]
  \centering
  \includegraphics[width=0.7\linewidth]{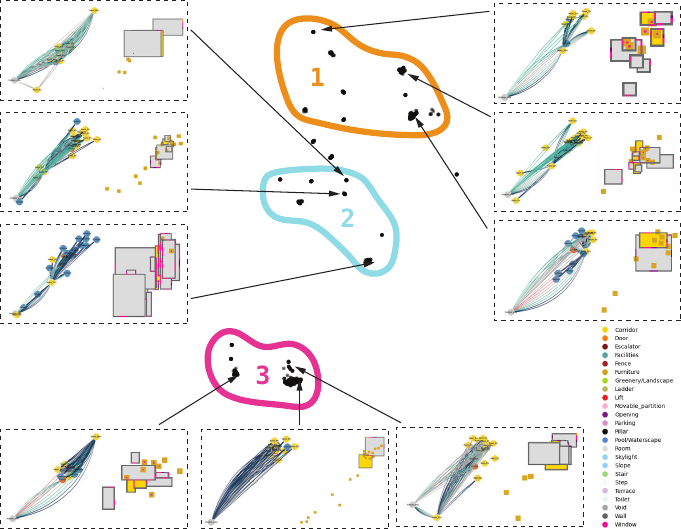}
  % \vspace{\BeforeCaptionVSpace}
  \caption{Generated graph samples and their corresponding locations in the learned latent space using a trained framework with edge-augmented encoder, vanilla VAE disentanglement module, MLP-based decoder, SVD embeddings, 25 categories of architectural elements, extra features of polygon vertices' coordinates, and boosted dimensions of the latent space}
  \label{fig:sum_svd8_betavae_FP25_undirected_fullNode_multiEdge_1024_15}
\end{figure}

The exploration of spatial relationship patterns through the proposed graph representation learning framework, as depicted in the series of demonstrations (Fig.~\ref{fig:sum_svd8_betavae_FP25_undirected_basicNode_multiEdge_512_21}, Fig.~\ref{fig:sum_svd8_vqvae_FP6_undirected_basicNode_multiEdge_512_41}, Fig.~\ref{fig:sum_svd8_betavae_FP25_undirected_fullNode_multiEdge_1024_15}), reveals some insights into the capabilities of these models in disentangling complex architectural elements. This analysis aligns with the findings presented in the quantitative evaluation section (Section \ref{results:gnn:quant}). 

Specifically, the model options with the SVD embeddings and extra architectural element categories show a high level of diverse clustering of learned latent graph patterns. The marked variation factor trends in nearby clusters indicate a rich and varied understanding of spatial relationships. This diversity in clustering highlights the model's ability to capture a wide array of spatial patterns effectively (Fig.~\ref{fig:sum_svd8_betavae_FP25_undirected_basicNode_multiEdge_512_21}). Contrasting with the previous model, the same GNN setup, except for the employment of the VQ mechanism, has learned a more distinctly disentangled space. However, it exhibits fewer clusters and a lower level of diversity. This suggests a more focused but less varied understanding of spatial relationships (Fig.~\ref{fig:sum_svd8_vqvae_FP6_undirected_basicNode_multiEdge_512_41}). Finally, the most complex setup, which includes SVD embeddings, an increased number of architectural element categories, features of polygon vertices' coordinates, and boosted dimensions of the latent space, demonstrates both a high level of disentanglement and diversity, indicative of an advanced understanding and representation of layout design patterns (Fig.~\ref{fig:sum_svd8_betavae_FP25_undirected_fullNode_multiEdge_1024_15}). More clustering visualization samples can be found in Appendix \ref{qual_app}.

\begin{figure}[!t]
  \centering
  \includegraphics[width=1.0\linewidth]{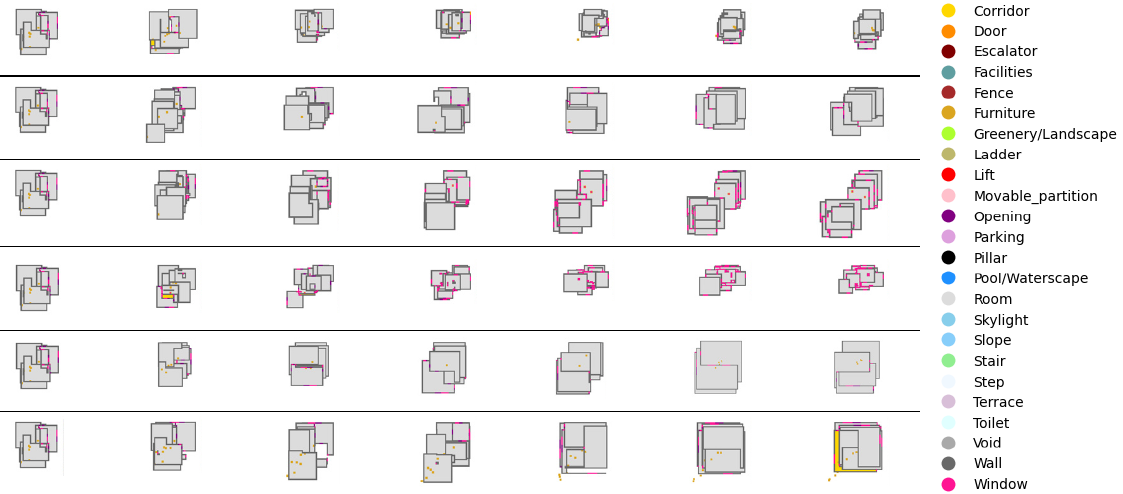}
  % \vspace{\BeforeCaptionVSpace}
  \caption{Linear interpolation samples starting from the same latent code $z$ of the learned latent space of a trained framework with edge-augmented encoder, vanilla VAE disentanglement module, MLP-based decoder, SVD embeddings, 25 categories of architectural elements, extra features of polygon vertices' coordinates, and boosted dimensions of the latent space}
  \label{fig:FP-interpolation_sum_svd8_betavae_FP25_undirected_fullNode_multiEdge_1024_one_}
\end{figure}

\begin{figure}[!t]
  \centering
  \includegraphics[width=1.0\linewidth]{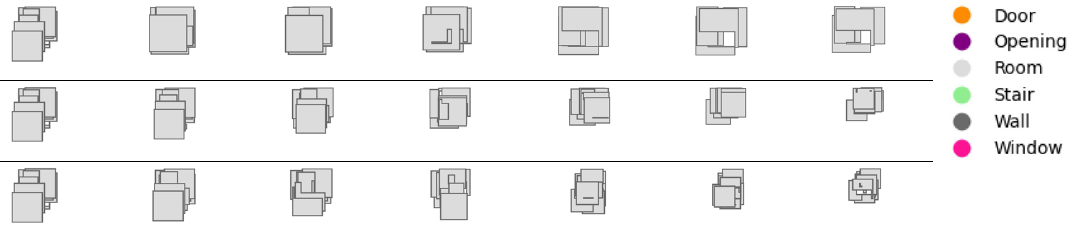}
  % \vspace{\BeforeCaptionVSpace}
  \caption{Linear interpolation samples starting from the same latent code $z$ of the learned latent space of a trained framework with edge-augmented encoder, vanilla VAE disentanglement module, MLP-based decoder, SVD embeddings, 6 categories of architectural elements, extra features of polygon vertices' coordinates, and boosted dimensions of the latent space}
  \label{fig:FP-interpolation_sum_svd8_betavae_FP6_undirected_fullNode_multiEdge_1024_one_}
\end{figure}

\begin{figure}[!t]
  \centering
  \includegraphics[width=1.0\linewidth]{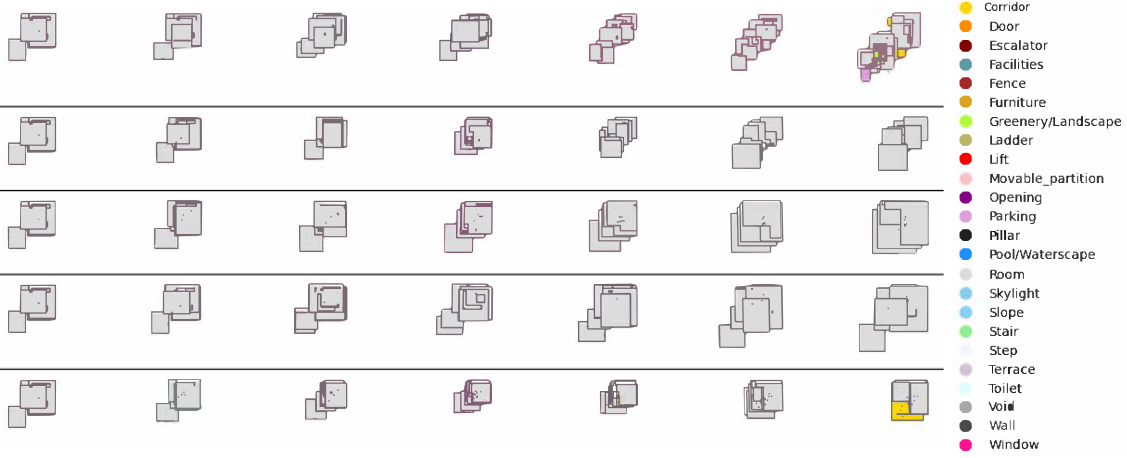}
  % \vspace{\BeforeCaptionVSpace}
  \caption{Linear interpolation samples starting from the same latent code $z$ of the learned latent space of a trained framework with edge-augmented encoder, vanilla VAE disentanglement module, MLP-based decoder, SVD embeddings, and 25 categories of architectural elements}
  \label{fig:FP-interpolation_sum_svd8_betavae_FP25_undirected_fullNode_multiEdge_512_one_}
\end{figure}

Meanwhile, when learning a latent code representation of a graph, we assume that each variable in the latent code corresponds to a certain factor or property used to generate the graphs’ edge and node attributes. Thus, by continuously changing the value of one variable and fixing the remaining variables, we can visualize the corresponding change in the generated graphs. Given the absence of a predefined list of layout design variables within the curated graph dataset and the impracticality of manually encoding such information, we opt for an unsupervised approach to assess the manipulation capabilities of the trained graph representation learning models. Specifically, we employ linear interpolation techniques on the learned latent space of selected models, utilizing a series of randomly generated latent code pairs to evaluate the disentanglement performance of our proposed framework. This involves simulating the graph feature manipulation process through linear interpolation operations between each pair of latent codes $z$. Fig.~\ref{fig:FP-interpolation_sum_svd8_betavae_FP25_undirected_fullNode_multiEdge_1024_one_} illustrate a series of linear interpolation samples of the learned latent space of a framework composed of an edge-augmented encoder, a vanilla VAE disentanglement module, and an MLP-based decoder, and trained with SVD embeddings, 25 categories of architectural elements, extra features of polygon vertices' coordinates, and boosted dimensionality of the latent space. Distinct trends in graph feature alterations are evident when examining interpolation samples derived from diverse latent code pairs residing across various regions of the learned latent graph feature space, which has been enhanced in dimensionality. These trends encompass modifications in layout features such as the proportion of space area, orientation of room layout, density of spaces, and organizational flow within the layout. Similarly, linear interpolation samples of the learned latent space of the same framework configuration but limited to 6 categories of architectural elements (Fig.~\ref{fig:FP-interpolation_sum_svd8_betavae_FP6_undirected_fullNode_multiEdge_1024_one_}) demonstrate similar identifiable trends in the modification of separable layout features. Likewise, Fig.~\ref{fig:FP-interpolation_sum_svd8_betavae_FP25_undirected_fullNode_multiEdge_512_one_} showcase samples derived from the learned latent space of the identical framework, albeit without additional polygon vertices' coordinates and a lesser dimensionality of the latent space. While similar trends in layout feature manipulation are discernible, the generated layouts exhibit a tendency towards oversimplification and a lack of detailed complexity, likely due to the absence of supplementary polygon vertices' coordinates and the reduced latent space dimensionality. More linear interpolation samples are demonstrated in Appendix \ref{qual_app}.

Exploring different configurations of the proposed framework demonstrates varied capabilities and performance levels in disentangling and interpreting the latent architectural layout design space. Certain configurations excel in disentangling spatial patterns, while others provide a richer diversity in the representation learning of layout features. The selection of a graph representation learning model setup plays a crucial role in balancing disentanglement and diversity. This underscores the significance of thoughtful model configuration in effectively interpreting architectural design data spaces.

\section{Discussion and Future Works}

Our empirical experiments have demonstrated the robustness and generalizability of our approaches in learning disentangled graph representations and interpreting the graph-based latent architectural design layout space. Meanwhile, this study's extensive quantitative and qualitative experiments have also shed light on a few critical aspects concerning disentangled representation learning and deep generative modelling of graph data. These insights pave the way for a series of promising future research avenues, setting the stage for further exploration and refinement in this field.

\subsection{Trade-off between disentanglement, fidelity, and diversity}

One crucial aspect identified in our study is the trade-off between disentanglement, fidelity, and diversity of the learned architectural layout graph representations, highlighting the complexity of learning and disentangling architectural layout graph representations. The empirical experiments have shown that different structural modifications and feature enhancements significantly affect the performance of graph representation learning in different aspects. For instance, using a style-based decoder instead of a vanilla MLP-based decoder results in improved fidelity but does not positively impact diversity. Similarly, adding polygon vertice coordinates, increasing the number of architectural element label categories, and using a vector quantization mechanism for latent space disentanglement also show trade-offs between improved fidelity and reduced diversity. Furthermore, elevating the latent code dimensionality improves diversity but slightly compromises fidelity. Nevertheless, incorporating SVD-based positional encoding enhances both fidelity and diversity. These findings underline the complexities in balancing these aspects through various implementations, highlighting the need for strategic model design to optimize performance across fidelity and diversity metrics.

The consistent positive effect of incorporating the SVD encoding scheme underscores the importance of integrating positional encoding techniques in developing architectural layout design graph generation models. Thus, future works may consider testing different positional and structural encoding mechanisms to explore further how spatial relationships and features of layout graphs can be more effectively learned and interpreted. By further investigating different positional and structural encoding strategies, it would be possible to identify more efficient positional and structural encoding schemes for capturing the nuances of layout design features, which are critical for generating more accurate and functional architectural layout graphs with a higher level of diversity.

Meanwhile, this study has yet to explore certain model implementation variations that may also be related to the trade-off issue, presenting opportunities for future research. These include adjusting the regularization coefficients for various loss terms and incorporating domain-specific knowledge into node ordering schemes. Exploring the regularization coefficients may optimize the model’s capacity to manage competing objectives, potentially improving its overall effectiveness concerning both fidelity and diversity. Additionally, different graph domains might see improvements from customized node orderings \cite{liao2019efficient}, and the application of domain-specific insights into standard orderings could also be potentially beneficial.

\subsection{Evaluation metric effectiveness and suitability}

Another key aspect identified is the consistency issue across different evaluation metrics. Our quantitative results indicate potential discrepancies between image and graph generation metrics. Specifically, while ``FID (FD)'' is widely used to assess both the fidelity and diversity of generated images, its effectiveness in measuring diversity within graph generation is questionable. This is illustrated by conflicting outcomes when comparing ``FID (FD)'' with other diversity-oriented metrics such as ``MMD RBF'', ``F1 PR'', ``Recall'', and ``Coverage''. The ``MMD Linear'' and ``MMD RBF'' measures also show potential conflicts, emphasizing their subtle differences in their focused data properties.

These discrepancies underscore the complexity involved in the metric selection and emphasize the importance of choosing appropriate metrics that align with specific research objectives for various graph generation tasks. This also highlights the need for further research into metric effectiveness and suitability, ensuring that the metrics employed provide meaningful insights and support the intended outcomes of the graph generation tasks.

\subsection{Generalization across other graph generation domains}

Exploring the applicability of the proposed framework across other graph generation domains represents an intriguing avenue for future research. While testing this framework in other contexts is feasible, it was beyond the scope of the current study. Extending this research to other domains could provide valuable insights into the framework’s versatility and effectiveness in different settings. Such investigations could further validate the framework’s broader utility and potentially uncover domain-specific challenges and opportunities for refinement. For instance, as our findings underscore the efficacy of the SVD encoding scheme and highlight the pivotal role of positional encoding techniques in enhancing the capabilities of architectural layout design graph generation models, future research may further investigate this aspect and extend our framework across various domains to ascertain its generalizability and efficiency in different settings.

\section{Conclusion}

In conclusion, this study represents a pioneering effort to address significant research gaps in the domain of architectural layout design graph generation and graph-based design representation space interpretation. We have initiated the disentangled representation learning of architectural layout design graphs by introducing the Style-based Edge-augmented Variational Graph Auto-Encoder (SE-VGAE) framework. The proposed framework allows for a nuanced exploration of the complex interrelationships of different model design configurations, facilitating a deeper understanding of graph representation learning concerning the generation of architectural layout graphs. 

Moreover, the introduction of a novel benchmark large-scale architectural layout design graph dataset marks another significant contribution. This dataset provides a comprehensive resource for training and evaluating graph generation models in this domain. This dataset enriches the field and sets a foundation for future research to explore and identify latent architectural design layout patterns and relationships.

Our study advances the theoretical understanding of graph-based architectural design and offers practical insights and tools for researchers and practitioners in relevant fields. The exploration of disentangled representation learning in the context of architectural layout design graphs illuminates an innovative path forward, suggesting that much can be gained by continuing to explore and refine techniques in this field. This work lays the groundwork for future explorations aimed at enhancing the robustness, accuracy, and diversity of graph generation models of architectural layout design and beyond.

\section*{Acknowledgments}

The computational work for this article was performed on resources of the National Supercomputing Centre, Singapore (https://www.nscc.sg). The data sources used in this study are also gratefully acknowledged. This research was supported by the President's Graduate Fellowship of the National University of Singapore and the Singapore Data Science Consortium (SDSC) Dissertation Research Fellowship.

\vspace{11pt}

\begin{IEEEbiography}[{\includegraphics[width=1in,height=1.25in,clip,keepaspectratio]{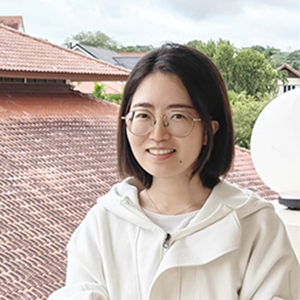}}]{Jielin Chen}
is a PhD candidate in Architecture at the National University of Singapore. She obtained her MLA (Distinction) from the University of Hong Kong and a BEng in Urban Planning from Zhejiang University. Her research specializes in design computing, with a focus on computational methods for design representation space interpretation. She is dedicated to disentangling computational design representation and seeking innovative solutions to complex architectural design research challenges.
\end{IEEEbiography}
\vspace{-33pt}
\begin{IEEEbiography}[{\includegraphics[width=1in,height=1.25in,clip,keepaspectratio]{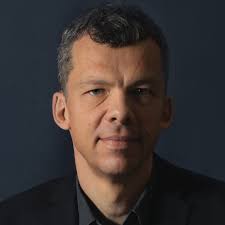}}]{Rudi Stouffs}
is Associate Professor at the Department of Architecture and Assistant Dean (Research) at the College of Design and Engineering, National University of Singapore. He leads the Architectural and Urban Prototyping lab. His research interests include computational issues of description, modelling and representation for design, in the areas of shape recognition and design generation, building information modelling and analysis, virtual cities and digital twins.
\end{IEEEbiography}

\newpage

% {\appendix[Proof of the Zonklar Equations]
% Use $\backslash${\tt{appendix}} if you have a single appendix:
% Do not use $\backslash${\tt{section}} anymore after $\backslash${\tt{appendix}}, only $\backslash${\tt{section*}}.
% If you have multiple appendixes use $\backslash${\tt{appendices}} then use $\backslash${\tt{section}} to start each appendix.
% You must declare a $\backslash${\tt{section}} before using any $\backslash${\tt{subsection}} or using $\backslash${\tt{label}} ($\backslash${\tt{appendices}} by itself
%  starts a section numbered zero.)

\appendices

\section{Methodology}
\label{method_app}

The pseudo-code of the vanilla VAE module is demonstrated in Algorithm \ref{alg:alg1}. It is worth noting that $z_{\sigma}$ is calculated as the log-variance for numerical stability reasons, and the standard deviation $\sigma$ is computed by $e^{\frac{z_{\sigma}}{2}}$, as $e^{\frac{z_{\sigma}}{2}} = e^{\frac{log (\sigma^{2})}{2}} = \sigma$.

\begin{algorithm}[H]
\caption{Pseudo code of the vanilla VAE module}\label{alg:alg1}
\begin{algorithmic}[1]
\STATE {\textsc{\textbf{input}: }}$X' \in \mathbb{R} ^{n\times d}, A^{e'} \in \mathbb{R} ^{n\times n \times c}$
\STATE {\textsc{\textbf{hyperparameter}:}}
\STATE \hspace{0.5cm}dimension of latent space: $M$
\STATE {\textsc{\textbf{trainable parameters}:}}
\STATE \hspace{0.5cm}$\epsilon \in \mathbb{R}$
\STATE \hspace{0.5cm}Edge feature mapping layer $f_{nn}^{e}: \mathbb{R}^{c}\rightarrow \mathbb{R}$
\STATE \hspace{0.5cm}Linear layer $l^{nn}$
\STATE \hspace{0.5cm}Parametric Rectified Linear Unit (PReLU) activation layer $pr$
\STATE \hspace{0.5cm}Layer normalization layer $l_n$
\STATE {\textsc{\textbf{training process}: }}
\STATE \hspace{0.5cm}$\widehat{X}' = (1 + \epsilon) \odot X'$
\STATE \hspace{0.5cm}$\widehat{A}^{e'} = f_{nn}^{e}\left(A^{e'}\right)$
\STATE \hspace{0.5cm}$\widehat{\widehat{X}}' = \widehat{A}^{e'} \cdot \widehat{X}'$ 
\STATE \hspace{0.5cm}$\widehat{\widehat{\widehat{X}}}' = \widehat{X}' + \widehat{\widehat{X}}'$
\STATE \hspace{0.5cm}$\widehat{\widehat{\widehat{X}}}' = l_n\left(pr(\left(l^{nn}_{d\rightarrow d}\left(\widehat{\widehat{\widehat{X}}}'\right)\right)\right), \widehat{\widehat{\widehat{X}}}' \in \mathbb{R} ^{n\times d}$
\STATE \hspace{0.5cm}$\overline{z} = \sum ^{n}\widehat{\widehat{\widehat{X}}}'_{n,d}, \overline{z} \in \mathbb{R} ^d$
\STATE \hspace{0.5cm}$z_{\mu} = l_n\left(pr(\left(l^{nn}_{M\rightarrow M}\left(l_n\left(pr(\left(l^{nn}_{d\rightarrow M}\left(\overline{z}\right)\right)\right)\right)\right)\right), z_{\mu} \in \mathbb{R} ^M$
\STATE \hspace{0.5cm}$z_{\sigma} = l_n\left(pr(\left(l^{nn}_{M\rightarrow M}\left(l_n\left(pr(\left(l^{nn}_{d\rightarrow M}\left(\overline{z}\right)\right)\right)\right)\right)\right), z_{\mu} \in \mathbb{R} ^M$
\STATE \hspace{0.5cm}$z = z_{\mu} + e^{\frac{z_{\sigma}}{2}} \odot r, z \sim \mathcal{N} \left(z_{\mu},e^{z_{\sigma}}\right)$,
\STATE \hspace{3cm}$r \sim \mathcal{N} \left( 0,\textbf{I}\right), r \in \mathbb{R} ^M$
\STATE {\textsc{\textbf{return}: }}$z$
\end{algorithmic}
\label{alg1}
\end{algorithm}
% \

The pseudo-code of the NED-based disentanglement module is demonstrated in Algorithm \ref{alg:alg3}. Specifically, the node-edge co-encoder learns the mean $z_{\mu}^{graph}$ and standard deviation $z_{\sigma}^{graph}$ of the latent representation of the entire graph and the corresponding $z^{graph} \sim \mathcal{N} \left(z_{\mu}^{graph},e^{2z_{\sigma}^{graph}} \textbf{I}\right)$. Similarly, the node encoder learns the mean $z_{\mu}^{node}$ and standard deviation $z_{\sigma}^{node}$ of node representation and the corresponding $z^{node} \sim \mathcal{N} \left(z_{\mu}^{node},e^{2z_{\sigma}^{node}} \textbf{I}\right)$, and the edge encoder learns the mean $z_{\mu}^{edge}$ and standard deviation $z_{\sigma}^{edge}$ of edge representation and the corresponding $z^{edge} \sim \mathcal{N} \left(z_{\mu}^{edge},e^{2z_{\sigma}^{edge}} \textbf{I}\right)$. Following this, $z^{node}$ and $z^{edge}$ are respectively fused with $z^{graph}$ to generate $z^{node+graph}$ and $z^{edge+graph}$, which are subsequently inputted into the node and edge sub-decoders concurrently.

\begin{algorithm}[H]
\caption{Pseudo code of the NED-based disentanglement module}\label{alg:alg3}
\begin{algorithmic}[1]
\STATE {\textsc{\textbf{input}: }}$X' \in \mathbb{R} ^{n\times d}, A^{e'} \in \mathbb{R} ^{n\times n \times c}$
\STATE {\textsc{\textbf{hyperparameter}:}}
\STATE \hspace{0.5cm}dimension of latent space: $M$
\STATE {\textsc{\textbf{trainable parameters}:}}
\STATE \hspace{0.5cm}$\epsilon \in \mathbb{R}$
\STATE \hspace{0.5cm}Edge feature mapping layer $f_{nn}^{e}: \mathbb{R}^{c}\rightarrow \mathbb{R}$
\STATE \hspace{0.5cm}Linear layer $l^{nn}$
\STATE \hspace{0.5cm}Parametric Rectified Linear Unit (PReLU) activation layer $pr$
\STATE \hspace{0.5cm}Layer normalization layer $l_n$
\STATE {\textsc{\textbf{training process}: }}
\STATE \hspace{0.5cm}Obtain $\overline{z}$ using the process provided in Algorithm \ref{alg:alg1} up till line 16
\STATE \hspace{0.5cm}$z_{\mu}^{graph} = l_n\left(pr(\left(l^{nn}_{M\rightarrow M}\left(l_n\left(pr(\left(l^{nn}_{d\rightarrow M}\left(\overline{z}\right)\right)\right)\right)\right)\right), z_{\mu} \in \mathbb{R} ^M$
\STATE \hspace{0.5cm}$z_{\sigma}^{graph} = l_n\left(pr(\left(l^{nn}_{M\rightarrow M}\left(l_n\left(pr(\left(l^{nn}_{d\rightarrow M}\left(\overline{z}\right)\right)\right)\right)\right)\right), z_{\mu} \in \mathbb{R} ^M$
\STATE \hspace{0.5cm}$z^{graph} = z_{\mu}^{graph} + e^{\frac{z_{\sigma}^{graph}}{2}} \odot r$,
\STATE \hspace{3.5cm}$ z^{graph} \sim \mathcal{N} \left(z_{\mu}^{graph},e^{z_{\sigma}^{graph}} \textbf{I}\right)$,
\STATE \hspace{3.5cm}$r \sim \mathcal{N} \left( 0,\textbf{I}\right), r \in \mathbb{R} ^M$
\STATE \hspace{0.5cm}$\overline{z}^{node} = \sum ^{n}X'_{n,d}, \overline{z}^{node} \in \mathbb{R} ^d$
\STATE \hspace{0.5cm}$z_{\mu}^{node} = l_n\left(pr(\left(l^{nn}_{M\rightarrow M}\left(l_n\left(pr(\left(l^{nn}_{d\rightarrow M}\left(\overline{z}^{node}\right)\right)\right)\right)\right)\right)$,
\STATE \hspace{5cm}$z_{\mu}^{node} \in \mathbb{R} ^M$
\STATE \hspace{0.5cm}$z_{\sigma}^{node} = l_n\left(pr(\left(l^{nn}_{M\rightarrow M}\left(l_n\left(pr(\left(l^{nn}_{d\rightarrow M}\left(\overline{z}^{node}\right)\right)\right)\right)\right)\right)$,
\STATE \hspace{5cm}$z_{\mu}^{node} \in \mathbb{R} ^M$
\STATE \hspace{0.5cm}$z^{node} = z_{\mu}^{node} + e^{\frac{e^{z_{\sigma}^{node}}}{2}} \odot r, z^{node} \sim \mathcal{N} \left(z_{\mu}^{node},e^{z_{\sigma}^{node}} \textbf{I}\right)$,
\STATE \hspace{5cm}$r \sim \mathcal{N} \left( 0,\textbf{I}\right), r \in \mathbb{R} ^M$
\STATE \hspace{0.5cm}$z^{node+graph} = l^{nn}_{2M\rightarrow M}\left(z^{node}||z^{graph}\right)$,
\STATE \hspace{5cm}$z^{node+graph} \in \mathbb{R} ^M$
\STATE \hspace{0.5cm}$\overline{z}^{edge} = l^{nn}_{n^{2}\rightarrow M}\left(flatten\left(f_{nn}^{e}\left(A^{e'}\right)\right)\right)$
\STATE \hspace{0.5cm}$z_{\mu}^{edge} = l_n\left(pr(\left(l^{nn}_{M\rightarrow M}\left(l_n\left(pr(\left(l^{nn}_{M\rightarrow M}\left(\overline{z}^{edge}\right)\right)\right)\right)\right)\right)$,
\STATE \hspace{5cm}$z_{\mu}^{edge} \in \mathbb{R} ^M$
\STATE \hspace{0.5cm}$z_{\sigma}^{edge} = l_n\left(pr(\left(l^{nn}_{M\rightarrow M}\left(l_n\left(pr(\left(l^{nn}_{M\rightarrow M}\left(\overline{z}^{edge}\right)\right)\right)\right)\right)\right)$,
\STATE \hspace{5cm}$z_{\mu}^{edge} \in \mathbb{R} ^M$
\STATE \hspace{0.5cm}$z^{edge} = z_{\mu}^{edge} + e^{\frac{e^{z_{\sigma}^{edge}}}{2}} \odot r, z^{edge} \sim \mathcal{N} \left(z_{\mu}^{edge},e^{z_{\sigma}^{edge}} \textbf{I}\right)$,
\STATE \hspace{5cm}$r \sim \mathcal{N} \left( 0,\textbf{I}\right), r \in \mathbb{R} ^M$
\STATE \hspace{0.5cm}$z^{edge+graph} = l^{nn}_{2M\rightarrow M}\left(z^{edge}||z^{graph}\right)$,
\STATE \hspace{5cm}$z^{edge+graph} \in \mathbb{R} ^M$
\STATE {\textsc{\textbf{return}: }}$z^{node+graph}, z^{edge+graph}$
\end{algorithmic}
\label{alg3}
\end{algorithm}

\section{Quantitative evaluation}
\label{quan_app}

\begin{figure}[!t]
  \centering
  \includegraphics[width=1.0\linewidth]{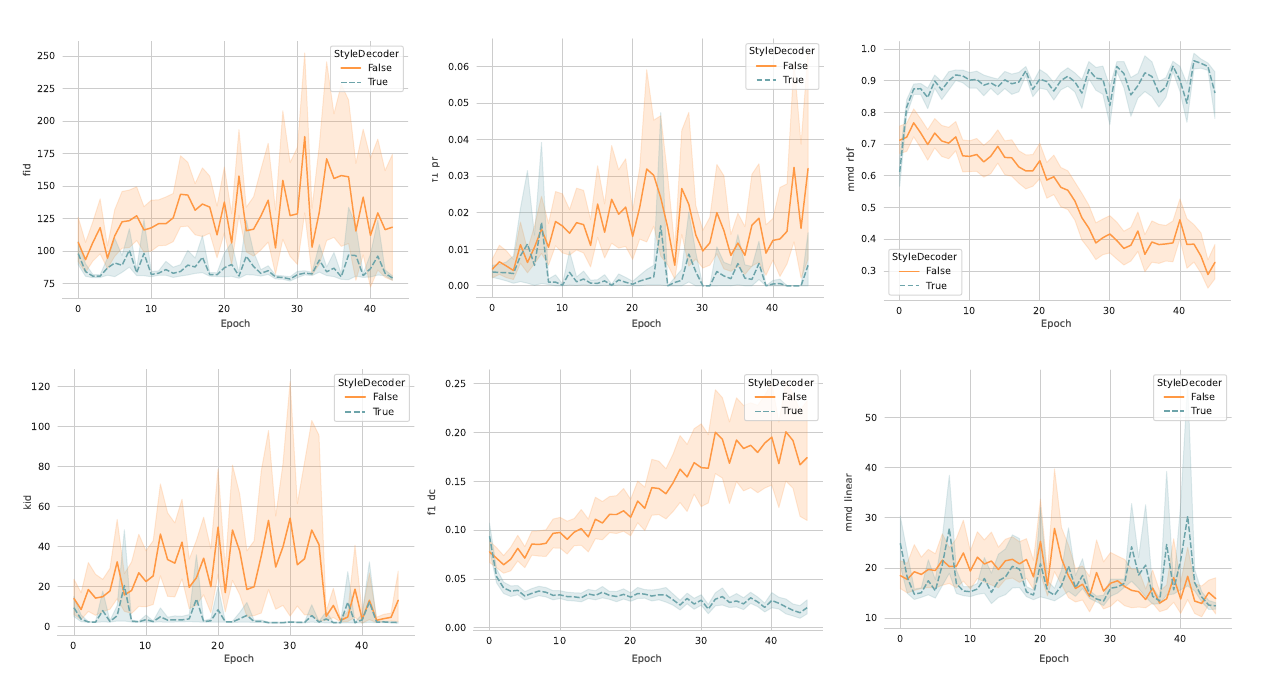}
  % \vspace{\BeforeCaptionVSpace}
  \caption{Comparison of style-based decoder and vanilla MLP-based decoder based on the 'FID', 'KID', 'F1 PR', 'F1 DC', 'MMD Linear', and 'MMD RBF' measures}
  \label{fig:style1}
\end{figure}

\begin{figure}[!t]
  \centering
  \includegraphics[width=1.0\linewidth]{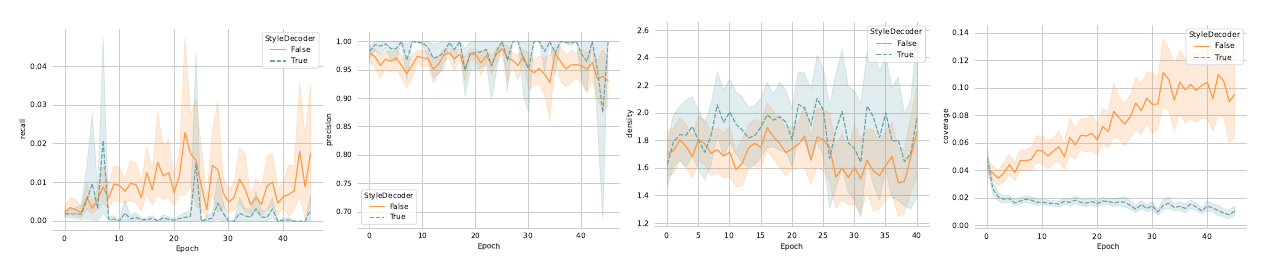}
  % \vspace{\BeforeCaptionVSpace}
  \caption{Comparison of style-based decoder and vanilla MLP-based decoder based on the 'precision', 'recall', 'density', and 'coverage' measures}
  \label{fig:style2}
\end{figure}

Our investigation indicates that incorporating a style-based decoder (``Style'' is True) compared to the vanilla MLP-based decoder (``Style'' is False) results in significantly lower ``FD'' and ``MMD Linear'' values, coupled with higher ``Precision'' and ``Density'' values (Fig.~\ref{fig:style1}, Fig.~\ref{fig:style2}). These findings highlight the effectiveness of the style-based decoder in enhancing fidelity levels within learned graph representations. However, it's notable that despite these improvements in fidelity, there is no discernible impact on diversity, as evidenced by comparable ``F1 PR'', ``F1 DC'', ``MMD RBF'', ``Recall'', and ``Coverage'' scores between the two decoder types. These results underscore the nuanced role of the style-based decoder, primarily contributing to fidelity enhancement while exhibiting limited influence on diversity under the specified conditions in our study.

\begin{figure}[!t]
  \centering
  \includegraphics[width=1.0\linewidth]{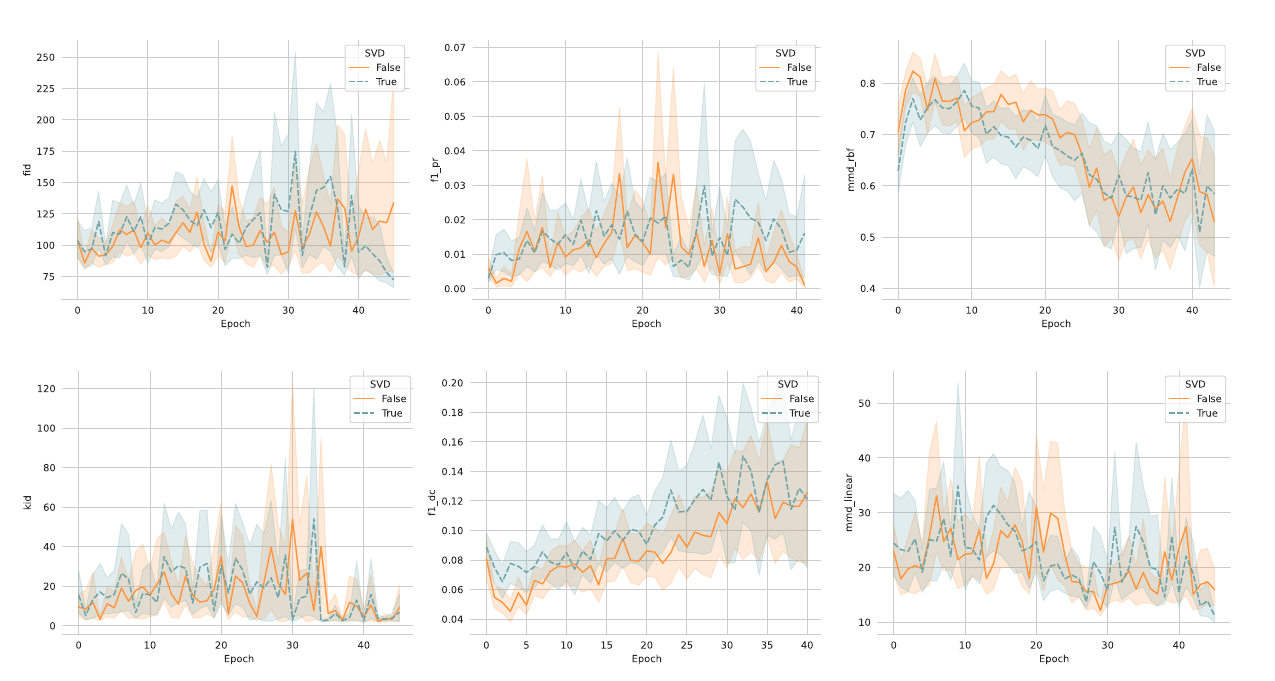}
  % \vspace{\BeforeCaptionVSpace}
  \caption{Comparison of the intervention of SVD positional embedding based on the 'FID', 'KID', 'F1 PR', 'F1 DC', 'MMD Linear', and 'MMD RBF' measures}
  \label{fig:SVD1}
\end{figure}

\begin{figure}[!t]
  \centering
  \includegraphics[width=1.0\linewidth]{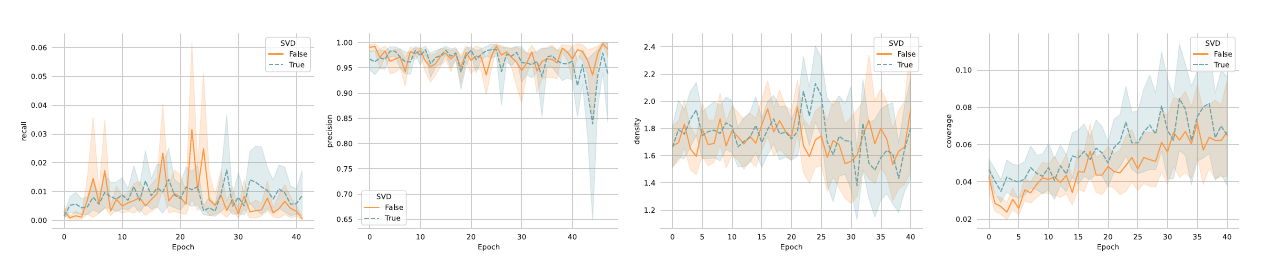}
  % \vspace{\BeforeCaptionVSpace}
  \caption{Comparison of the intervention of SVD positional embedding based on the 'precision', 'recall', 'density', and 'coverage' measures}
  \label{fig:SVD2}
\end{figure}

Our findings also reveal that the incorporation of singular value decomposition-based positional encoding (``SVD'' is True) generally leads to better graph representation learning performance. With this augmentation, ``MMD RBF'' registers a significantly lower value (Fig.~\ref{fig:SVD1}), while ``Coverage'' records a substantially higher value (Fig.~\ref{fig:SVD2}). The lower value of ``MMD RBF'' suggests improved fidelity and diversity in the representations, as it indicates a closer similarity to the target set, while concurrently, the increase in ``Coverage'' value also points to a heightened level of diversity within the learned representations. These results indicate that adding SVD positional encoding to node features can effectively enhance the model's ability to learn graph representations with higher fidelity and diversity levels.

\begin{figure}[!t]
  \centering
  \includegraphics[width=1.0\linewidth]{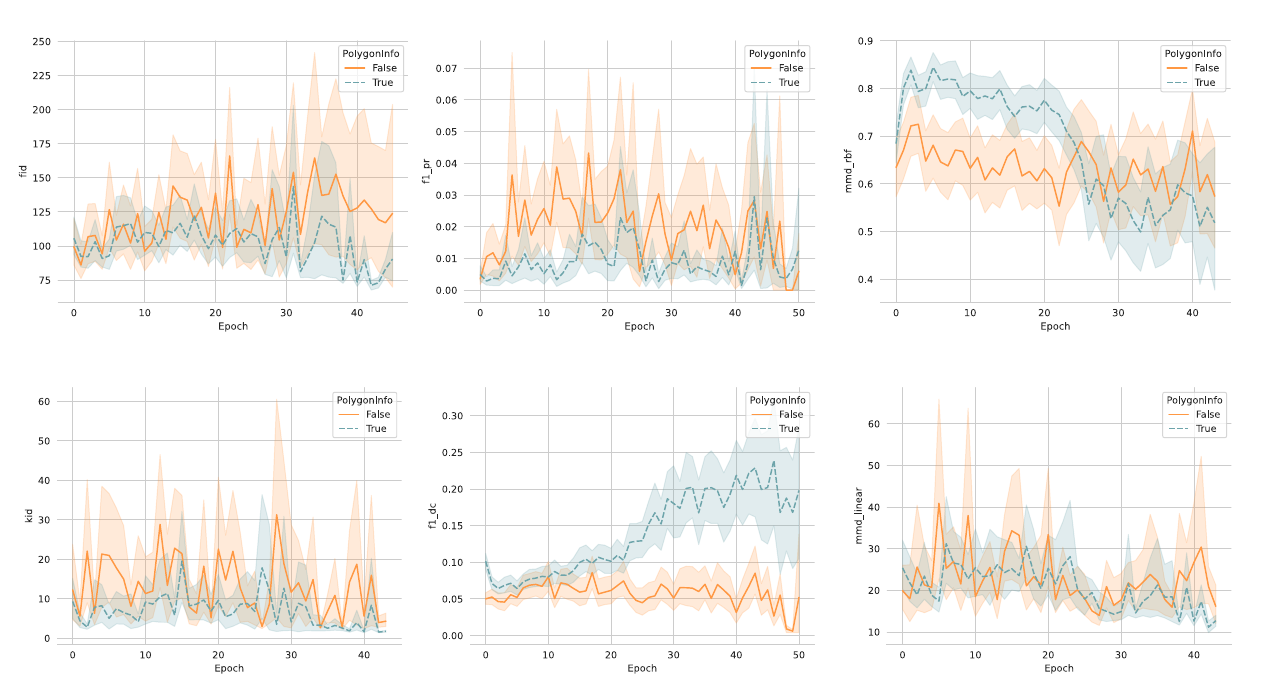}
  % \vspace{\BeforeCaptionVSpace}
  \caption{Comparison of the intervention of extra information of polygon vertices coordinates based on the 'FID', 'KID', 'F1 PR', 'F1 DC', 'MMD Linear', and 'MMD RBF' measures}
  \label{fig:PolygonInfo1}
\end{figure}

\begin{figure}[!t]
  \centering
  \includegraphics[width=1.0\linewidth]{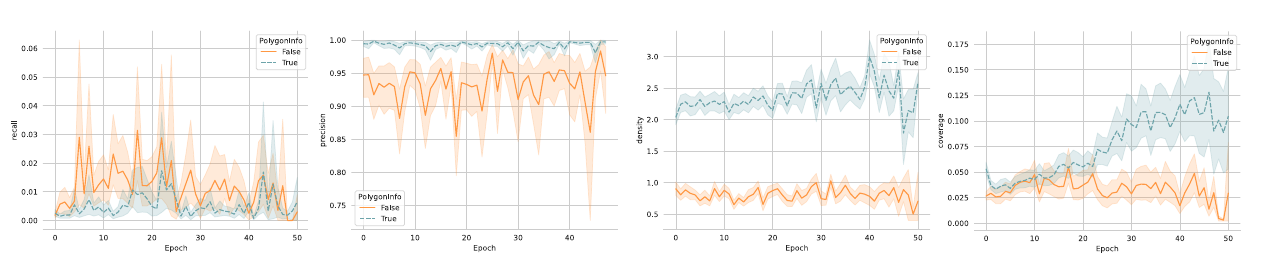}
  % \vspace{\BeforeCaptionVSpace}
  \caption{Comparison of the intervention of extra information of polygon vertices coordinates based on 'precision', 'recall', 'density', and 'coverage' measures}
  \label{fig:PolygonInfo2}
\end{figure}

Meanwhile, significant effects can also be observed when incorporating additional polygon vertices coordinates information into node features (``poly'' is True). With this enhancement, the ``FD'' value showed a notable decrease, while ``F1 DC'', ``Precision'', ``Density'', and ``Coverage'' values increased significantly (Fig.~\ref{fig:PolygonInfo1}, Fig.~\ref{fig:PolygonInfo2}). This trend suggests that integrating polygon vertices coordinate information into node features enhances the fidelity of the graph representations learned by the model. However, this improvement in fidelity comes with a trade-off. We can also note a significant increase in the ``MMD RBF'' value and a decrease in the ``F1 PR'' and the ``Recall'' values. This shift points to a potential reduction in the diversity of the learned graph representations. The increase in ``MMD RBF'' and the decrease in ``F1 PR'' and ``Recall'' suggest a reduction in the model's ability to capture the full range of the target graph representation distribution. Therefore, these results indicate a nuanced trade-off effect when adding polygon vertices coordinates information to node features. While it leads to higher fidelity in the graph representation learning model, it appears to do so at the expense of diversity. This phenomenon is somewhat predictable, as the inclusion of polygon vertices coordinates undoubtedly enriches the node representation with more pertinent information, enhancing the model's ability to capture detailed features and improve fidelity. Yet, this addition also substantially increases the dimensionality of the latent representation that the model needs to capture and inherently escalates the complexity of the representation learning task. This increased complexity can pose significant challenges for the model as it strives to accommodate the broader range of information within the higher-dimensional latent space. Consequently, this can lead to difficulties in effectively learning and representing the entire scope of the data, potentially resulting in mode dropping, i.e., the model fails to represent certain modes or variations within the data distribution, which can diminish the diversity of the learned representations.

\begin{figure}[!t]
  \centering
  \includegraphics[width=1.0\linewidth]{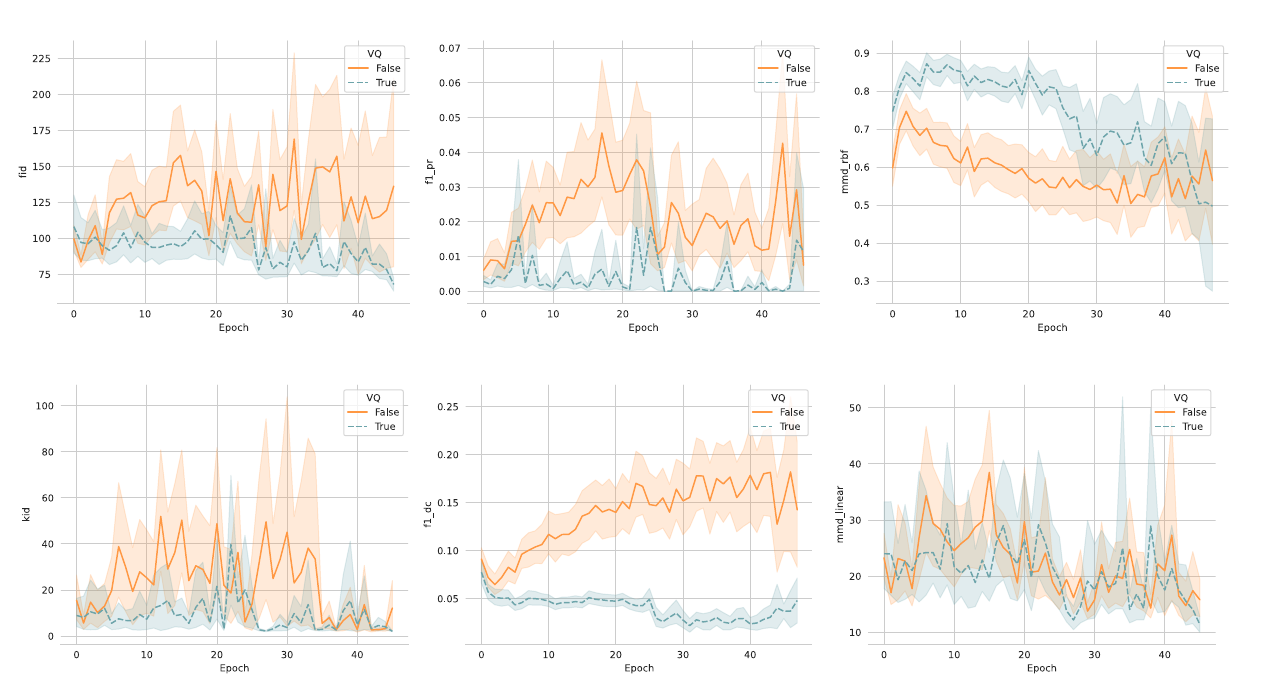}
  % \vspace{\BeforeCaptionVSpace}
  \caption{Comparison of the intervention of vector quantization mechanism based on the 'FID', 'KID', 'F1 PR', 'F1 DC', 'MMD Linear', and 'MMD RBF' measures}
  \label{fig:VQ1}
\end{figure}

\begin{figure}[!t]
  \centering
  \includegraphics[width=1.0\linewidth]{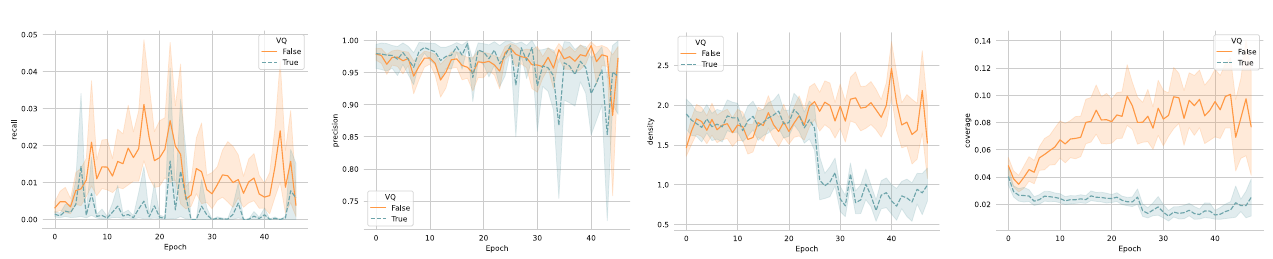}
  % \vspace{\BeforeCaptionVSpace}
  \caption{Comparison of the intervention of vector quantization mechanism based on the 'precision', 'recall', 'density', and 'coverage' measures}
  \label{fig:VQ2}
\end{figure}

Similarly, implementing a vector quantization mechanism for latent space disentanglement (``VQ'' is True) yielded a significant decrease in ``FD'' (Fig.~\ref{fig:VQ1}), while ``Precision'' increased considerably (Fig.~\ref{fig:VQ2}), suggesting that the incorporation of a vector quantization mechanism enhances the fidelity of graph representations learned by the model and pointing to a more accurate and closer match between the generated and real graph data. However, alongside these improvements, we also observed a trade-off regarding diversity: there was a significant increase in the ``MMD RBF'' value and marked decreases in ``Recall'', ``Density'', and ``Coverage'' values, suggesting a growing dissimilarity between the overall learned and target distributions and a diminished ability of the model to capture the full range and variety of the target distribution. Therefore, while the vector quantization mechanism improves fidelity, it does so at the expense of diversity within the learned representations as well. This trade-off underscores the complexity involved in latent space disentanglement: enhancing one aspect of the model's performance can potentially inadvertently impact another. The involvement of the vector quantization mechanism, while beneficial for achieving higher fidelity, also necessitates careful consideration of its effects on the diversity of the generated graph representations, highlighting the intricate balance required in the design and implementation of graph representation learning models.

\begin{figure}[!t]
  \centering
  \includegraphics[width=1.0\linewidth]{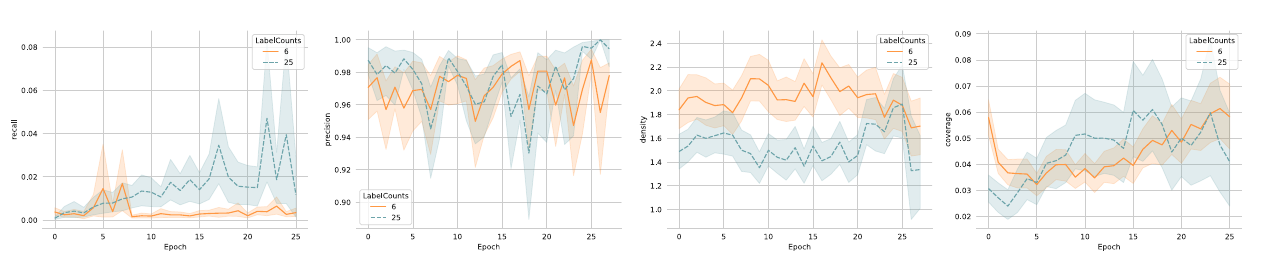}
  % \vspace{\BeforeCaptionVSpace}
  \caption{Comparison of the intervention of architectural element label categories based on the 'FID', 'KID', 'F1 PR', 'F1 DC', 'MMD Linear', and 'MMD RBF' measures}
  \label{fig:LabelCounts1}
\end{figure}

\begin{figure}[!t]
  \centering
  \includegraphics[width=1.0\linewidth]{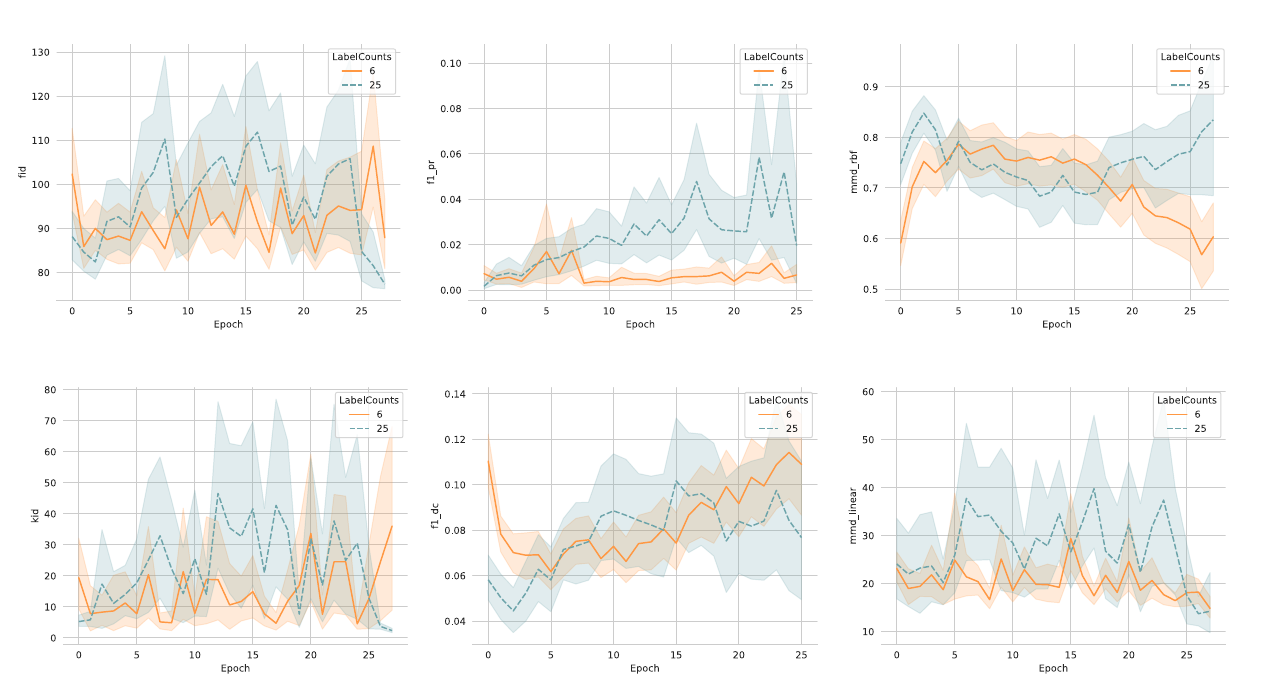}
  % \vspace{\BeforeCaptionVSpace}
  \caption{Comparison of the intervention of architectural element label categories based on the 'precision', 'recall', 'density', and 'coverage' measures}
  \label{fig:LabelCounts2}
\end{figure}

When considering the number of architectural element label categories involved, our analysis reveals that while increasing the number of architectural element label categories involved in the graph representation learning model training,  the ``F1 PR'', ``Precision'', and ``Recall'' values significantly increase (Fig.~\ref{fig:LabelCounts1}, Fig.~\ref{fig:LabelCounts2}), suggesting that the model can achieve a higher fidelity level in graph representation learning. Nevertheless, this improvement in fidelity also comes at a cost to diversity, as evidenced by the substantial increase in ``MMD Linear'' and ``MMD RBF'' values, coupled with a notable decrease in ``F1 DC'', ``Density'', and ``Coverage''. These results indicate that an increase in the number of architectural element label categories can also lead to a trade-off effect in the fidelity and diversity level of the learned graph representations. This phenomenon is somewhat predictable, similar to the situation of involving extra polygon vertices coordinates to node features, as adding more architectural element label categories offers more detailed and relevant information for graph representation yet also imposes a heightened challenge on the representation learning process. Thus, balancing the enhanced fidelity with the increased risk of mode dropping underscores a critical aspect of model design and feature selection in graph representation learning. This balance is crucial for developing graph representation learning models that can effectively capture both the characteristics and diversity of the architectural graph data they are designed to represent.

\begin{figure}[!t]
  \centering
  \includegraphics[width=1.0\linewidth]{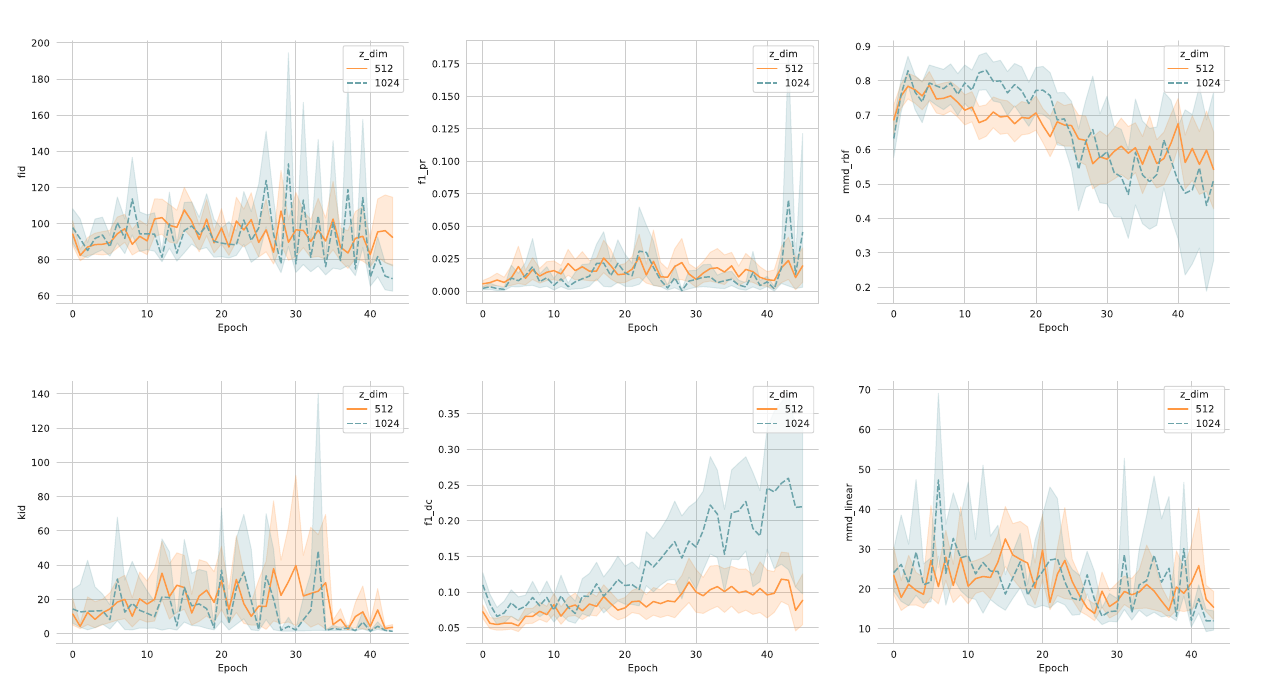}
  % \vspace{\BeforeCaptionVSpace}
  \caption{Comparison of the intervention of the dimension of latent code space based on the 'FID', 'KID', 'F1 PR', 'F1 DC', 'MMD Linear', and 'MMD RBF' measures}
  \label{fig:z_dim1}
\end{figure}

\begin{figure}[!t]
  \centering
  \includegraphics[width=1.0\linewidth]{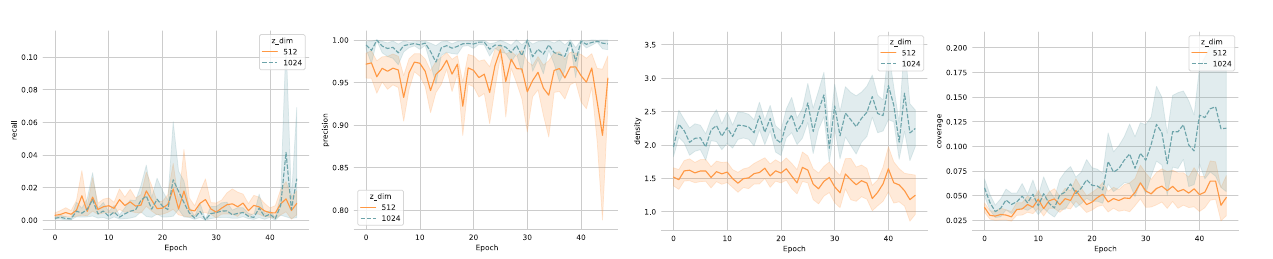}
  % \vspace{\BeforeCaptionVSpace}
  \caption{Comparison of the intervention of the dimension of latent code space based on the 'precision', 'recall', 'density', and 'coverage' measures}
  \label{fig:z_dim2}
\end{figure}

Moreover, evaluation of the impact of varying dimensions of latent codes \textit{z} on learning high-dimensional representations of architectural design data graphs in latent space reveals that increasing the dimension of latent codes \textit{z} significantly improves ``F1 DC'', ``Precision'', ``Density'', and ``Coverage'' values (Fig.~\ref{fig:z_dim1}, Fig.~\ref{fig:z_dim2}). This enhancement indicates that the graph representation learning model is capable of learning graph representations with a better level of diversity; a higher dimensional latent space may provide a more expansive and nuanced space for the model to capture a wider range of variations and complexities present in the architectural design graph data. However, this improvement in diversity comes with a trade-off in terms of fidelity, evidenced by an increase in the ``MMD RBF'' value, which suggests a slight derogation in the fidelity level of the learned graph representations. The increase in ``MMD RBF'' implies that the representations generated by the model in the higher-dimensional latent space are somewhat less similar to the real data distribution, indicating a minor compromise in the accuracy and precision of the representations. Therefore, while a higher dimension of latent codes \textit{z} enhances the model's ability to capture diversity in the graph representations, it also appears to affect the fidelity of the learned representations to some extent. This finding highlights again the delicate balance between achieving a diverse representation and maintaining high fidelity in graph representation learning by underscoring the need for careful consideration of the latent space dimensionality in designing models for architectural design graph data representation.

\begin{figure}[!t]
  \centering
  \includegraphics[width=1.0\linewidth]{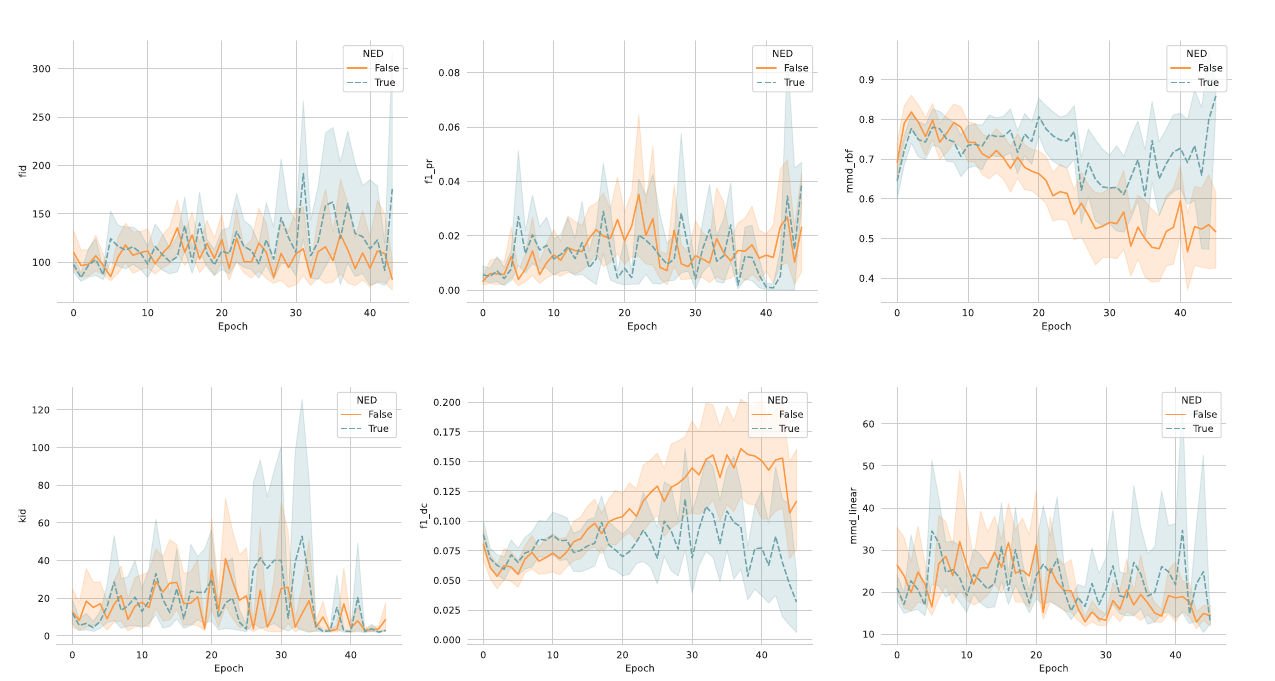}
  % \vspace{\BeforeCaptionVSpace}
  \caption{Comparison of the intervention of incorporating the node-edge co-disentanglement mechanism based on the 'FID', 'KID', 'F1 PR', 'F1 DC', 'MMD Linear', and 'MMD RBF' measures}
  \label{fig:NED1}
\end{figure}

\begin{figure}[!t]
  \centering
  \includegraphics[width=1.0\linewidth]{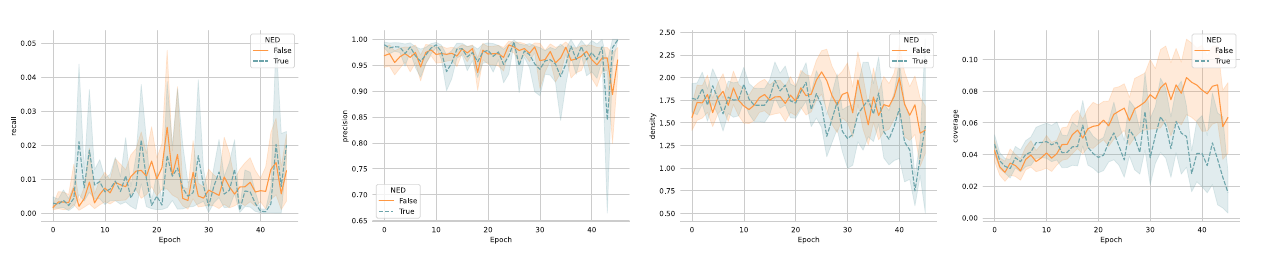}
  % \vspace{\BeforeCaptionVSpace}
  \caption{Comparison of the intervention of incorporating the node-edge co-disentanglement mechanism based on the 'precision', 'recall', 'density', and 'coverage' measures}
  \label{fig:NED2}
\end{figure}

Similar to increasing the latent code dimensionality, the effects of incorporating a node-edge co-disentanglement mechanism into the structure of the graph representation learning model resulted in significant increases in both ``F1 DC'' and ``Coverage'' (Fig.~\ref{fig:NED1}, Fig.~\ref{fig:NED2}). This indicates an enhancement in the diversity level of the graph representations learned by the model, allowing for a more nuanced and detailed representation of the relationships and interactions between nodes and edges within the graph, which, in turn, facilitates the model's ability to capture a broader spectrum of variations and intricacies inherent in the architectural design data. Yet, this increased diversity comes at a certain cost to fidelity, with an accompanying rise in the ``MMD RBF'' value, meaning that the representations generated by the model with the node-edge co-disentanglement mechanism are less congruent with the real data distribution, indicating a minor compromise in how accurately and precisely the model captures the details of the architectural designs. Therefore, implementing a node-edge co-disentanglement mechanism in the graph representation learning model also creates a trade-off between diversity and fidelity. While it significantly enriches the diversity of the representations, enabling the model to encompass a wider range of patterns and relationships, it also slightly impacts the fidelity of these representations. This trade-off underscores the complexity of designing graph representation learning models, particularly in balancing the need to capture diverse architectural elements while maintaining high accuracy and precision.

To deepen our understanding of the impacts of various modelling choices on graph representation learning performance, we compare different model design choices with One-way ANOVA analysis across all possible combination groups, allowing us to systematically compare the effects of varied design choices on different graph representation learning performance metrics. One-way ANOVA analysis offers the ability to determine the statistical significance of differences among different graph representation model design choice combinations. Concretely, the One-way ANOVA results of all group comparisons yield significant F statistics (Table \ref{table:ANOVA}), indicating that the differences in performance metrics across various model design choice combinations – whether they involve structural modifications or feature enhancements – are statistically significant and not due to random variations.

\begin{table}
\caption{One-way ANOVA results of all possible combination groups of graph representation model structure and feature intervention choices over different graph representation learning model performance evaluation metrics}
\centering
\begin{tabular}{ccc}
\hline\hline
Metric&  F statistics& p-value\\
\hline\hline
FID&  12.23& .00\\
\hline
MMD Linear&  6.08& .00\\
\hline
MMD RBF&  107.75& .00\\
\hline
F1 PR&  36.90& .00\\
\hline
F1 DC&  123.18& .00\\
\hline
Precision&  13.25& .00\\
\hline
Recall&  22.47& .00\\
\hline
Density&  103.09& .00\\
\hline
Coverage&  122.20& .00\\
\hline\hline
\end{tabular}
\label{table:ANOVA}
\end{table}

\section{Qualitative evaluation}
\label{qual_app}

\begin{figure}[!t]
  \centering
  \includegraphics[width=0.5\linewidth]{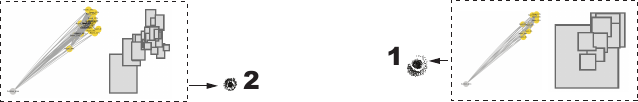}
  % \vspace{\BeforeCaptionVSpace}
  \caption{Generated graph samples and their corresponding locations in the learned latent space using a trained framework with edge-augmented encoder, vector quantisation disentanglement module, MLP-based decoder, SVD embeddings, 6 categories of architectural elements and extra features of polygon vertices' coordinates}
  \label{fig:sum_svd8_vqvae_FP6_undirected_fullNode_multiEdge_512_21}
\end{figure}

\begin{figure}[!t]
  \centering
  \includegraphics[width=0.8\linewidth]{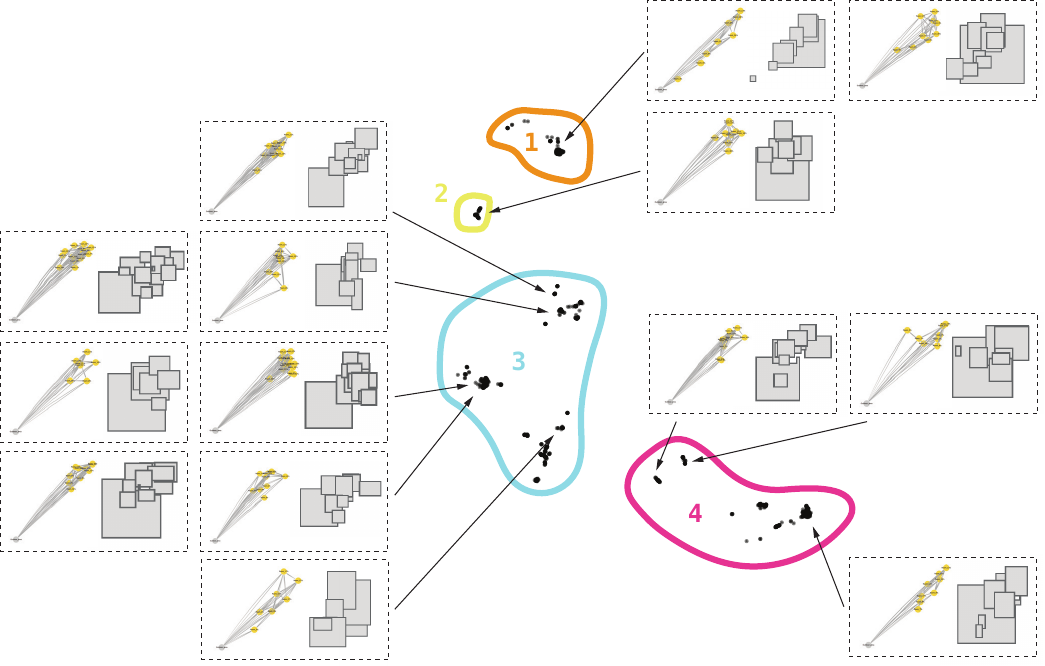}
  % \vspace{\BeforeCaptionVSpace}
  \caption{Generated graph samples and their corresponding locations in the learned latent space using a trained framework with edge-augmented encoder, vanilla VAE disentanglement module, MLP-based decoder, SVD embeddings, 6 categories of architectural elements, extra features of polygon vertices' coordinates, and boosted dimensions of the latent space}
  \label{fig:sum_svd8_betavae_FP6_undirected_fullNode_multiEdge_1024_40}
\end{figure}

Similar to the model with the VQ mechanism, the model with extra features of polygon vertices' coordinates also demonstrates a significant level of disentanglement but with a comparable limitation in diversity (Fig.~\ref{fig:sum_svd8_vqvae_FP6_undirected_fullNode_multiEdge_512_21}). The expansion of the latent space dimensions, combined with SVD embeddings and additional features, results in a boosted performance in terms of diversity while maintaining a high level of disentanglement. This setup seems to strike a balance between diverse clustering and clear disentanglement (Fig.~\ref{fig:sum_svd8_betavae_FP6_undirected_fullNode_multiEdge_1024_40}).

\begin{figure}[!t]
  \centering
  \includegraphics[width=1.0\linewidth]{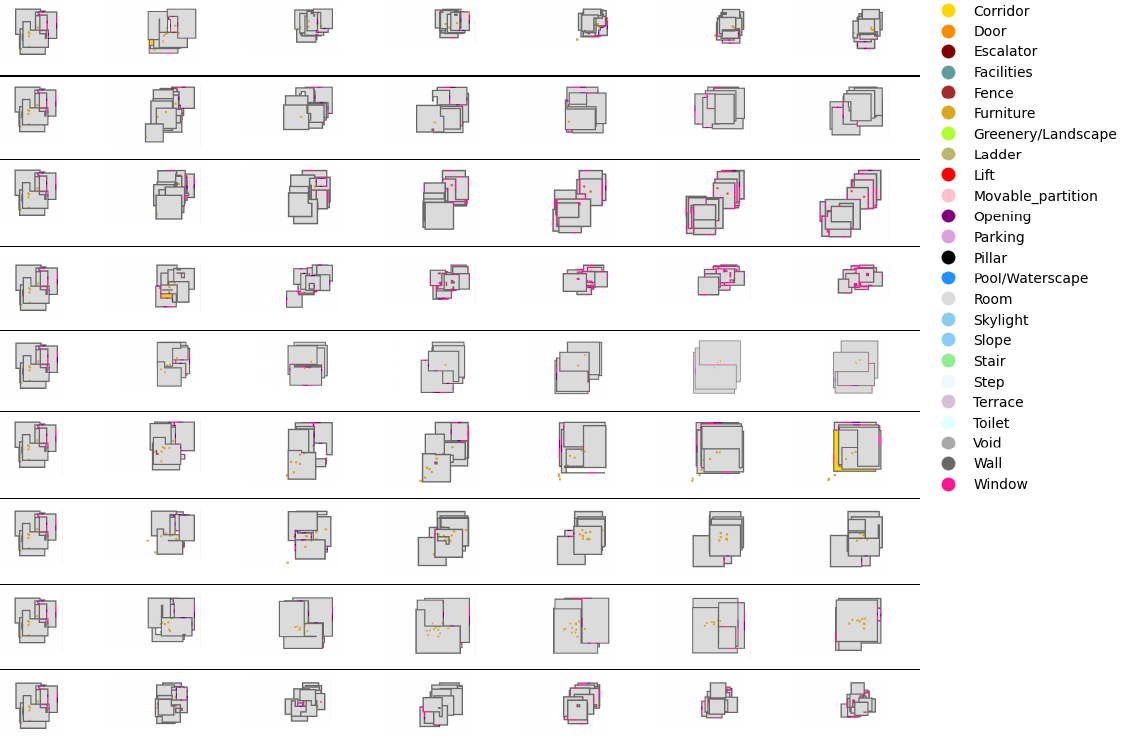}
  % \vspace{\BeforeCaptionVSpace}
  \caption{Linear interpolation samples starting from the same latent code $z$ of the learned latent space of a trained framework with edge-augmented encoder, vanilla VAE disentanglement module, MLP-based decoder, SVD embeddings, 25 categories of architectural elements, extra features of polygon vertices' coordinates, and boosted dimensions of the latent space}
  \label{fig:FP-interpolation_sum_svd8_betavae_FP25_undirected_fullNode_multiEdge_1024_one}
\end{figure}

\begin{figure}[!t]
  \centering
  \includegraphics[width=1.0\linewidth]{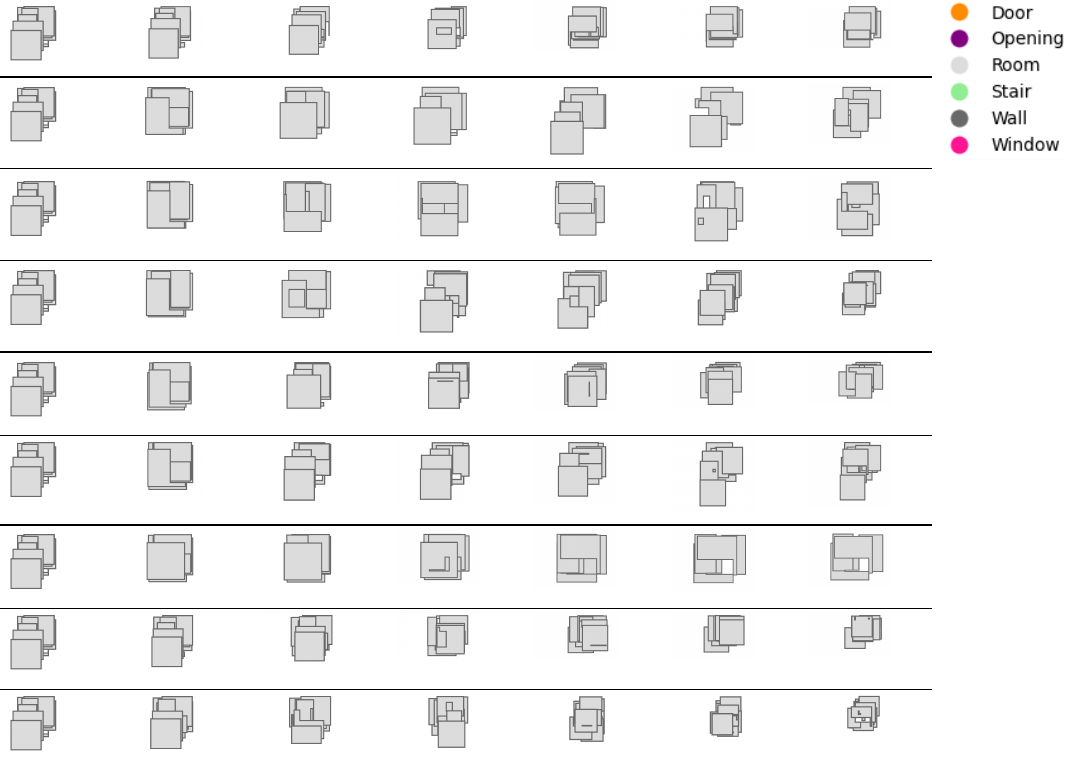}
  % \vspace{\BeforeCaptionVSpace}
  \caption{Linear interpolation samples starting from the same latent code $z$ of the learned latent space of a trained framework with edge-augmented encoder, vanilla VAE disentanglement module, MLP-based decoder, SVD embeddings, 6 categories of architectural elements, extra features of polygon vertices' coordinates, and boosted dimensions of the latent space}
  \label{fig:FP-interpolation_sum_svd8_betavae_FP6_undirected_fullNode_multiEdge_1024_one}
\end{figure}

\begin{figure}[!t]
  \centering
  \includegraphics[width=1.0\linewidth]{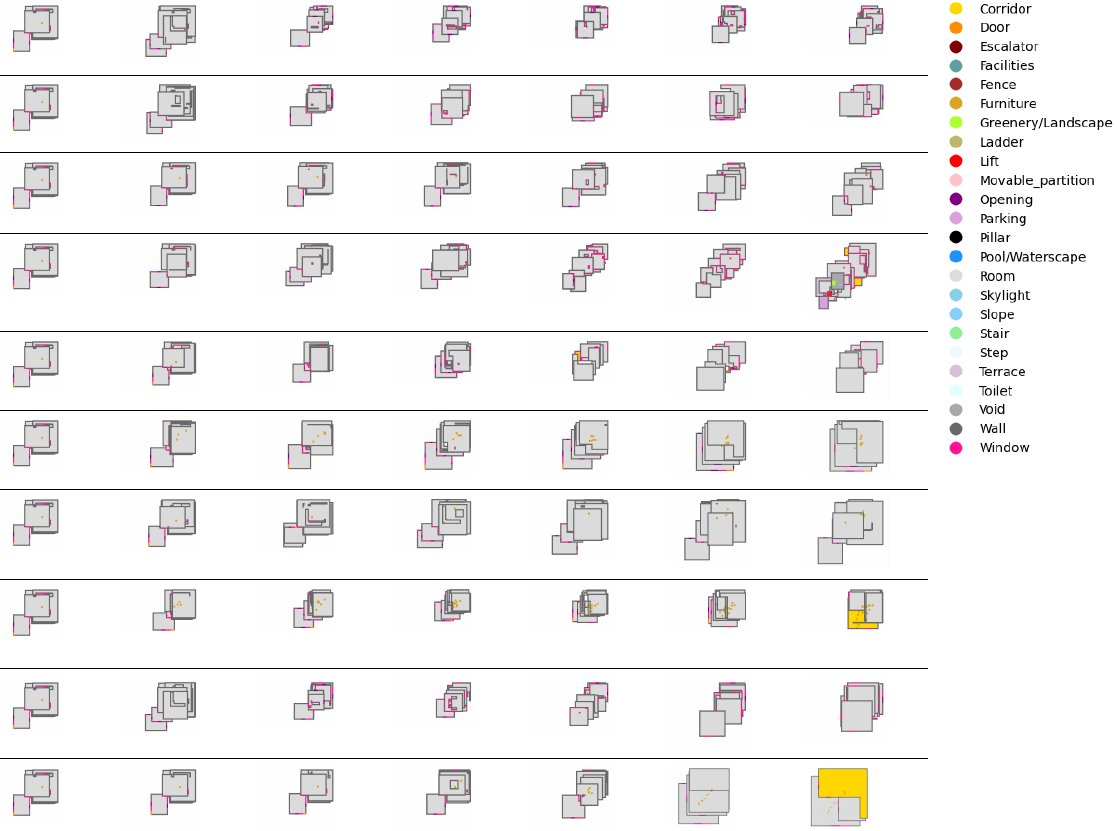}
  % \vspace{\BeforeCaptionVSpace}
  \caption{Linear interpolation samples starting from the same latent code $z$ of the learned latent space of a trained framework with edge-augmented encoder, vanilla VAE disentanglement module, MLP-based decoder, SVD embeddings, and 25 categories of architectural elements}
  \label{fig:FP-interpolation_sum_svd8_betavae_FP25_undirected_basicNode_multiEdge_512_one}
\end{figure}

\begin{figure*}[!t]
  \centering
  \includegraphics[width=1.0\linewidth]{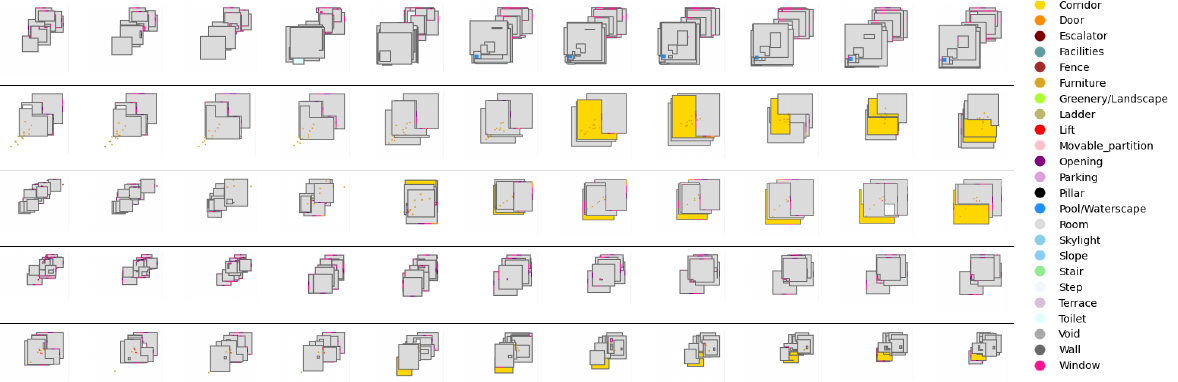}
  % \vspace{\BeforeCaptionVSpace}
  \caption{Linear interpolation samples between pairs of randomly generated latent codes $z$ of the learned latent space of a trained framework with edge-augmented encoder, vanilla VAE disentanglement module, MLP-based decoder, SVD embeddings, 25 categories of architectural elements, extra features of polygon vertices' coordinates, and boosted dimensions of the latent space}
  \label{fig:FP-interpolation_sum_svd8_betavae_FP25_undirected_fullNode_multiEdge_1024_paired}
\end{figure*}

\begin{figure*}[!t]
  \centering
  \includegraphics[width=1.0\linewidth]{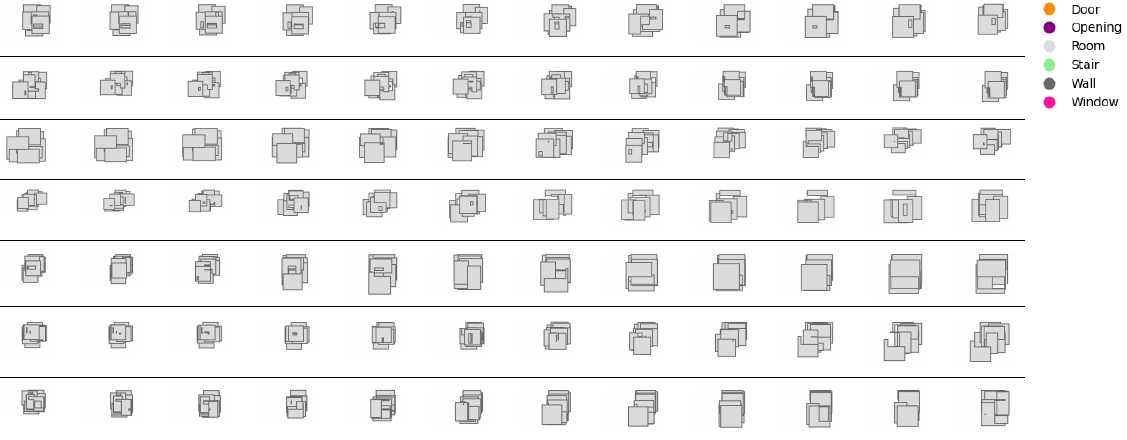}
  % \vspace{\BeforeCaptionVSpace}
  \caption{Linear interpolation samples between pairs of randomly generated latent codes $z$ of the learned latent space of a trained framework with edge-augmented encoder, vanilla VAE disentanglement module, MLP-based decoder, SVD embeddings, 6 categories of architectural elements, extra features of polygon vertices' coordinates, and boosted dimensions of the latent space}
  \label{fig:FP-interpolation_sum_svd8_betavae_FP6_undirected_fullNode_multiEdge_1024_paired}
\end{figure*}

\begin{figure*}[!t]
  \centering
  \includegraphics[width=1.0\linewidth]{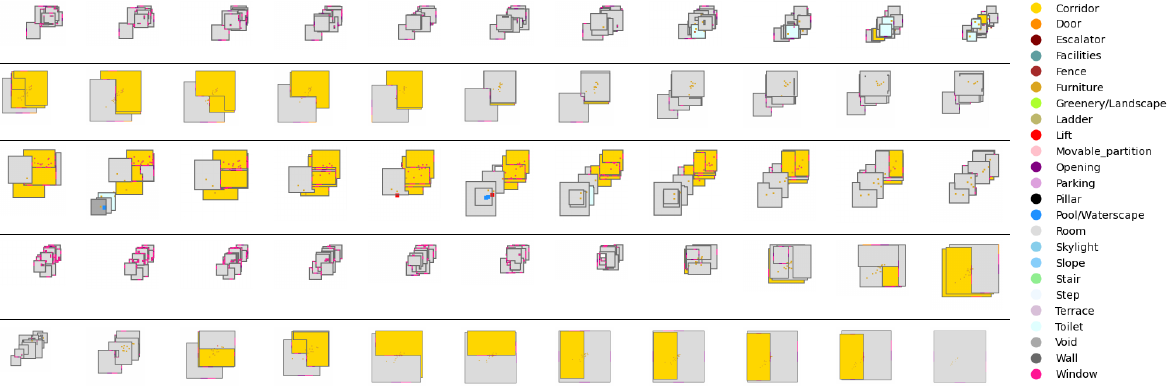}
  % \vspace{\BeforeCaptionVSpace}
  \caption{Linear interpolation samples between pairs of randomly generated latent codes $z$ of the learned latent space of a trained framework with edge-augmented encoder, vanilla VAE disentanglement module, MLP-based decoder, SVD embeddings, and 25 categories of architectural elements}
  \label{fig:FP-interpolation_sum_svd8_betavae_FP25_undirected_basicNode_multiEdge_512_paired}
\end{figure*}

\begin{figure}[!t]
  \centering
  \includegraphics[width=1.0\linewidth]{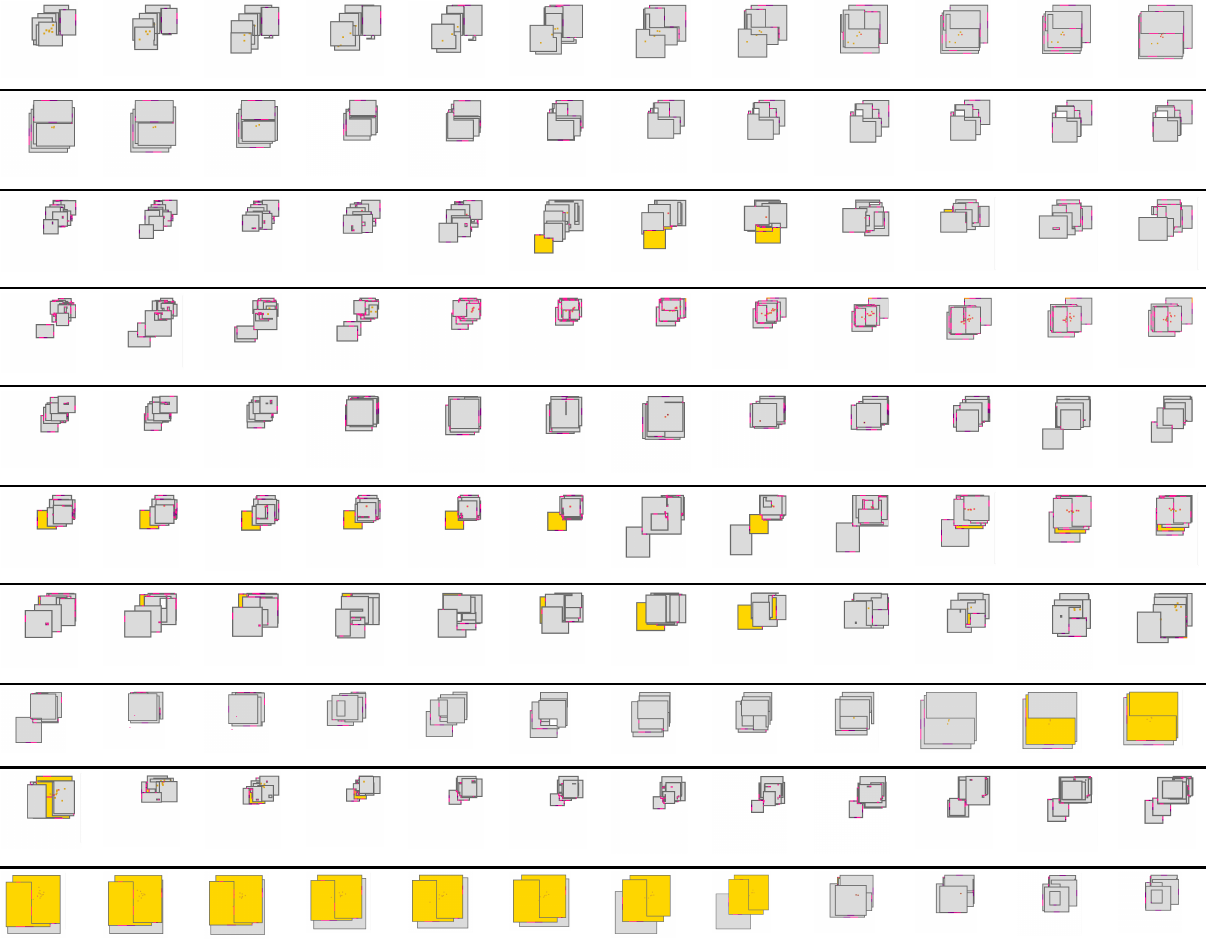}
  % \vspace{\BeforeCaptionVSpace}
  \caption{Linear interpolation samples of the learned latent space of a trained framework with edge-augmented encoder, vanilla VAE disentanglement module, MLP-based decoder, SVD embeddings, 25 categories of architectural elements, extra features of polygon vertices' coordinates, and boosted dimensions of the latent space}
  \label{fig:FP-interpolation_sum_svd8_betavae_FP25_undirected_fullNode_multiEdge_1024_extra}
\end{figure}

\begin{figure}[!t]
  \centering
  \includegraphics[width=1.0\linewidth]{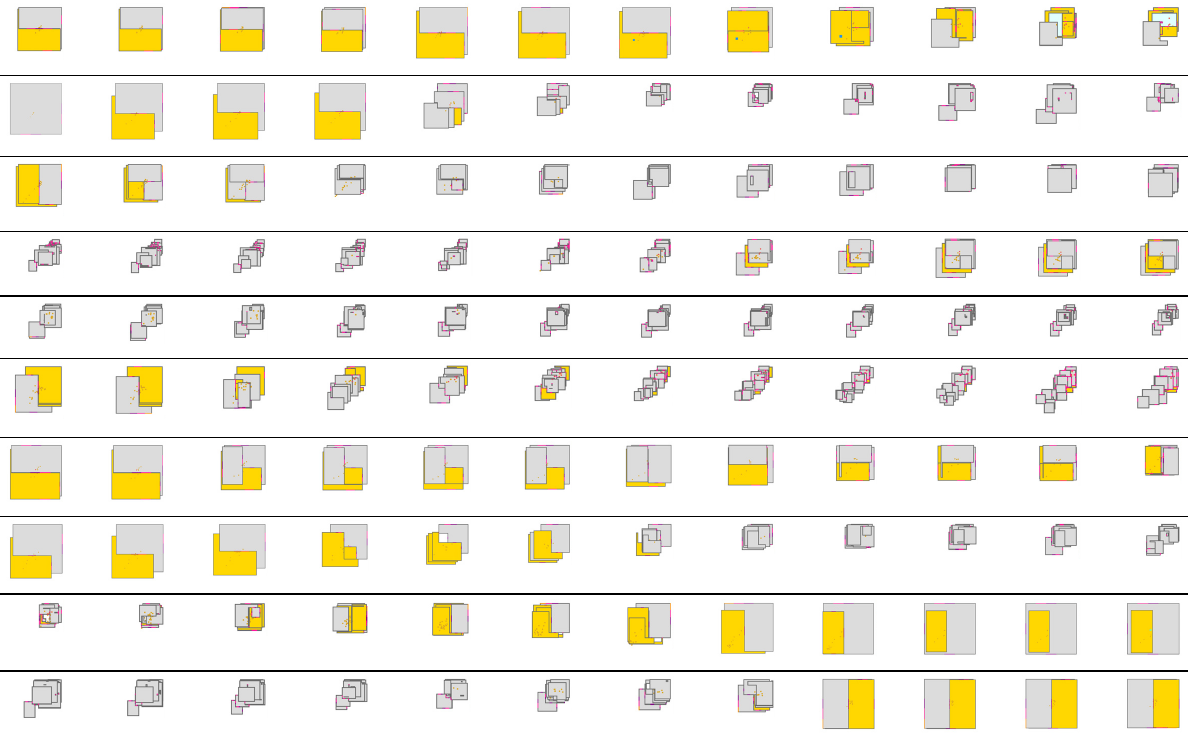}
  % \vspace{\BeforeCaptionVSpace}
  \caption{Linear interpolation samples of the learned latent space of a trained framework with edge-augmented encoder, vanilla VAE disentanglement module, MLP-based decoder, SVD embeddings, and 25 categories of architectural elements}
  \label{fig:FP-interpolation_sum_svd8_betavae_FP25_undirected_basicNode_multiEdge_512_extra}
\end{figure}

\begin{figure}[!t]
  \centering
  \includegraphics[width=1.0\linewidth]{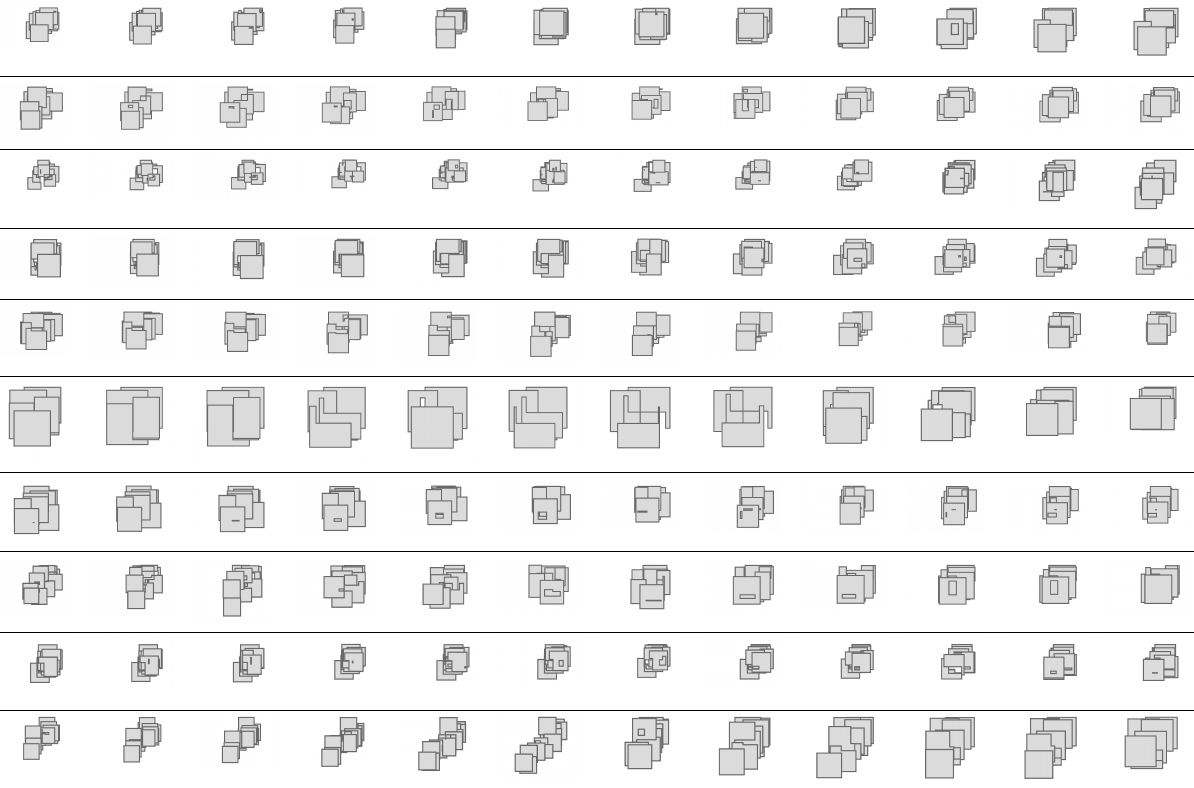}
  % \vspace{\BeforeCaptionVSpace}
  \caption{Linear interpolation samples of the learned latent space of a trained framework with edge-augmented encoder, vanilla VAE disentanglement module, MLP-based decoder, SVD embeddings, 6 categories of architectural elements, extra features of polygon vertices' coordinates, and boosted dimensions of the latent space}
  \label{fig:FP-interpolation_sum_svd8_betavae_FP6_undirected_fullNode_multiEdge_1024_extra}
\end{figure}

More linear interpolation samples are demonstrated in Fig.~\ref{fig:FP-interpolation_sum_svd8_betavae_FP25_undirected_fullNode_multiEdge_1024_one}, Fig.~\ref{fig:FP-interpolation_sum_svd8_betavae_FP6_undirected_fullNode_multiEdge_1024_one}, Fig.~\ref{fig:FP-interpolation_sum_svd8_betavae_FP25_undirected_basicNode_multiEdge_512_one}, Fig.~\ref{fig:FP-interpolation_sum_svd8_betavae_FP25_undirected_fullNode_multiEdge_1024_paired}, Fig.~\ref{fig:FP-interpolation_sum_svd8_betavae_FP6_undirected_fullNode_multiEdge_1024_paired}, Fig.~\ref{fig:FP-interpolation_sum_svd8_betavae_FP25_undirected_basicNode_multiEdge_512_paired}, Fig.~\ref{fig:FP-interpolation_sum_svd8_betavae_FP25_undirected_fullNode_multiEdge_1024_extra}, Fig.~\ref{fig:FP-interpolation_sum_svd8_betavae_FP25_undirected_basicNode_multiEdge_512_extra}, and Fig.~\ref{fig:FP-interpolation_sum_svd8_betavae_FP6_undirected_fullNode_multiEdge_1024_extra}, corresponding to what has been discussed in section \ref{results:gnn:qual}.

\section{Attributed adjacency multi-graph datasets}
\label{AAG_app}

\begin{figure}[!t]
  \centering
  \includegraphics[width=1.0\linewidth]{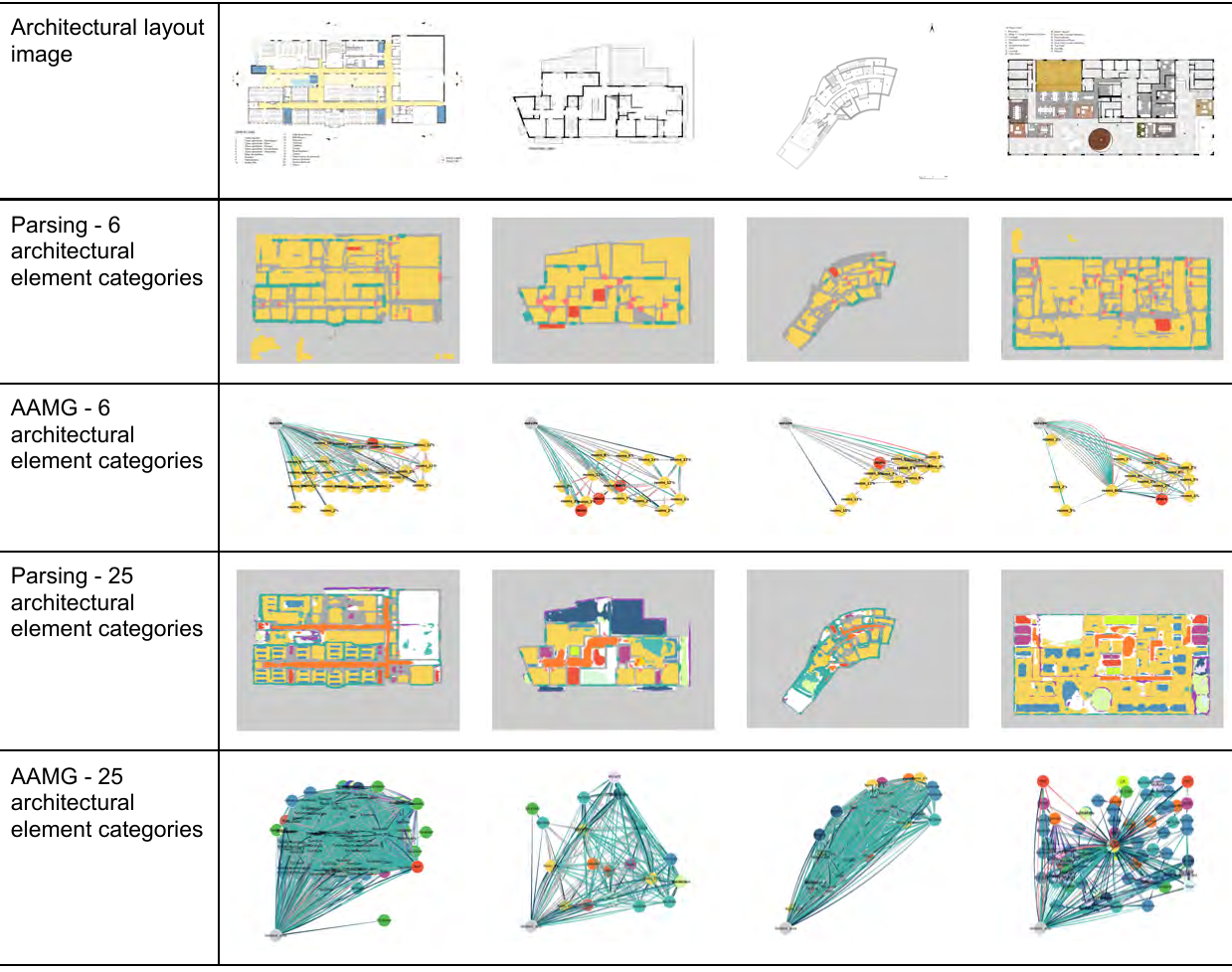}
  % \vspace{\BeforeCaptionVSpace}
  \caption{Randomly selected floor plan samples with corresponding parsing and AAMG extraction outputs}
  \label{fig:parsing_pipeline1samples}
\end{figure}

\begin{figure}[!t]
  \centering
  \includegraphics[width=1.0\linewidth]{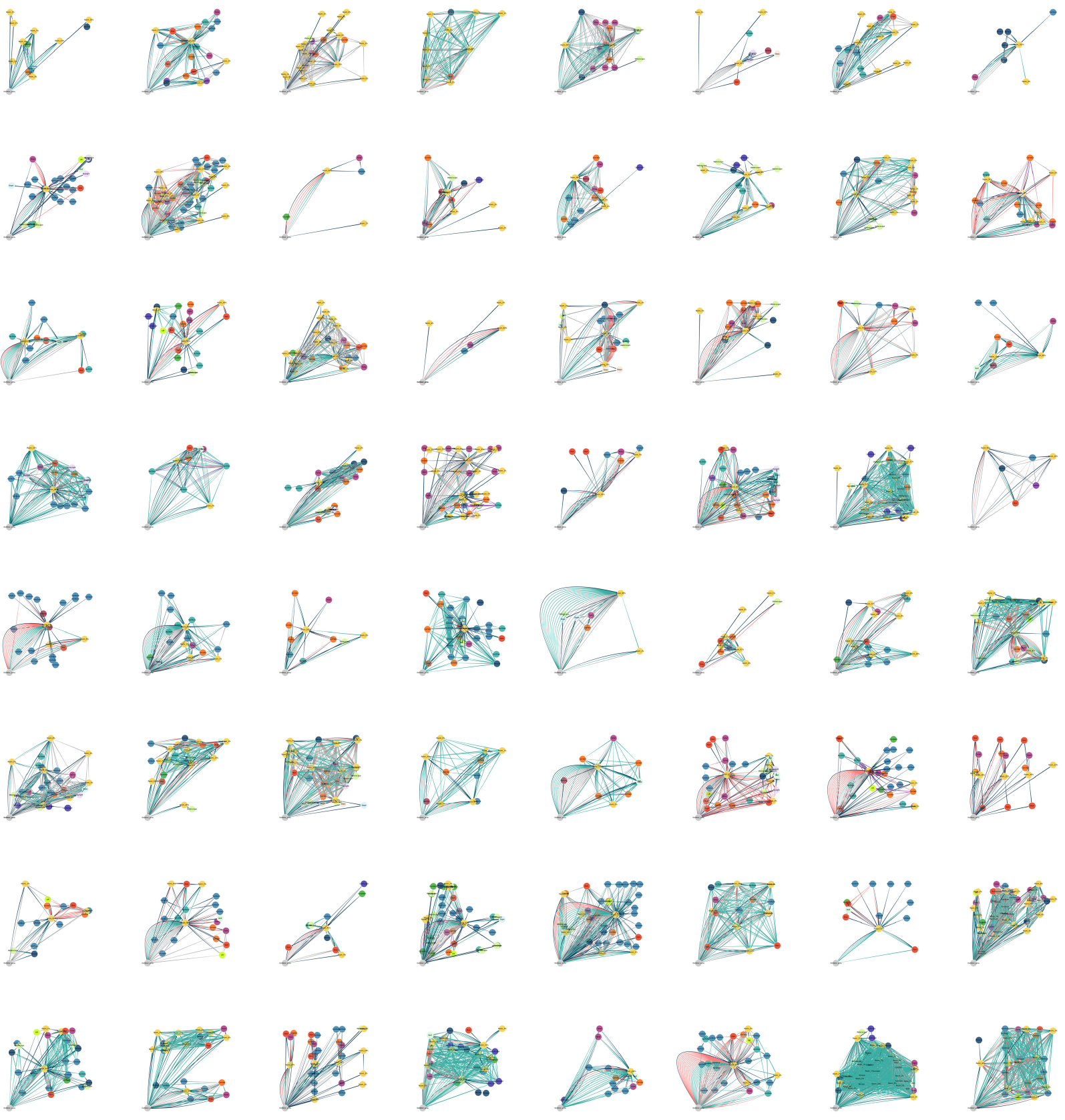}
  % \vspace{\BeforeCaptionVSpace}
  \caption{Selected AAMG samples of the training dataset with 25 architectural element categories}
  \label{fig:multigraph_samples}
\end{figure}

Some floor plan image samples with corresponding parsing and attributed adjacency multi-graph (AAMG) extraction outputs are shown in Fig.~\ref{fig:parsing_pipeline1samples}. 
More AAMG samples extracted from the floor plan image repository are shown in Fig.~\ref{fig:multigraph_samples}. The graph datasets can also be used to explore other learning-based design tools or training tasks.

\section{Conversion from generated attributed adjacency multigraph to floor plans}
\label{study2:results:gnn:visualization}

Converting the generated attributed adjacency graph into graphical floor plans is a strategic step for the qualitative evaluation of model performance, particularly concerning the interpretation of graph data space. This conversion can be essential as graphical floor plans provide a clear and tangible representation of the complex information in the adjacency and node feature matrices. This visual form allows a more intuitive understanding of the graph's spatial relationships and encoded architectural elements. By representing the generated graph data as graphical floor plans, qualitatively evaluating the model’s performance becomes significantly easier, enabling direct comparison between the model-generated graph layouts. This comparison is vital for assessing the fidelity of the model's interpretation, identifying areas of strength, and pinpointing aspects that may require further refinement.

Typically, the task of automatically converting an attributed adjacency graph into practical floor plan layouts involves interpreting the graph––where nodes represent different spaces and edges represent various connections––and translating this abstract representation into a coherent, spatially accurate floor plan. This reverse engineering process is conventionally not trivial. One primary challenge in converting attributed adjacency graphs back into floor plans is the non-uniqueness of the task, as a single graph can correspond to multiple feasible floor plan layouts, each varying in spatial arrangement while adhering to the same structural relationships and constraints defined by the graph. However, our generated graphs contain more detailed positional and area information for each space node and specified space type. This level of detail aids significantly in reducing the ambiguity typically associated with the conversion, providing a clearer blueprint for the corresponding floor plan layout. Despite the detailed information available in the graphs, a certain level of compromise might still be required to adapt the abstract graph data into functional floor plans. This involves balancing the rigid constraints of the graph with the practical considerations of architectural design.
% \cite{nauata2021house, tang2024graph}

\begin{figure}[!t]
  \centering
  \includegraphics[width=1.0\linewidth]{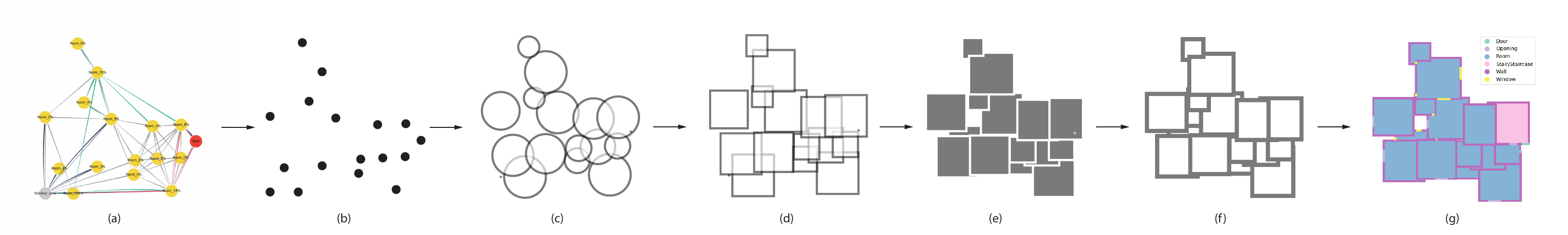}
  % \vspace{\BeforeCaptionVSpace}
  \caption{Converting attributed adjacency graphs with 6 distinct architectural design elements into floor plan layouts}
  \label{fig:GRAPH2PLAN6}
\end{figure}

\begin{figure}[!t]
  \centering
  \includegraphics[width=1.0\linewidth]{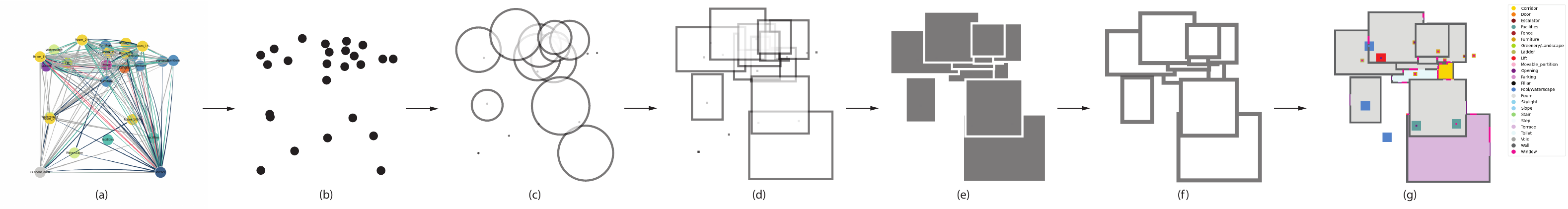}
  % \vspace{\BeforeCaptionVSpace}
  \caption{Converting attributed adjacency graphs with 25 distinct architectural design elements into floor plan layouts}
  \label{fig:GRAPH2PLAN25}
\end{figure}

For the task of converting attributed adjacency graphs back into floor plan layouts, our approach is designed to be straightforward and efficient, primarily serving the purpose of qualitative evaluation of model performance in interpreting graph data space. This simplified conversion process is not the primary focus of our study but rather a means to validate the effectiveness of our model (Fig.~\ref{fig:GRAPH2PLAN6} and Fig.~\ref{fig:GRAPH2PLAN25}). We begin by extracting the centre point coordinates of all space nodes from the corresponding node feature matrix. Each space node is represented as a simple located point, streamlining the initial layout. In this first step, we exclude the outdoor node, which will be used as a reference point to generate different connection elements (such as walls, doors, windows, etc.) for spaces that are connected to the outdoor node. We utilize the area ratio data from the node feature matrix for space nodes requiring more specific area information. This information is used to convert points into rectangles with areas proportional to the provided ratios. This step excludes certain space nodes like elevators, staircases, and toilets. These spaces typically have standard occupied areas and do not require detailed area information for our purpose. Afterwards, we conduct a series of shape transformations based on the information provided by the corresponding adjacency matrix. If two rectangles overlap, we divide their overlapping area and adjust their boundaries accordingly. For rectangles representing spaces connected in the graph but not adjacent in the layout, we translate their sides to reflect these connections. The boundaries of the generated polygon shapes are then buffered to represent walls, providing a structural outline to the layout. This is followed by segmenting the adjacency sides of pairs of polygons connected by $k$ different edge types (other than ``wall'') into $k+1$ segments. Each segment is then assigned a connection type like ``door'', ``window'', or ``opening'' if indicated by the adjacency matrix. This segmentation also applies to polygons connected to the outdoor node. After all these steps, the remaining space nodes whose occupied areas are usually standard will be plotted as rectangles with consistent areas for demonstration purposes.

\vfill

\end{document}